\documentclass[twoside,11pt]{article}

\usepackage{jmlr2e}
\newcommand{\BlackBox}{\rule{1.5ex}{1.5ex}}  
\newenvironment{JMLRproof}{\par\noindent{\bf Proof\ }}{\hfill\BlackBox\\[2mm]}

\ShortHeadings{The exploding gradient problem demystified}{Philipp, Song and Carbonell}
\firstpageno{1}

\usepackage{varioref}
\usepackage{hyperref}
\usepackage{url}
\usepackage{amsfonts}
\usepackage{amsthm}
\usepackage{amsmath}
\usepackage[linesnumbered]{algorithm2e}
\usepackage{graphicx}
\usepackage{adjustbox}
\usepackage{array}
\usepackage{multirow}
\usepackage{placeins}
\usepackage{enumitem}
\usepackage{bbm}
\adjustboxset*{center}
\usepackage{xcolor}

\author{George Philipp\footnote{work done while at University of California, Berkeley} \\
Carnegie Mellon University\\
\texttt{george.philipp@email.de}\\
\And
Dawn Song \\
University of California, Berkeley \\
\texttt{dawnsong@gmail.com}
\And
Jaime G. Carbonell\\
Carnegie Mellon University\\
\texttt{jgc@cs.cmu.edu}
}

\allowdisplaybreaks

\makeatletter
\newtheorem*{rep@theorem}{\rep@title}
\newcommand{\newreptheorem}[2]{%
\newenvironment{rep#1}[1]{%
 \def\rep@title{#2 \ref{##1}}%
 \begin{rep@theorem}}%
 {\end{rep@theorem}}}
\makeatother

\newtheorem{theorem}{Theorem}
\newreptheorem{theorem}{Theorem}
\newtheorem{proposition}{Proposition}
\newreptheorem{proposition}{Proposition}

\newtheorem{corollary}{Corollary}

\begin{document}

\title{The exploding gradient problem demystified - definition, prevalence, impact, origin, tradeoffs, and solutions}

\author{\name George Philipp \thanks{corresponding author; majority of work completed while at UC Berkeley} \email george.philipp@email.de \\
       \addr School of Computer Science\\
       Carnegie Mellon University\\
       Pittsburgh, PA, 15213, USA
       \AND
      \name Dawn Song \email dawnsong@gmail.com \\
       \addr Department of Electrical Engineering and Computer Science\\
       University of California, Berkeley\\
       Berkeley, CA, 94720, USA
       \AND
       \name Jaime G. Carbonell \email jgc@cs.cmu.edu \\
       \addr School of Computer Science\\
       Carnegie Mellon University\\
       Pittsburgh, PA, 15213, USA}

\editor{TBD}

\maketitle

\begin{abstract}
Whereas it is believed that techniques such as Adam, batch normalization and, more recently, SELU nonlinearities ``solve'' the exploding gradient problem, we show that this is not the case in general and that in a range of popular MLP architectures, exploding gradients exist and that they limit the depth to which networks can be effectively trained, both in theory and in practice. We explain why exploding gradients occur and highlight the {\it collapsing domain problem}, which can arise in architectures that avoid exploding gradients. 

ResNets have significantly lower gradients and thus can circumvent the exploding gradient problem, enabling the effective training of much deeper networks. We show this is a direct consequence of the Pythagorean equation. By noticing that {\it any neural network is a residual network}, we devise the {\it residual trick}, which reveals that introducing skip connections simplifies the network mathematically, and that this simplicity may be the major cause for their success.\\
\end{abstract}


\section{Introduction}

Arguably, the primary reason for the success of neural networks is their ``depth'', i.e. their ability to compose and jointly train nonlinear functions so that they co-adapt. A large body of work has detailed the benefits of depth (e.g. \cite{depth1,depth2,depth3,depth4,depth5,depth6,depth7}).

The exploding gradient problem has been a major challenge for training very deep feedforward neural networks at least since the advent of gradient-based parameter learning \citep{hochreiterThesis}. In a nutshell, it describes the phenomenon that as the gradient is backpropagated through the network, it may grow exponentially from layer to layer. This can, for example, make the application of vanilla SGD impossible. Either the step size is too large for updates to lower layers to be useful or it is too small for updates to higher layers to be useful. While this intuitive notion is widely understood, there are significant gaps in the understanding of this important phenomenon. In this paper, we take a significant step towards closing those gaps.

\paragraph{Defining exploding gradients} To begin with, there is no well-accepted metric for determining the presence of pathological exploding gradients. Should we care about the length of the gradient vector? Should we care about the size of individual components of the gradient vector? Should we care about the eigenvalues of the Jacobians of individual layers? Depending on the metric used, different strategies arise for combating exploding gradients. For example, manipulating the width of layers as suggested by e.g. \cite{widthHacking,pyramidal} can greatly impact the size of gradient vector components but leaves the length of the gradient vector relatively unchanged. 

The problem is that it is unknown whether exploding gradients, when defined according to any of these metrics, necessarily lead to training difficulties. There is much evidence that gradient explosion when defined according to some metrics is associated with poor results when certain architectures are paired with certian optimization algorithms (e.g. \cite{meanFieldExplosion,xavierInit}). But, can we make general statements about entire classes of algorithms and architectures?

\paragraph{Prevalence} It has become a common notion that techniques such as introducing normalization layers (e.g. \cite{batchnorm}, \cite{layernorm}, \cite{cosineNormalization}, \cite{weightNormalization}) or careful initial scaling of weights (e.g. \citet{heInit}, \citet{xavierInit}, \cite{orthogonalInitialization}, \cite{saneInit}) largely eliminate exploding gradients by stabilizing forward activations. This notion was espoused in landmark papers. The paper that introduced batch normalization \citep{batchnorm} states: \\


{\it In traditional deep networks, too-high learning rate may result in the gradients that explode or vanish, as well as getting stuck in poor local minima. Batch Normalization helps address these issues.}\\

The paper that introduced ResNet \citep{resNet} states: \\

{\it Is learning better networks as easy as stacking more layers? An obstacle to answering this question was the notorious problem of vanishing/exploding gradients, which hamper convergence from the beginning. This problem, however, has been largely addressed by normalized initialization and intermediate normalization layers, ...}\\

We argue that these claims are too optimistic. While scaling weights or introducing normalization layers can reduce gradient growth defined according to some metrics in some situations, these techniques are not effective in general and can cause other problems even when they are effective. We intend to add nuance to these ideas which have been widely adopted by the community (e.g. \cite{cosineNormalization,shattering}). In particular, we intend to correct the misconception that stabilizing forward activations is sufficient for avoiding exploding gradients (e.g. \cite{selu}).

\paragraph{Impact} Algorithms such as RMSprop \citep{RMSprop}, Adam \citep{Adam} or vSGD \citep{vSGD} are light modifications of SGD that rescale different parts of the gradient vector and are known to be able to lead to improved training outcomes. This raises an important unanswered question. Are exploding gradients merely a numerical quirk to be overcome by simply rescaling different parts of the gradient vector or are they reflective of an inherently difficult optimization problem that cannot be easily tackled by simple modifications to a stock algorithm?

\paragraph{Origins and tradeoffs} The exploding gradient problem is often discussed in conjunction with the vanishing gradient problem, and often the implication is that the best networks exist on the edge between the two phenomena and exhibit stable gradients (e.g. \cite{xavierInit}, \cite{meanFieldExplosion}). But is avoiding gradient pathology simply a matter of designing the network to have a rate of gradient change per layer as close to one as possible, or are there more fundamental reasons why gradient pathology arises in so many popular architectures? Are there tradeoffs that cannot be escaped as easily as the exploding / vanishing gradient dichotomy suggests?

\paragraph{Solutions} ResNet \citep{resNet} and other neural network architectures utilizing skip connections (e.g. \cite{denseNet}, \cite{inceptionv4}) have been highly successful recently. While the performance of networks without skip connections starts to degrade when depth is increased beyond a certain point, the performance of ResNet continues to improve until a much greater depth is reached. While favorable changes to properties of the gradient brought about by the introduction of skip connections have been demonstrated for specific architectures (e.g. \cite{meanFieldResNet,shattering}), a general explanation of the power of skip connections has not been given.\\

Our contributions are as follows:

\begin{enumerate}
\item We introduce the {\it `gradient scale coefficient' (GSC)}, a novel measurement for assessing the presence of pathological exploding gradients (section \ref{GSCsection}). It is robust to confounders such as network scaling (section \ref{GSCsection}) and layer width (section \ref{explodingsection}) and can be used directly to show that training is difficult (section \ref{shallowsection}). We propose that this metric can standardize research on gradient pathology.
\item We demonstrate that exploding gradients are in fact present in a variety of popular MLP architectures, including architectures utilizing techniques that supposedly combat exploding gradients. We show that introducing normalization layers may even exacerbate the exploding gradient problem (section \ref{explodingsection}).
\item We introduce the {\it `residual trick'} (section \ref{shallowsection}), which reveals that arbitrary networks can be viewed as residual networks. This enables the application of analysis devised for ResNet, including the popular notion of `effective depth', to arbitrary networks.
\item We show that exploding gradients as defined by the GSC are not a numerical quirk to be overcome by rescaling different parts of the gradient vector, but are indicative of an inherently complex optimization problem and that they limit the depth to which MLP architectures can be effectively trained, rendering very deep MLPs effectively much shallower (section \ref{shallowsection}).
\item For the first time, we explain why exploding gradients are likely to occur in deep networks even when the forward activations do not explode (section \ref{widerViewsection}). We argue that this is a fundamental reason for the difficulty of constructing very deep trainable networks.
\item We define the {\it `collapsing domain problem'} for training very deep feedforward networks. We show how this problem can arise precisely in architectures that avoid exploding gradients and that it can be at least as damaging to the training process (section \ref{collapsingSection}).
\item For the first time, we show that the introduction of skip connections has a strong gradient-reducing effect on general deep network architectures and that this follows directly from the Pythagorean equation (section \ref{ResNetsection}). 
\item We reveal that ResNets are mathematically simpler version of networks without skip connections and thus approximately achieve what we term the {\it `orthogonal initial state'}. This provides, we argue, the major reason for their superior performance at great depths as well as an important criterion for neural network design in general (section \ref{ResNetsection}).
\end{enumerate}

In section \ref{relatedWorksection}, we discuss related work. In section \ref{conclusionsection}, we conclude and derive practical recommendations for designing and training deep networks as well as key implications of our work for deep learning research in general. We provide further discussion in section \ref{discussionSection}, including future work (\ref{futureWorkSection}).


\section{Exploding gradients defined - the gradient scale coefficient} \label{GSCsection}

\subsection{Notation and terminology}

For the purpose of this paper, we define a neural network $f$ as a succession of layers $f_l$, $0\le l\le L$, where each layer is a function that maps a vector of fixed dimensionality to another vector of fixed but potentially different dimensionality. We assume a prediction framework, where the `prediction layer' $f_1$ is considered to output the prediction of the network and the goal is to minimize the value of the error layer $f_0$ over the network's prediction and the true label $y$, summed over some dataset $D$.

\begin{equation} \label{neuralNet}
\arg \min_{\theta} E\text{, where } E = \frac{1}{|D|}\sum_{(x,y) \in D}f_0(y,f_1(\theta_1,f_2(
\theta_2,f_3(..f_L(\theta_L,x)..))))
\end{equation}

Note that in contrast to standard notation, we denote by $f_L$ the lowest layer and by $f_0$ the highest layer of the network as we are primarily interested in the direction of gradient flow. Let the dimensionality / width of layer $l$ be $d_l$ with $d_0 = 1$ and the dimensionality of the data input $x$ be $d$.

Each layer except $f_0$ is associated with a parameter sub-vector $\theta_l$ that collectively make up the parameter vector $\theta = (\theta_1,..,\theta_L)$. This vector represents the trainable elements of the network. Depending on the type of the layer, the sub-vector might be empty. For example, a layer that computes the componentwise $\tanh$ function of the incoming vector has no trainable elements, so its parameter sub-vector is empty. We call these layers `unparametrized'. In contrast, a fully-connected linear layer has trainable weights, which are encompassed in the parameter sub-vector. We call these layers `parametrized'. 

We say a network that has layers $f_0$ through $f_L$ has `nominal depth' $L$. In contrast, we say the `compositional depth' is equal to the number of parametrized layers in the network, which approximately encapsulates what is commonly referred to as ``depth''. For example, a network composed of three linear layers, two tanh layers and a softmax layer has nominal depth 6, but compositional depth 3. We also refer to the data input $x$ as the `input layer' or the $L+1$'st layer and write $f_{L+1}$ for $x$ and $d_{L+1}$ for $d$.

Let $\mathcal{J}^l_k(\theta, x, y)$ be the Jacobian of the $l$'th layer $f_l$ with respect to the $k$'th layer $f_k$ evaluated with parameter $\theta$ at $(x,y)$, where $0\le l \le k\le L+1$. Similarly, let $\mathcal{T}^l_k(\theta, x, y)$ be the Jacobian of the $l$'th layer $f_l$ with respect to the parameter sub-vector of the $k$'th layer $\theta_k$.

Let the `quadratic expectation' $\mathbb{Q}$ of a random variable $X$ be defined as $\mathbb{Q}[X] = (\mathbb{E}[X^2])^\frac{1}{2}$, i.e. the generalization of the quadratic mean to random variables. Similarly, let the `inverse quadratic expectation' $\mathbb{Q}^{-1}$ of a random variable $X$ be defined as $\mathbb{Q}[X] = (\mathbb{E}[|X|^{-2}])^{-\frac{1}{2}}$. Further terminology, notation and conventions used only in the appendix are given in section \ref{terminologySection}.

\subsection{The colloquial notion of exploding gradients} \label{colloquialsection}

Colloquially, the exploding gradient problem is understood approximately as follows:\\

{\it When the error is backpropagated through a neural network, it may increase exponentially from layer to layer. In those cases, the gradient with respect to the parameters in lower layers may be exponentially greater than the gradient with respect to parameters in higher layers. This makes the network hard to train if it is sufficiently deep.}\\

We might take this colloquial notion to mean that if $||\mathcal{J}^l_k||$ and / or $||\mathcal{T}^l_k||$ grow exponentially in $k-l$, according to some to-be-determined norm $||.||$, the network is hard to train if it is sufficiently deep. However, this notion is insufficient because we can construct networks that can be trained successfully yet have Jacobians that grow exponentially at arbitrary rates. In a nutshell, all we have to do to construct such a network is to take an arbitrary network of desired depth that can be trained successfully and scale each layer function $f_l$ and each parameter sub-vector $\theta_l$ by $R^{-l}$ for some constant $R > 1$. During training, all we have to do to correct for this change is to scale the gradient sub-vector corresponding to each layer by $R^{-2l}$.

\begin{proposition} \label{equivalentprop}

Consider any $r > 1$ and any neural network $f$ which can be trained to some error level in a certain number of steps by some gradient-based algorithm. There exists a network $f'$ with the same nominal and compositional depth as $f$ that can also be trained to the same error level as $f$ and to make the same predictions as $f$ in the same number of steps using the same algorithm, and has exponentially growing Jacobians with rate $r$. (See section \ref{prop1proof} for details.)

\end{proposition}

Therefore, we need a definition of `exploding gradients' different from `exponentially growing Jacobians' if we hope to derive from it that training is difficult and that exploding gradients are not just a numerical issue to be overcome by gradient rescaling.

Note that all propositions and theorems are stated informally in the main body of the paper, for the purpose of readability and brevity. In the appendix in sections \ref{propProofSection} and \ref{theoProofSection} respectively, they are re-stated in rigorous terms, proofs are provided and the conditions are discussed.

\subsection{The gradient scale coefficient} \label{definitionSection}

In this section, we outline our definition of `exploding gradients' which can be used to show that training is difficult. It does not suffer from the confounding effect outlined in the previous section.

\paragraph{Definition 1.} Let the `quadratic mean norm' or `qm norm' of an $m \times n$ matrix $A$ be the quadratic mean of its singular values where the sum of squares is divided by the number of columns $n$. If $s_1$, $s_2$, .., $s_{\min(m,n)}$ are the singular values of $A$, we have:

\begin{equation*}
||A||_{qm} = \sqrt{\frac{s_1^2 + s_2^2 + .. + s_{\min(m,n)}^2}{n}}
\end{equation*}

\begin{proposition} \label{vectorEffect}
Let $A$ be an $m \times n$ matrix and $u$ a uniformly distributed, $n-dimensional$ unit length vector. Then $||A||_{qm} = \mathbb{Q}_{u}||Au||_2$. (See section \ref{propb21proof} for the proof.)
\end{proposition}

In plain language, the qm norm measures the expected impact the matrix has on the length of a vector with uniformly random orientation. The qm norm is also closely related to the $L2$ norm. We will use $||.||_2$ to denote the $L2$ norm of both vectors and matrices.

\begin{proposition} \label{L2relation}
Let $A$ be an $m \times n$ matrix. Then $||A||_{qm} = \frac{1}{\sqrt{n}}||A||_2$. (See section \ref{propb22proof} for the proof.)
\end{proposition}

\paragraph{Definition 2.} Let the {\it `gradient scale coefficient (GSC)'} for $0\le l\le k\le L+1$ be as follows:

\begin{equation*}
GSC(k,l,f,\theta,x,y) = \frac{||\mathcal{J}^l_k||_{qm}||f_k||_2}{||f_l||_2}
\end{equation*}

\paragraph{Definition 3.} We say that the network $f(\theta)$ has `exploding gradients with rate $r$ and intercept $c$' at some point $(x,y)$ if for all $k$ and $l$ we have $GSC(k,l,f,\theta,x,y) \ge cr^{k-l}$, and in particular $GSC(l,0,f,\theta,x,y) \ge cr^{l}$.\\

Of course, under this definition, any network of finite depth has exploding gradients for sufficiently small $c$ and $r$. There is no objective threshold for $c$ and $r$ beyond which exploding gradients become pathological. Informally, we will say that a network has `exploding gradients' if the GSC can be well-approximated by an exponential function, though we will restrict our attention to the GSC of the error, $GSC(l,0)$, in this paper.

The GSC combines the norm of the Jacobian with the ratio of the lengths of the forward activation vectors. In plain language, it measures the size of the gradient flowing backward relative to the size of the activations flowing forward. Equivalently, it measures the relative sensitivity of layer $l$ with respect to small random changes in layer $k$. 

\begin{proposition}
\label{relrel}
Let $u$ be a uniformly distributed, $d_k$-dimensional unit length vector. Then $GSC(k,l)$ measures the quadratic expectation $\mathbb{Q}$ of the relative size of the change in the value of $f_l$ in response to a change in $f_k$ that is a small multiple of $u$. (See section \ref{prop2proof} for details.)

\end{proposition}

What about the sensitivity of layers with respect to changes in the parameter? For fully-connected linear layers, we obtain a similar relationship. 

\begin{proposition}
\label{relrelparm}
Let $u$ be a uniformly distributed, $d_kd_{k+1}$-dimensional unit length vector. Assume $f_k$ is a fully-connected linear layer without trainable bias parameters and $\theta_k$ contains the entries of the weight matrix. Then $GSC(k,l)\frac{||\theta_k||_2||f_{k+1}||_2}{||f_k||_2\sqrt{d_{k+1}}}$ measures the quadratic expectation $\mathbb{Q}$ of the relative size of the change in the value of $f_l$ in response to a change in $\theta_k$ that is a small multiple of $u$. Further, if the weight matrix is randomly initialized, \begin{equation*}\mathbb{Q}^{-1}_{\theta_k}\frac{||\theta_k||_2||f_{k+1}||_2}{||f_k||_2\sqrt{d_{k+1}}} = 1\end{equation*} (See section \ref{prop3proof} for details.)
\end{proposition}

For reasons of space and mathematical simplicity, we focus our analysis for now on multi-layer perceptrons (MLPs) which are comprised only of fully-connected linear layers with no trainable bias parameters, and unparametrized layers. Therefore we also do not use trainable bias and variance parameters in the normalization layers. Note that using very deep MLPs with certain architectural constraints as a testbed to advance the study of exploding gradients and related concepts is a well-established practice (e.g. \cite{shattering,meanFieldResNet,pathLength}). As \cite{meanFieldExplosion}, we focus on training error rather than test error in our analysis as we do not consider the issue of generalization. While exploding gradients have important implications for generalization, this goes beyond the scope of this paper.

In section \ref{colloquialsection}, we showed that we can construct trainable networks with arbitrarily growing Jacobians by simple multiplicative rescaling of layers, parameters and gradients. Crucially, the GSC is invariant to this rescaling as it affects both the forward activations and the Jacobian equally, so the effects cancel out.

\begin{proposition}
\label{scalinginvariance}
$GSC(k,l)$ is invariant under multiplicative rescalings of the network that do not change the predictions or error values of the network. (See section \ref{prop4proof} for details.)
\end{proposition}

\section{The prevalence of exploding gradients - gradients explode despite bounded activations} \label{explosionCommon} \label{explodingsection}

In this section, we show that exploding gradients exist in a range of popular MLP architectures. Consider the decomposability of the GSC.

\begin{figure}[h]
\adjincludegraphics[width=\textwidth]{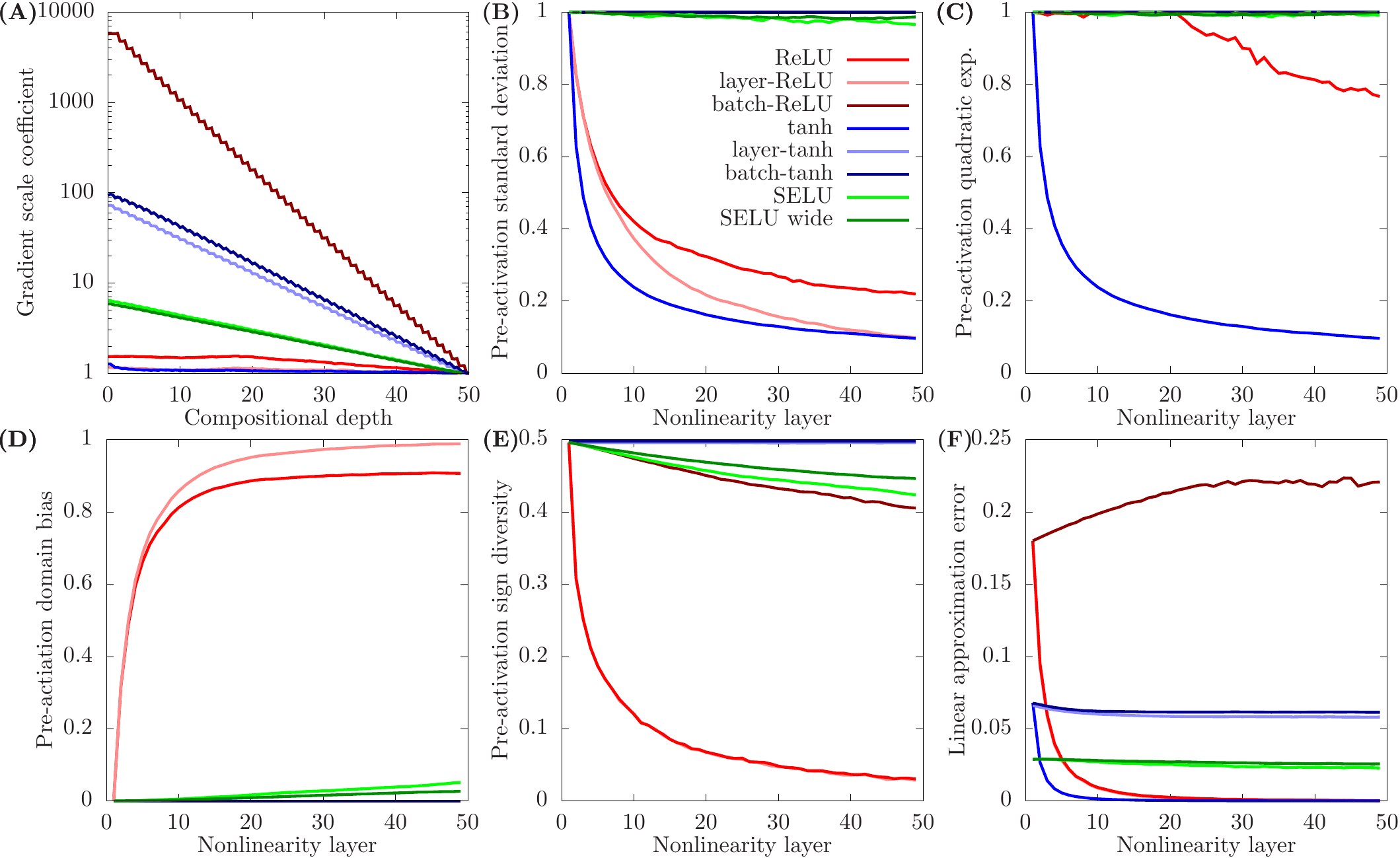}
\caption{Key metrics for architectures in their randomly initialized state evaluated on Gaussian noise. The x axis in A shows depth in terms of the number of linear layers counted from the input. The x axis in B-F counts nonlinearity layers, starting from the input. Note: The curve for layer-ReLU is shadowed by tanh in A, by ReLU in E and F and by SELU among others in C.}\label{GSCfigure}
\end{figure}

\begin{proposition}
\label{compositional}
Assuming the approximate decomposability of the qm norm of the product of Jacobians, i.e. $||\mathcal{J}^l_{l+1}\mathcal{J}^{l+1}_{l+2}..\mathcal{J}^{k-1}_{k}||_{qm} \approx ||\mathcal{J}^l_{l+1}||_{qm}||\mathcal{J}^{l+1}_{l+2}||_{qm}..||\mathcal{J}^{k-1}_{k}||_{qm}$, we have $GSC(k,l) \approx GSC(k,k-1)GSC(k-1,k-2)..GSC(l+1,l)$. (See section \ref{prop5proof} for the proof.)
\end{proposition}

This indicates that as long as the GSC of individual layers is approximately $r > 1$ and as long as the qm norm of the product of layer-wise Jacobians approximately decomposes, we have an exponential growth of $GSC(k,l)$ in $k-l$. In figure \ref{GSCfigure}A, we show $GSC(l,0)$ for seven MLP architectures. A linear layer is followed by (i) a ReLU nonlinearity (`ReLU'), (ii) layer normalization \citep{layernorm} followed by a ReLU nonlinearity (`layer-ReLU'), (iii) batch normalization plus ReLU (`batch-ReLU'), (iv) a tanh nonlinearity, (v) layer norm plus tanh (`layer-tanh'), (vi) batch norm plus tanh (`batch-tanh'), (vii) a SELU nonlinearity \citep{selu}. SELU was recently introduced by \cite{selu} and has since seen practical use (e.g. \cite{seluCNN,seluGAN,seluRL,seluAuto,seluMLP}). All networks have compositional depth 50 (i.e. 50 linear layers) and each layer has width 100. Both data input and labels are Gaussian noise and the error layer computes the dot product between the label and the prediction. The entries of the weight matrices are dawn independently from a Gaussian distributions with mean zero. Weight matrix entries for ReLU architectures are initialized with variance $\frac{2}{100}$ as suggested by \cite{heInit}, weight matrix entries for tanh architectures with variance $\frac{1}{100}$ as suggested by \cite{xavierInit}, and weight matrix entries for SELU architectures with variance $\frac{1}{100}$ as suggested by \cite{selu}. For further details about the experimental protocol, architecture composition and normalization / nonlinearity operations used, see section \ref{detailssection}. 

We find that in four architectures (batch-ReLU, layer-tanh, batch-tanh and SELU), $GSC(l,0)$ grows almost perfectly linearly in log-space. This corresponds to gradient explosion. We call those architectures `exploding architectures'. Among these architectures, a range of techniques that supposedly reduce or eliminate exploding gradients are used: careful initial scaling of weights, normalization layers, SELU nonlinearities. Adding normalization layers may even bring about exploding gradients, as observed when comparing ReLU with batch-ReLU or tanh with layer-tanh or batch-tanh. 

In light of proposition \ref{scalinginvariance}, it is not surprising that these techniques are not effective in general at combating exploding gradients as defined by the GSC, as this metric is invariant under multiplicative rescaling. Normalization layers scale the forward activations. Carefully choosing the initial scale of weights corresponds to a multiplicative scaling of the parameter. SELU nonlinearities, again, act to scale down large activations and scale up small activations. While these techniques may of course impact the GSC by changing the fundamental mathematical properties of the network, as observed when comparing e.g. ReLU and batch-ReLU, they do not reduce it simply by virtue of controlling the size of forward activations. Note that while we focus on gradient explosion as defined by the GSC in this section, the four exploding architectures would exhibit gradient explosion under any reasonable metric.

In contrast, the other three architectures (ReLU, layer-ReLU and tanh) do not exhibit exploding gradients. However, this apparent advantage comes at a cost, as we further explain in sections \ref{collapsingSection} and \ref{relationshipSection}.

All curves in figure \ref{GSCfigure}A exhibit small jitters. This is because we plotted the value of the GSC at every linear layer, every normalization layer and every nonlinearity layer in this figure and then connected the points corresponding to these values. Layers were placed equispaced on the x axis in the order they occurred in the network. Not every type of layer affects the GSC equally. In fact, we find that as gradients pass through linear layers, they tend to shrink relative to forward activations. In the exploding architectures, this is more than counterbalanced by the relative increase the gradient experiences as it passes through e.g. normalization layers. Despite these differences, it is worth noting that each individual layer used in the architectures studied has only a small impact on the GSC. This impact would be larger for either the forward activations or gradients taken by themselves. For example, passing through a ReLU layer reduces the length of both forward activation and gradient vector by $\approx \sqrt{2}$. The relative invariance of the GSC to individual layers suggests that it measures not just a superficial quantity, but a deep property of the network. This hypothesis is confirmed in the following sections.

Finally, we note that the GSC is also robust to changes in width and depth. Changing the depth has no impact on the rate of explosion of the four exploding architectures as the layer-wise GSC, i.e. $GSC(l+1,l)$, is itself independent of depth. In figure \ref{GSCfigure}A, we also show the results for the SELU architecture where each layer contains 200 neurons instead of 100 (`SELU wide'). We found that the rate of gradient explosion decreases slightly when width increases. We also studied networks with exploding architectures where the width oscillated from layer to layer. $GSC(l,0)$ still increased approximately exponentially and at a similar rate to corresponding networks with constant width. 

A summary of results can be found in table \ref{gaussianTable}.

\section{The impact of exploding gradients - exploding gradients limit depth} \label{shallowsection}

\subsection{Notation and terminology - ResNet} \label{ResNetTerminology}

We denote a ResNet as a sequence of `blocks' $f_b$, where each block is the sum of a fixed `skip block' $s_b$ and a `residual block' $\rho_b$. Residual blocks are composed of a sequence of layers. We define the optimization problem for a ResNet analogously to equation \ref{neuralNet}.

\begin{equation}\label{ResNetForm}
\arg \min_{\theta} E\text{, where } E = \frac{1}{|D|}\sum_{(x,y) \in D} f_0(y,(s_1 + \rho_1(\theta_1))\circ (s_2 + \rho_2(\theta_2)) \circ .. \circ (s_B + \rho_B(\theta_B)) \circ x)
\end{equation}

\subsection{Background: Effective depth} \label{backgroundSection}

In this section, we introduce the concept of `effective depth' as defined for ResNet architectures by \citet{effectiveDepth}. Let's assume for the sake of simplicity that the dimensionality of each block $f_b$ in a ResNet is identical. In that case, the skip block is generally chosen to be the identity function. Then, writing the identity matrix as $I$, we have

\begin{equation*} 
\frac{df_0}{dx} = \frac{df_0}{df_1}(I + \frac{d\rho_1}{df_2})(I + \frac{d\rho_2}{df_3}) .. (I + \frac{d\rho_{B-1}}{df_B})(I + \frac{d\rho_B}{dx})
\end{equation*} 

Multiplying out, this becomes the sum of $2^B$ terms. Almost all of those terms are the product of approximately $\frac{B}{2}$ identity matrices and $\frac{B}{2}$ residual block Jacobians. If the operator norm of the residual block Jacobians is less than $p$ for some $p < 1$, the norm of terms decreases exponentially in the number of residual block Jacobians they contain. Let the terms in $\frac{df_0}{dx}$ containing $\beta$ or more residual block Jacobians be called `$\beta$-residual' and let $res^\beta$ be the sum of all $\beta$-residual terms. Then:

\begin{equation*}
||res^\beta||_2 \le ||\frac{df_0}{df_1}||_2\sum_{b=\beta}^B p^b \binom{B}{b}
\end{equation*}

Again, if $p < 1$, the right hand side decreases exponentially in $\beta$ for sufficiently large $\beta$, for example when $\beta > \frac{B}{2}$. So the combined size of $\beta$-residual terms is exponentially small. Therefore, \citet{effectiveDepth} argue, the full set of blocks does not jointly co-adapt during training because the information necessary for such co-adaption is contained in gradient terms that contain many or all residual block Jacobians. Only sets of blocks of size at most $\beta$ where $res^\beta$ is not negligably small co-adapt. The largest such $\beta$, multiplied by the depth of each block, is called the `effective depth' of the network. \citet{effectiveDepth} argue that a ResNet is not really as deep as it appears, but rather behaves as an ensemble of relatively shallow networks where each member in that ensemble has depth less than or equal to the effective depth of the ResNet. This argument is bolstered by the success of the stochastic depth \citep{stochasticDepth} training technique, where random sets of residual blocks are deleted for each mini-batch update.

\citet{effectiveDepth} introduced the concept of effective depth  somewhat informally. We give our formal definition in section \ref{depthDetailsSection}. There, we also provide a more detailed discussion of the concept and point out limitations.

\subsection{The residual trick}

Now we make a crucial observation. Any neural network can be expressed in the framework of equation \ref{ResNetForm}. We can simply cast each layer as a block, choose arbitrary $s_b$ and define $\rho_b(\theta_b) := f_b(\theta_b) - s_b$. Specifically, if we train a network $f$ from some fixed initial parameter $\theta^{(0)}$, we can set $s_b := f_b(\theta^{(0)}_b)$ and thus $\rho_b(\theta_b) := f_b(\theta_b) - f_b(\theta^{(0)}_b)$. Then training begins with all the $\rho_b$ being zero functions. Therefore, all analysis devised for ResNet that relies on the small size of the residual block Jacobians can then be brought to bear on arbitrary networks. We term this the {\it `residual trick'}. Indeed, the analysis by \cite{effectiveDepth} does not rely on the network having skip connections in the computational sense, but only on the mathematical framework of equation \ref{ResNetForm}. Therefore, as long as the operator norms of $\frac{df_b(\theta_b)}{df_{b+1}} - \frac{df_b(\theta^{(0)}_b)}{df_{b+1}}$ are small, $f$ is effectively shallow.

\subsection{Notation and terminology - residual network} \label{residualTerminology}

 From now on, we will make a distinction between the terms `ResNet' and `residual network'. `ResNet'
 will be used to refer to networks that have an architecture as in \cite{resNet} that uses skip connections as outlined in section \ref{ResNetTerminology}. In contrast, networks without skip connections will be referred to as `vanilla networks'. Define the fixed `initial function' $i_l$ as $f_l(\theta^{(0)}_l)$ and the `residual function' $r_l$ as $f_l(\theta_l) - f_l(\theta^{(0)}_l)$. Then we refer to a `residual network' as an arbitrary network expressed as
 
\begin{equation}\label{residualForm}
f_0(y,(i_1 + r_1(\theta_1))\circ (i_2 + r_2(\theta_2)) \circ .. \circ (i_L + r_L(\theta_L)) \circ x)
\end{equation}

The gradient becomes

\begin{equation*} 
\frac{df_0}{dx} = \frac{df_0}{df_1}(\frac{di_1}{df_2} + \frac{dr_1}{df_2})(\frac{di_2}{df_3} + \frac{dr_2}{df_3}) .. (\frac{di_{L-1}}{df_L} + \frac{dr_{L-1}}{df_L})(\frac{di_L}{dx} + \frac{dr_L}{dx})
\end{equation*} 

After multiplying out, a term containing $\lambda$ or more residual Jacobians is called `$\lambda$-residual'.

\subsection{Theoretical analysis} \label{analysisSection}

In this section, we will show that an exploding gradient as defined by the GSC causes the effective training time of deep MLPs to be exponential in depth and thus limits the effective depth that can be achieved. 

The proof is based on the insight that the relative size of a gradient-based update $\Delta \theta_l$ on $\theta_l$ is bounded by the inverse of the GSC if that update is to be useful. The basic assumption underlying gradient-based optimization is that the function optimized is locally well-approximated by a linear function as indicated by the gradient. Any update made based on a local gradient computation must be small enough so that the updated value lies in the region around the original value where the linear approximation is sufficiently accurate. Let's assume we apply a random update to $\theta_l$ with relative size $\frac{1}{GSC(l,0)}$, i.e. $\frac{||\Delta \theta_l||_2}{||\theta_l||_2} = \frac{1}{GSC(l,0)} $. Then under the local linear approximation, according to proposition \ref{relrelparm}, this would change the value of $f_0$ approximately by a value with quadratic expectation $f_0$. Hence, with significant probability, the error would become negative. This is not reflective of the true behavior of $f$ in response to changes in $\theta_l$ of this magnitude. Since $f$ is even more sensitive to updates in the direction of the gradient than it is to random updates, useful gradient-based updates are even more likely to be bounded in relative magnitude by $\frac{1}{GSC(l,0)}$.

\begin{figure}
\adjincludegraphics[width=0.95\textwidth]{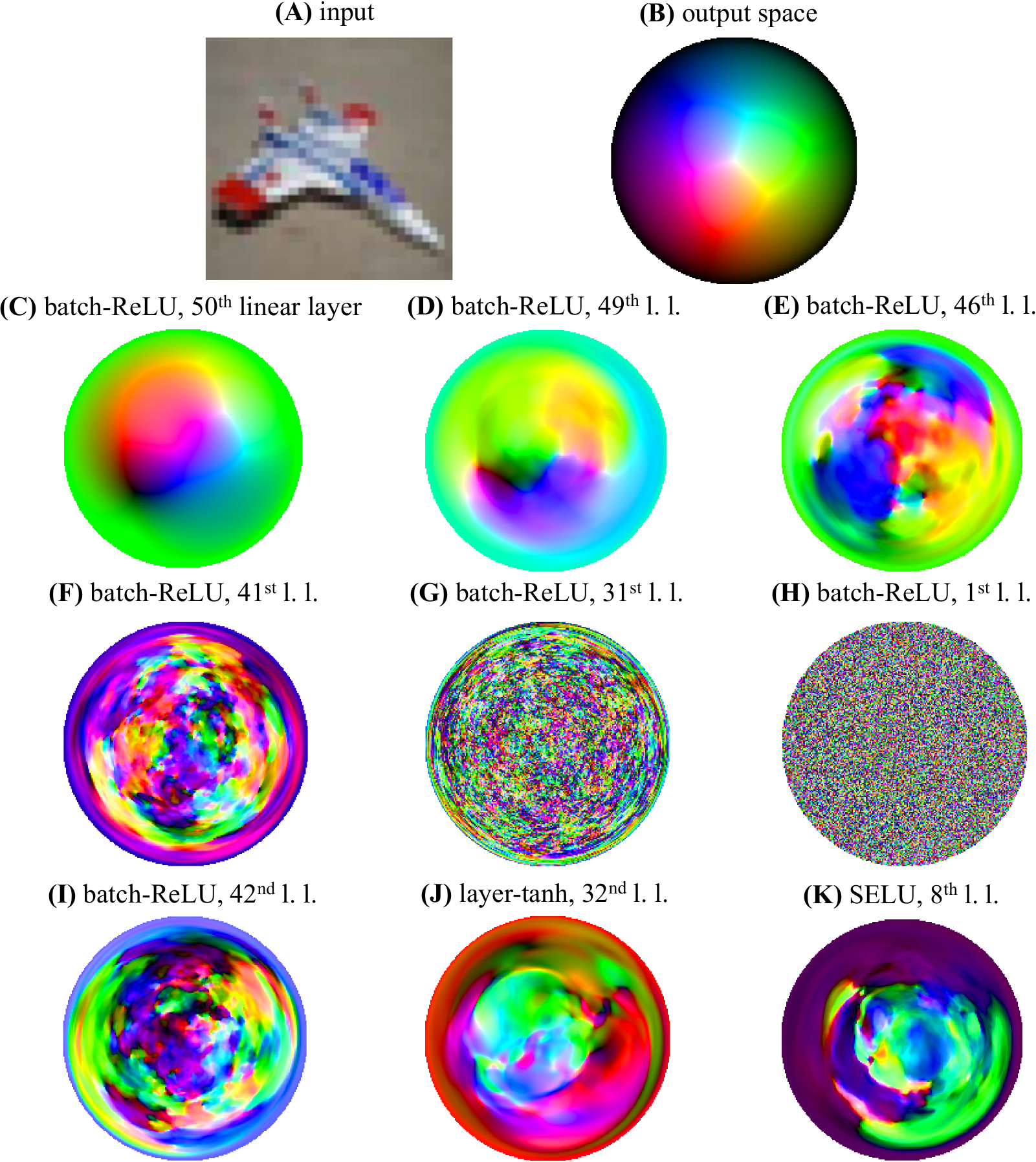}
\caption{Illustrations of networks of different architectures as functions of the parameter in a single linear layer. For each network architecture as indicated under (C-K) with 50 linear layers, three random weight configurations are chosen that differ only at a single linear layer as indicated. For each location on the sphere centered on the origin containing those three configurations, the input shown in A from the CIFAR10 dataset is propagated through the network with weights indicated by that location. The length of the 3-dimensional output of the prediction layer is then normalized. Each location on the sphere is colored according to this output as shown in B. Weight configurations where the input is assigned class 1/2/3 are shown in red/green/blue respectively. Discs B through K are azimuthal projections. See section \ref{detailssection} for details.} \label{im_expl}
\end{figure}

Figure \ref{im_expl} illustrates this phenomenon. There, we depict the value of the output of the prediction layer of various network architectures given the input in \ref{im_expl}A over a 2-dimensional subspace of the parameter space of a single linear layer. We set $d_1 = 3$. Different colors correspond to different regions of the output space, which is shown in \ref{im_expl}B. Figures \ref{im_expl}C-H illustrate that the complexity of a batch-ReLU network as a function of the parameter $\theta_l$ grows exponentially with $l$ as indicated by figure \ref{GSCfigure}A. As this complexity grows, the region of the parameter space that is well-approximated by the local gradient vanishes accordingly. Figures \ref{im_expl}I-K compares 3 linear layers in different architectures with comparable $GSC(l,0)$ values. We can observe that while the visual properties of $f$ vary, the complexity is comparable.

In a nutshell, if $\frac{1}{GSC(l,0)}$ decreases exponentially in $l$, so must the relative size of updates. So for a residual function to reach a certain size relative to the corresponding initial function, an exponential number of updates is required. But to reach a certain effective depth, a certain magnitude of $\lambda$-residual terms is required and thus a certain magnitude of residual functions, and thus exponentially many updates.

\begin{theorem} \label{depthTheorem}
Under certain conditions, if an MLP has exploding gradients with explosion rate $r$ and intercept $c$ on some dataset, then there exists a constant $c'$ such that training this MLP with a gradient-based algorithm to have effective depth $\lambda$ takes at least $c'cr^\frac{\lambda}{4}$ updates. (See section \ref{theo1proof} for details.)
\end{theorem}


Importantly, this lower bound on the number of updates required to reach a certain effective depth is independent of the nominal depth of the network. While the constant $c'$ depends on some constants that arise in the conditions of the theorem, as long as those constants do not change when depth is increased, neither does the lower bound on the number of updates.

\begin{corollary}
In the scenario of theorem \ref{depthTheorem}, if the number of updates to convergence is bounded, so is effective depth.
\end{corollary}

Here we simply state that if we reach convergence after a certain number of updates, but theorem \ref{depthTheorem} indicates that more would be required to attain a greater effective depth, then that greater effective depth is unreachable with that algorithm.

\subsection{Experiments} \label{depthexperimentssection}

To practically validate our theory of limited effective depth, we train our four exploding architectures (batch-ReLU, layer-tanh, batch-tanh and SELU) on CIFAR10. All networks studied have a compositional depth of 51, i.e. there are 51 linear layers. The width of each layer is 100, except for the input, prediction and error layers. Full experimental details can be found in section \ref{detailssection}. 

First, we determined the approximate best step size for SGD for each individual linear layer. We started by pre-training the highest layers of each network with a small uniform step size until the training classification error was below 85\%, but at most for 10 epochs. Then, for each linear layer, we trained only that layer for 1 epoch with various step sizes while freezing the other layers. The step size that achieved the lowest training classification error after that epoch was selected. Note that we only considered step sizes that induce relative update sizes of 0.1 or less, because larger updates often cause weight instability. The full algorithm for step size selection and a justification is given in section \ref{stepSizeSection}.

\begin{figure}
\adjincludegraphics[width=0.96\textwidth]{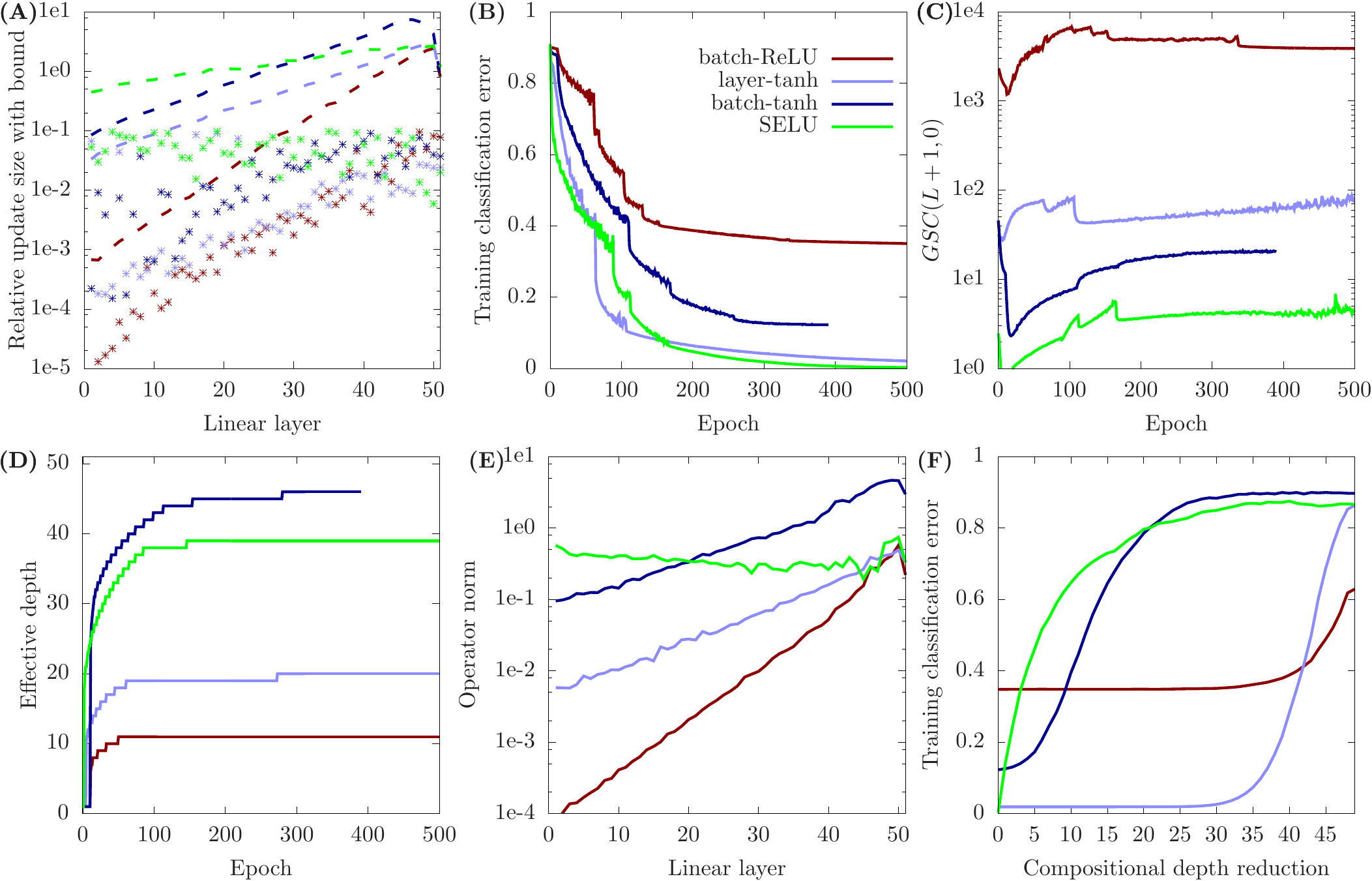}
\caption{Key metrics for exploding architectures trained on CIFAR10. See main text for explanation.} \label{dyna}
\end{figure}

In figure \ref{dyna}A, we show the relative update size induced on each linear layer by what was selected to be the best step size as well as $\frac{1}{GSC(l,0)}$ as a dashed line. In section \ref{analysisSection}, we argued that $\frac{1}{GSC(l,0)}$ is an upper bound for the relative size of a useful update. We find that this bound holds and is conservative except for a small number of outliers. Even though our algorithm for determining the best step size for each layer gives noisy results, there is a clear trend that lower layers require relatively smaller updates, and that this effect is more pronounced if the gradient explodes with a larger rate. Therefore the foundational assumption underlying theorem \ref{depthTheorem} holds.

We then smoothed these best step size estimates and trained each network for 500 epochs with those smoothed estimates. Periodically, we scaled all step sizes jointly by $\frac{1}{3}$. In figure \ref{dyna}B, we show the training classification error of each architecture. Final error values are shown in table \ref{CIFARtable}. There is a trend that architectures with less gradient explosion attain a lower final error. In fact, the final error values are ordered according to the value of $GSC(L+1,0)$ in the initialized state. Note that, of course, all these error values are still much higher than the state of the art on CIFAR10. This is not a shortcoming of our analysis however, as the goal of this section is to study and understand pathological architectures rather than find optimal ones. Those architectures, by definition, attain high errors.

In figure \ref{dyna}C, we show $GSC(L+1,0)$ as training progresses. During the initial pre-training phase, this value drops significantly but later regains or even exceeds its original value. In figure \ref{dyna}A, the dashed line indicates the inverse of $GSC(l,0)$ for each $l$ after pre-training. We find that the GSC actually falls below 1 as the gradient passes through the pre-trained layers, but then resumes explosion once it reached the layers that were not pre-trained. We find this behavior surprising and unexpected. We conclude that nonstandard training procedures can have a significant impact on the GSC but that there is no evidence that when all layers are trained jointly, which is the norm, the GSC either significantly increases or decreases during training.

We then went on to measure the effective depth of each network. We devised a conservative, computationally tractable estimate of the cumulative size of updates that stem from $\lambda$-residual terms. See section \ref{depthComputationSection} for details. The effective depth depicted in figure \ref{dyna}D is the largest value of $\lambda$ such that this estimate has a length exceeding $10^{-6}$. As expected, none of the architectures reach an effective depth equal to their compositional depth, and there is a trend that architectures that use relatively smaller updates achieve a lower effective depth. It is worth noting that the effective depth increases most sharply at the beginning of training. Once all step sizes have been multiplied by $\frac{1}{3}$ several times, effective depth no longer changes significantly while the error, on the other hand, is still going down. This suggests that, somewhat surprisingly, high-order co-adaption of layers takes place towards the beginning of training and that as the step size is reduced, layers are fine-tuned relatively independently of each other.

SELU and especially tanh-batch reach an effective depth close to their compositional depth according to our estimate. In figure \ref{dyna}E, we show the operator norm of the residual weight matrices after training. All architectures except SELU, which has a $GSC(L+1,0)$ close to 1 after pre-training, show a clear downward trend in the direction away from the error layer. If this trend were to continue for networks that have a much greater compositional depth, then those networks would not achieve an effective depth significantly greater than our 51-linear layer networks.

\cite{effectiveDepth} argue that a limited effective depth indicates a lack of high-order co-adaptation. We wanted to verify that our networks, especially layer-tanh and batch-ReLU, indeed lack these high-order co-adaptations by using a strategy independent of the concept of effective depth to measure this effect. We used Taylor expansions to do this. Specifically, we replaced the bottom $k$ layers of the fully-trained networks by their first-order Taylor expansion around the initial functions. See section \ref{taylorDetailsSection} for how this is done. This reduces the compositional depth of the network by $k-2$. In figure \ref{dyna}F, we show the training classification error in response to compositional depth reduction. We find that the compositional depth of layer-tanh and batch-ReLU can be reduced enormously without suffering a significant increase in error. In fact, the resulting layer-tanh network of compositional depth 15 greatly outperforms the original batch-tanh and batch-ReLU networks. This confirms that these networks lack high-order co-adaptations. Note that cutting the depth by using the Taylor expansion not only eliminates high-order co-adaptions among layers, but also co-adaptions of groups of 3 or more layers among the bottom $k$ layers. Hence, we expect the increase in error induced by removing only high-order co-adaptions to be even lower than what is shown in figure \ref{dyna}F. Unfortunately, this cannot be tractably computed.

Finally, we trained each of the exploding architectures by using only a single step size for each layer that was determined by grid search, instead of custom layer-wise step sizes. As expected, the final error was higher. The results are found in table \ref{CIFARtable}.

\paragraph{Summary} For the first time, we established a direct link between exploding gradients and severe training difficulties for general gradient-based training algorithms. These difficulties arise in MLPs composed of popular layer types, even if those MLPs utilize techniques that are believed to combat exploding gradients by stabilizing forward activations. The gradient scale coefficient not only underpins this analysis, but is largely invariant to the confounders of network scaling (section \ref{definitionSection}), layer width and individual layers (section \ref{explodingsection}). Therefore we propose that the GSC can standardize research on gradient pathology.

\subsection{A note on batch normalization and other sources of noise} \label{batchNormSection}

We used minibatches of size 1000 to train all architectures except batch-ReLU, for which we conducted full-batch training. When minibatches were used on batch-ReLU, the training classification error stayed above 89\% throughout training. (Random guessing achieves a 90\% error.) In essence, no learning took place. This is because of the pathological interplay between exploding gradients and the noise inherent in batch normalization. Under batch normalization, the activations at a neuron are normalized by their mean and standard deviation. These values are estimated using the current batch. Hence, if a minibatch has size $b$, we expect the noise induced by this process to have relative size $\approx \frac{1}{\sqrt{b}}$. But we know that according to proposition \ref{relrel}, under the local linear approximation, this noise leads to a change in the error layer of relative size $\approx \frac{GSC}{\sqrt{b}}$. Hence, if the GSC between the error layer and the first batch normalization layer is larger than ${\sqrt{b}}$, learning should be seriously impaired. For the batch-ReLU architecture, this condition was satisfied and consequently, the architecture was untrainable using minibatches. Ironically, the gradient explosion that renders the noise pathological was introduced in the first place by adding batch normalization layers. Note that techniques exist to reduce the dependence of batch normalization on the current minibatch, such as using running averages \citep{renorm}. Other prominent techniques that induce noise and thus can cause problems in conjunction with large gradients are dropout \citep{dropout}, stochastic nonlinearities (e.g. \cite{stochasticNonlinearity}) and network quantization (e.g. \cite{quantization}).

\section{The origins of exploding gradients - quadratic vs geometric means} \label{widerViewsection}

Why do exploding gradients occur? As mentioned in section \ref{explodingsection}, gradients explode with rate $r > 1$ as long as we have (i) $GSC(k,l) \approx GSC(l+1,l)GSC(l+2,l+1)..GSC(k,k-1)$ and (ii) $GSC(l+1,l) \approx r$ for all $k$ and $l$. Our results from figure \ref{GSCfigure}A suggest that (i) holds in practical networks. Indeed, we can justify (i) theoretically by viewing the parameters of linear layers as random variables.

\begin{theorem} \label{explodeTheorem}
Under certain conditions, for any neural network $f$ composed of layer functions $f_l$ that are parametrized by randomly initialized $\theta_l$, \begin{equation*}\sqrt{\frac{d_k}{d_k-1}}\mathbb{Q}_{\theta} GSC(k,l) = \prod_{l'=l+1}^{k} \sqrt{\frac{d_{l'}}{d_{l'}-1}} \mathbb{Q}_\theta GSC(l',l'-1)\end{equation*}. (See section \ref{theo2proof} for details.)
\end{theorem}

Let's turn to (ii). The common perception of the exploding gradient problem is that it lies on a continuum with the vanishing gradient problem and that all we need to do to avoid both is to hit a sweet spot by avoiding design mistakes. According to this viewpoint, building a network with $GSC(l+1,l) \approx 1$ should not be difficult.

\paragraph{Definition 4.} Assume $X$ is a random variable with a real-valued probability density function $p$ on $\mathbb{R}^d$. Then we define its `exponential entropy' $E(X)$ as $\mathbb{E}_X p(X)^{-\frac{1}{d}}$.\\

In comparison, entropy $H(X)$ is defined as $\mathbb{E}_X-\log p(X)$.

\begin{theorem} \label{bigGradTheorem}
Let $X$ be a random variable with a real-valued probability density function $p$ on $\mathbb{R}^d$ and let $f$ be an endomorphism on $\mathbb{R}^d$. Let $\sigma_{|s|}$ be the standard deviation of the absolute singular values of $\mathcal{J}^f_X$ and let $\mu_{|s|}$ be the mean of the absolute singular values of $\mathcal{J}^f_X$. Assume $\sigma_{|s|} \ge \sqrt{\epsilon(\epsilon+2\mu_{|s|})}$ with probability $\delta$. Then \begin{equation*}\mathbb{E}_X||\mathcal{J}^f_X||_{qm} \ge \epsilon\delta + e^{\frac{1}{d}(H(f(X)) - H(X))}\end{equation*}

Further, assume that the value of $p$ is independent of $\mathcal{J}^f_X$. Then \begin{equation*}\mathbb{E}_X||\mathcal{J}^f_X||_{qm} \ge \epsilon\delta + \frac{E(f(X))}{E(X)}\end{equation*}

(See section \ref{theo2p5proof} for details.)  
\end{theorem}

\begin{figure}
\includegraphics[width=\textwidth]{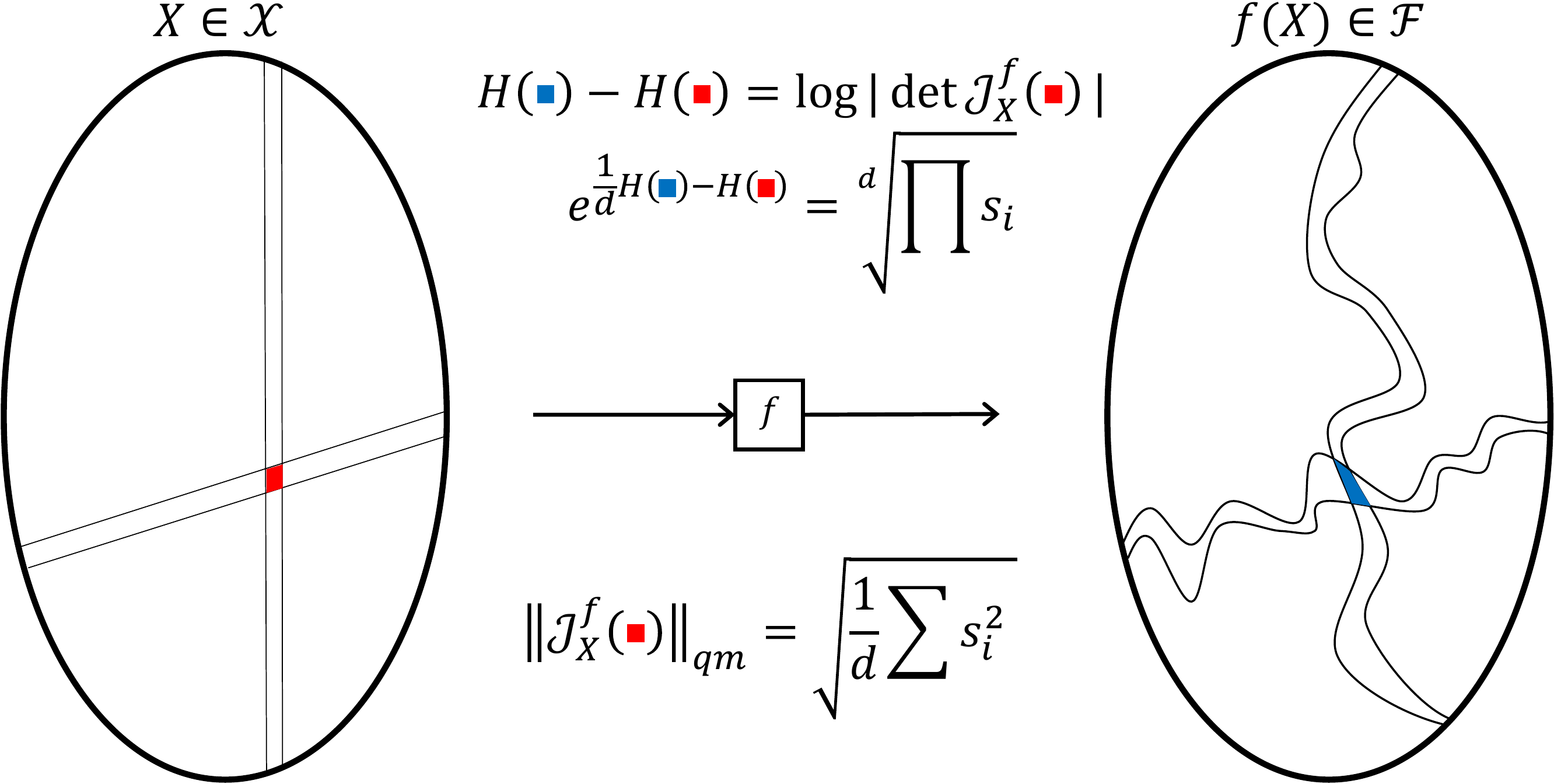}
\caption{Illustration of theorem \ref{bigGradTheorem}. See main text for details.} \label{im_origin}
\end{figure}

\begin{corollary} \label{bigGradCorollary}
Let $f_{l+1}(f_{l+2}(..f_L(x)..))$ and $f_l$ fulfill the same conditions as $X$ and $f$ above. Also let $||f_l||_2 = c_l$ and $||f_{l+1}||_2 = c_{l+1}$ be fixed. Then \begin{equation*}\mathbb{E}GSC(l+1,l) \ge \frac{c_{l+1}}{c_l}\epsilon\delta + e^{\frac{1}{d}((H(f_l) - \log c_l)  - (H(f_{l+1}) - \log c_{l+1}))}\end{equation*}

And under the further assumption from above we have \footnote{We present results in terms of $E$ and $H$ because $H$ requires less assumptions but $E$ provides the tighter bound.} \begin{equation*}\mathbb{E}GSC(l+1,l) \ge \frac{c_{l+1}}{c_l}\epsilon\delta + \frac{\frac{E(f_l)}{c_l}}{\frac{E(f_{l+1})}{c_{l+1}}}\end{equation*}

\end{corollary}

The assumption of layer functions having a fixed output length is fulfilled approximately in sufficiently wide networks due to the law of large numbers. 

Theorem \ref{bigGradTheorem} suggests that layer-wise GSC's larger than 1 occur even in networks where forward activations are stable and cannot be eliminated via simple design choices, unless (I) information loss is present or (II) Jacobians are constrained. It turns out that these are exactly the strategies employed by popular architectures ReLU and ResNet to avoid exploding gradients, as we will show in the next two sections respectively. Both strategies incur drawbacks in practice. While we have not observed a vanishing gradient in any architecture we studied in this paper, we conjecture that an architecture built on popular design principles that exhibits them would suffer those drawbacks to an even larger degree. 

The mechanism underlying theorem \ref{bigGradTheorem} is that information propagation is governed by the geometric mean of absolute singular values of the Jacobian, whereas the GSC is governed by the quadratic mean of the absolute singular values. We illustrate this in figure \ref{im_origin}. Say a random input variable $X$ with domain $\mathcal{X}$ is mapped by a nonlinear function $f$ onto a domain $\mathcal{F}$ and say the small red patch is mapped onto the small blue patch. The difference in entropy between the patches is equal to the logarithm of the absolute determinant of the Jacobian at the red patch. This quantity is related to the geometric mean of the absolute singular values. Conversely, the GSC at the red patch is based on the qm norm of the Jacobian, which is itself the quadratic mean of the absolute singular values. As the quadratic mean is larger than the geometric mean, with the size of the difference governed by how ``spread out'' the absolute singular values are, we obtain the result of the theorem. 

In current practice, avoiding exploding gradients does not seem to be a matter of simply avoiding design mistakes, but involves tradeoffs with other potentially harmful effects. As our theoretical analysis in this section does rely on several conditions such as a constant width and the presence of a probability density function, we leave open the possibility of designing novel architectures that avoid exploding gradients by exploiting these conditions.

\section{Exploding gradient tradeoffs - the collapsing domain problem} \label{collapsingSection}

In the previous section, we showed how gradients explode if the entropy or exponential entropy is perserved relative to the scale of forward activations. This suggests that we can avoid exploding gradients via a sufficiently large entropy reduction. This corresponds to a contraction of the latent representations of different datapoints, a {\it collapsing domain}. 

Consider a contraction of the domain around a single point. If we shrink the co-domain of some layer function $f_l$ by a factor $c$, we reduce the eigenvalues of the Jacobian and hence its qm norm by $c$. If we also ensure that the length of the output stays the same, the GSC is also reduced by $c$. Similarly, inflating the co-domain would cause the qm norm to increase. A contraction around a single point would cause the activation values at each individual neuron to be biased. We call this {\it domain bias}.

This is precisely what we find. Returning to figure \ref{GSCfigure}, we now turn our attention to graphs B through F. In B, we plot the standard deviation of the activation values in the layers before each nonlinearity layer (`pre-activations'). Each standard deviation is taken at an individual neuron and across datapoints. The standard deviations of neurons in the same layer are combined by taking their quadratic mean. In C, we plot the quadratic expectation of the pre-activations. The two quantities diverge significantly for 2 architectures: ReLU and layer-ReLU. This divergence implies that activation values become more and more clustered away from zero with increasing depth, which implies domain bias. In D, we plot the fraction of the signal explained by the bias (bias squared divided by quadratic expectation squared). This value increases significantly only for ReLU and layer-ReLU. Hence, we term those two architectures `biased architectures'.

But why can domain bias be a problem? There are at least two reasons.

\paragraph{Domain bias can cause pseudo-linearity}

\begin{figure}
\includegraphics[width=\textwidth]{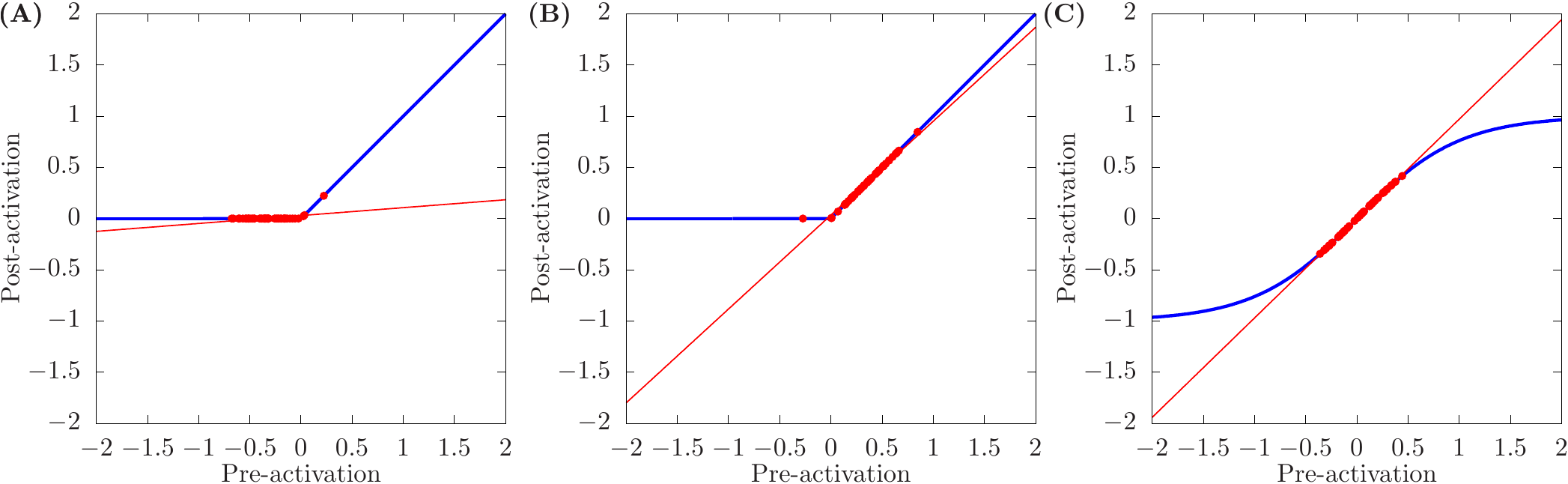}
\caption{The phenomenon of pseudo-linearity in ReLU and tanh nonlinearities. The nonlinearity function is shown in blue, the nonlinearity applied to 50 individual pre-activations drawn from a Gaussian with mean $\mu$ and standard deviation $\sigma$ are shown as red dots. The closest linear fit to the 50 post-activations is shown as a red line, and it approximates these post-activations very closely. A: ReLU, $\mu=-0.3$, $\sigma=0.2$. B: ReLU, $\mu=0.3$, $\sigma=0.2$. C: tanh, $\mu=0$, $\sigma=0.2$} \label{pseudoLin}
\end{figure}

If the pre-activations that are fed into a nonlinearity are sufficiently similar, the nonlinearity can be well-approximated by a linear function. In an architecture employing ReLU nonlinearities, if either all or most pre-activations are positive or all or most pre-activations are negative, the nonlinearity can be well-approximated by a linear function. If all or most pre-activations are negative, ReLU can be approximated by the zero function (figure \ref{pseudoLin}A). If all or most pre-activations are positive, ReLU can be approximated by the identity function (figure \ref{pseudoLin}B). But if nonlinearity layers become well-approximated by linear layers, the entire network becomes equivalent to a linear network. We say the network becomes {\it `pseudo-linear'}. Of course, linear networks of any depth have the representational capacity of a linear network of depth 1 and are unable to model nonlinear functions. Hence, a network that is pseudo-linear beyond compositional depth $k$ approximately has the representational capacity of a compositional depth $k+1$  network.

In figure \ref{GSCfigure}E, we plot the proportion of the pre-activations at each neuron that are positive or negative, whichever is smaller for that neuron. Values are averaged over each layer. We call this metric `sign diversity'. For ReLU and layer-ReLU, sign diversity decreases rapidly. Because of the properties of ReLU discussed above, this implies pseudo-linearity. Finally, in figure \ref{GSCfigure}F we plot the error incurred by replacing each neuron in a nonlinearity layer by its respective best fit linear function, measured as one minus the ratio of the signal power (squared quadratic expectation) of the approximated post-activations over the signal power of the true post-activation. We find that in the two biased architectures, pseudo-linearity takes hold substantially after layer 10 and completely after layer 25.

\paragraph{Domain bias can mask exploding gradients}

In theorem \ref{depthTheorem}, we used the fact that the output of the error layer of the network was positive to bound the size of a useful gradient-based update. In other words, we used the fact that the domain of the error layer is bounded. However, domain bias causes not just a reduction of the size of the domain of the error layer, but of all intermediate layers. This should ultimately have the same effect on the largest useful update size as exploding gradients, that is to reduce them and thus cause a low effective depth.

\begin{figure}
\includegraphics[width=\textwidth]{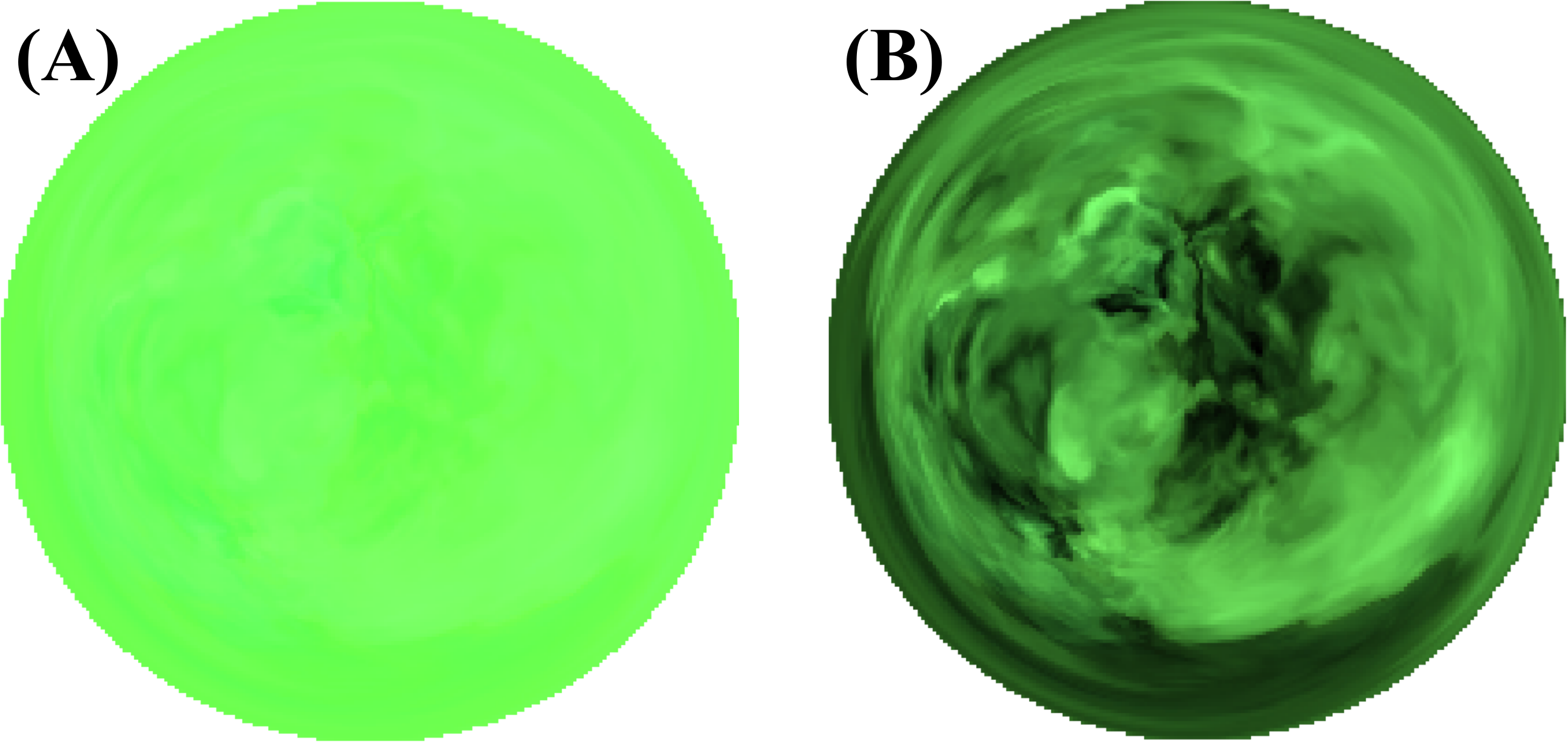}
\caption{Illustration of a 50-layer ReLU network as a function of the parameter of the first linear layer, using the same methodology as figure \ref{im_expl}. Figure B is a contrast-heightened version of figure A. See section \ref{detailssection} for details.} \label{im_collapse}
\end{figure}

We illustrate this effect in figure \ref{im_collapse}. In figure \ref{im_collapse}A, we depict the output of a 50-layer ReLU network over a 2-dimensional subspace of the parameter space of the first linear layer, using the same methodology as in figure \ref{im_expl}. Domain bias causes the output to be restricted to a narrow range. While at first glance the function looks simple, heightening the contrast (\ref{im_collapse}B) reveals that there are oscillations of small amplitude that were not present in exploding architectures. Because any color shift is confined to a small amplitude, we necessarily obtain oscillations. Those oscillations then lead to local gradient information being uninformative, just as with exploding architectures. 

\begin{figure}[p]
\adjincludegraphics[width=\textwidth]{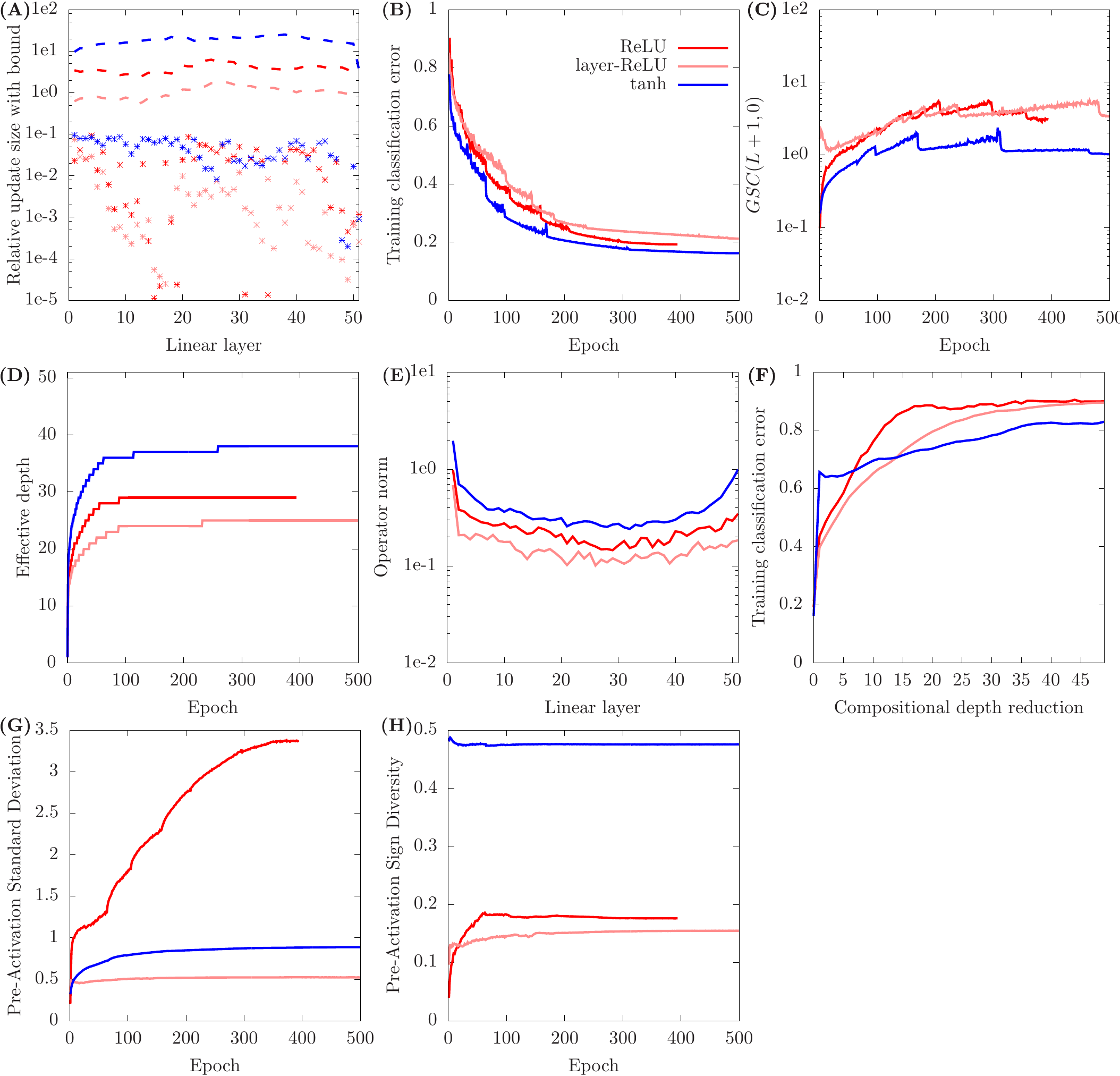}
\caption{Key metrics for architectures that avoid exploding gradients trained on CIFAR10. The top left graph shows the estimated optimal relative update size in each layer according to the algorithm described in section \ref{stepSizeSection}. Remaining graphs show results obtained from training with a single step size as this achieved lower error than training with layer-wise step sizes (see table \ref{CIFARtable}). The top two rows are equivalent to graphs in figure \ref{dyna}. The bottom row shows pre-activation standard deviation and pre-activation sign diversity (see section \ref{gaussianProtocol} for definition) of the highest nonlinearity layer as training progresses.} \label{dynaCollapse}
\end{figure}

In table \ref{CIFARtable}, we show the final error values achieved by training ReLU and layer-ReLU on CIFAR10. The errors are substantially higher than those achieved by the exploding architectures, except for batch-ReLU. Also, training with layer-wise step sizes did not help compared to training with a single step size. In figure \ref{dynaCollapse}A, we show the estimated best relative update size for each layer. This time, there is no downward trend towards lower layers, which is likely why training with a single step size is ``sufficient''. As conjectured, the difference between the $\frac{1}{GSC}$ bound and the empirical estimate is much larger for the biased architectures than it is for the exploding architectures (see figure \ref{dyna}A), indicating that both suffer from reduced useful update sizes. In figure \ref{dynaCollapse}D, we find the effective depth reached by ReLU and layer-ReLU is significantly lower than the compositional depth of the network and is comparable to that of architectures with exploding gradients (see figure \ref{dyna}D).

In figure \ref{dynaCollapse}G and H, we plot the pre-activation standard deviation and sign diversity at the highest nonlinearity layer throughout training. Interestingly, sign diversity increases significantly early in training. The networks become less linear through training.

\paragraph{Summary} In neural network design, there is an inherent tension between avoiding exploding gradients and collapsing domains. Avoiding one effect can bring about or exacerbate the other. Both effects are capable of severely hampering training. This tension is brought about by the discrepancy of the geometric and quadratic mean of the singular values of layer-wise Jacobians and is a foundational reason for the difficulty in constructing very deep trainable networks. 

Many open questions remain. Is it possible to measure or at least approximate the entropy of latent representations? What about latent representations that have varying dimensionality or lack a probability density function? In what ways other than domain bias can collapsing domains manifest and how would those manifestations hamper training? An example of such a manifestation would be a clustering of latent representations around a small number of principle components as observed in Gaussian initialized linear networks \citep{orthogonalInitialization,eigenspectrumGram}.

\section{Exploding gradient solutions - ResNet and the orthogonal initial state} \label{ResNetsection}

ResNet and related architectures that utilize skip connections have been very successful recently. One reason for this is that they can be successfully trained to much greater depths than corresponding vanilla networks. In this section, we show how skip connections are able to greatly reduce the GSC and thus largely circumvent the exploding gradient problem. Please refer back to section \ref{ResNetTerminology} for the notation and terminology we employ for ResNets.

\paragraph{Definition 4.} We say a function $f_b$ is `$k$-diluted' with respect to a random vector $v$, a matrix $S_b$ and a function $\rho_b$ if $f_b(v) = S_bv + \rho_b(v)$ and $\frac{\mathbb{Q}_v||S_bv||_{2}}{\mathbb{Q}_v||\rho_b(v)||_{2}} = k$.\\

$k$-dilution expresses the idea that the kinds of functions that a block $f_b$ represents are of a certain form if $s_b$ is restricted to matrix multiplication. (Note that the identity function can be viewed as matrix multiplication with the identity matrix.) The larger the value of $k$, the more $\rho_b$ is ``diluted'' by a linear function, bringing $f_b$ itself closer and closer to a linear function. $v$ represents the incoming forward activations to the block $f_b$. 

\begin{theorem} \label{dilutionTheorem}
If a block $\rho_b(v)$ would cause the $GSC$ to grow with expected rate $r$, $k$-diluting $\rho_b(v)$ with an uncorrelated linear transformation $S_bv$ reduces this rate to $1 + \frac{r-1}{k^2+1} + O((r-1)^2)$. (See section \ref{theo3proof} for details.)
\end{theorem}

This reveals the reason why ResNet circumvents the exploding gradient problem. $k$-diluting $\rho_b$ does not just reduce the growth of the GSC by $k$, but by $k^2+1$. Therefore what appears to be a relatively mild reduction in representational capacity achieves, surprisingly, a relatively large amount of gradient reduction, and therefore ResNet can be trained successfully to ``unreasonably'' great depths for general architectures.

\begin{figure}[p]
\includegraphics[width=\textwidth]{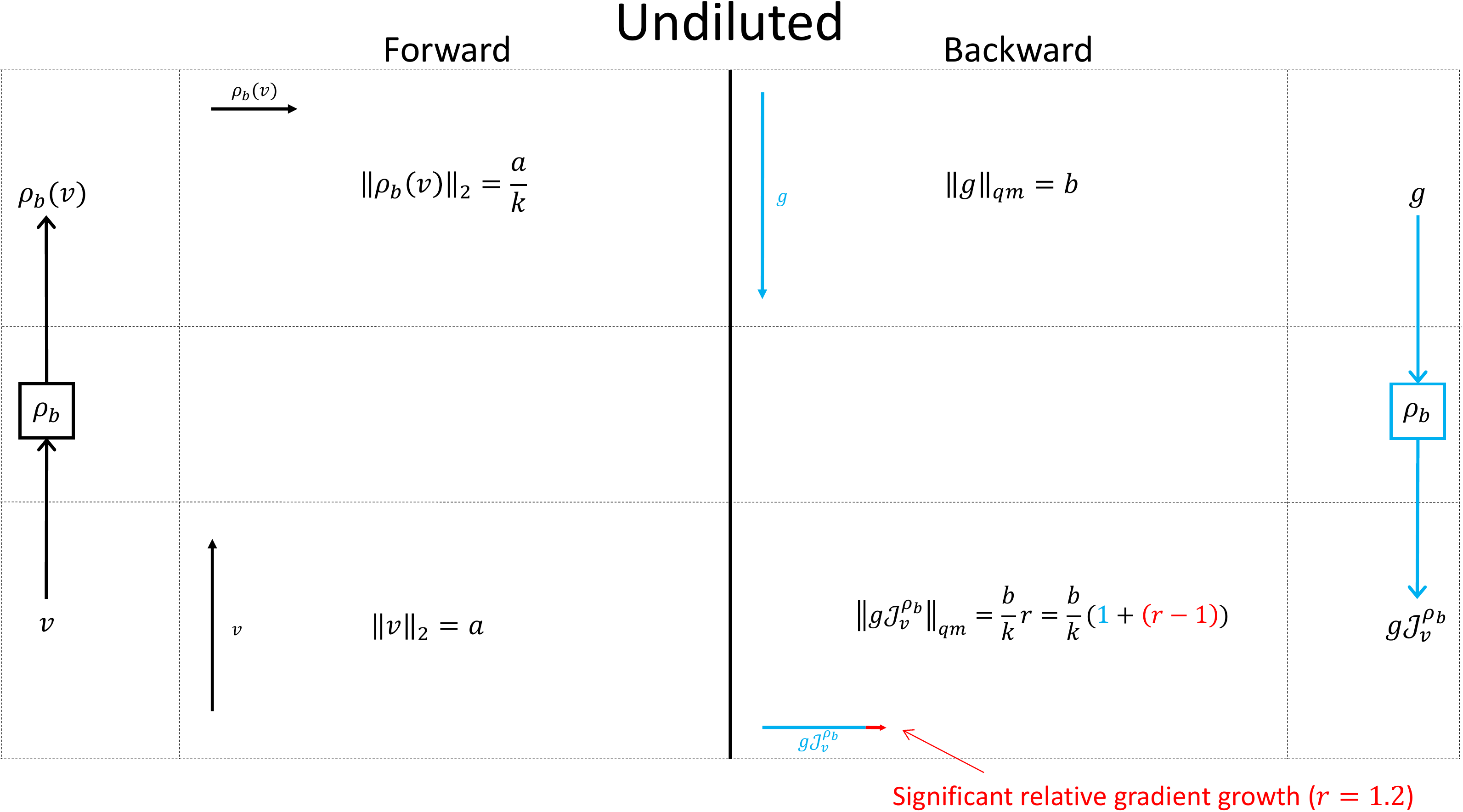}
\includegraphics[width=\textwidth]{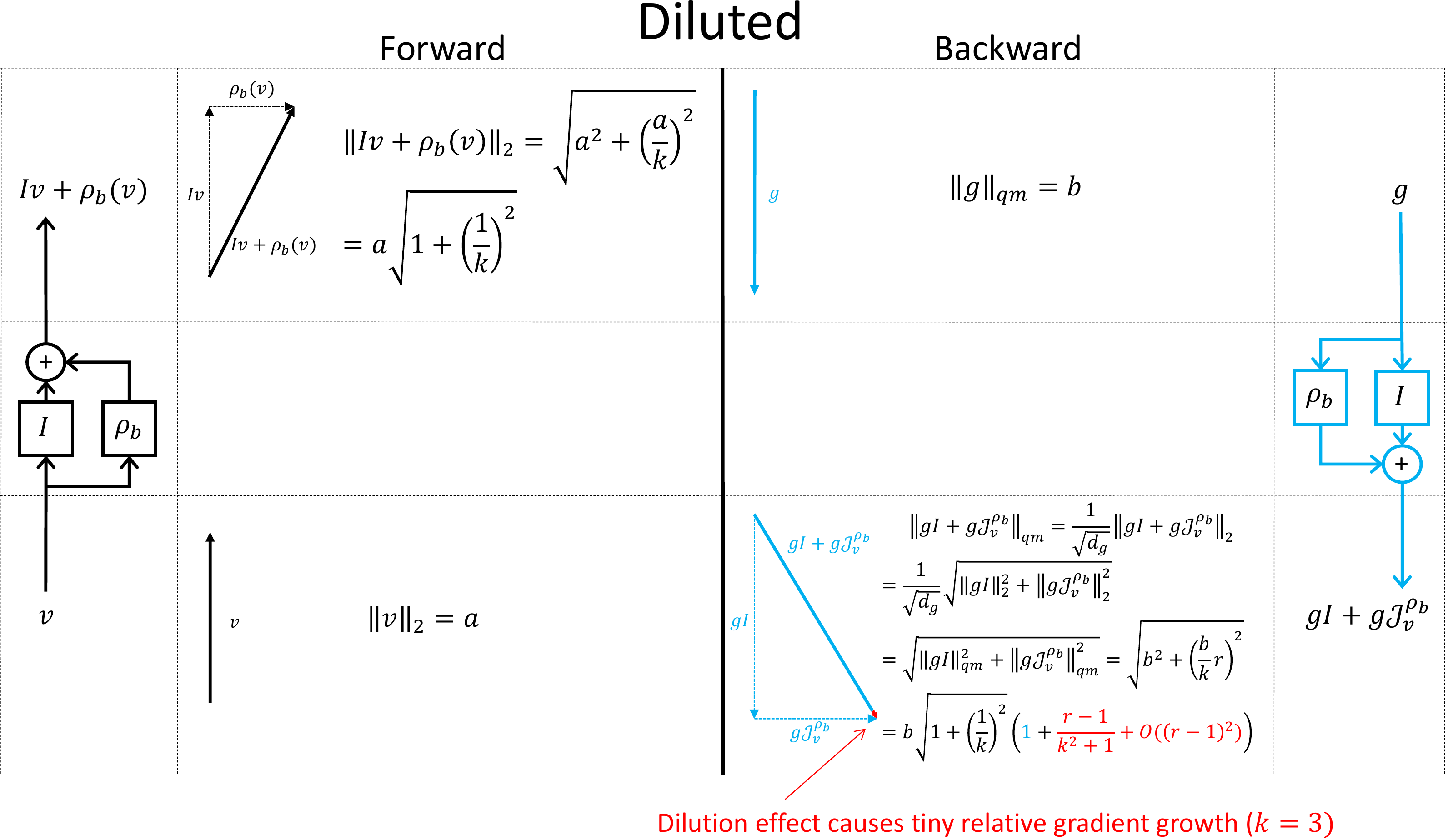}
\caption{Illustration of theorem \ref{dilutionTheorem}. See main text for details.} \label{im_dilution}
\end{figure}

The mechanism underlying theorem \ref{dilutionTheorem} is illustrated in figure \ref{im_dilution}. In the upper half of this figure, we show a block function $\rho_b$ that causes the GSC to grow with rate $r$, i.e. $\frac{GSC(b+1,0)}{GSC(b,0)} = \frac{||v||_2||g\mathcal{J}_v^{\rho_b}||_{qm}}{||\rho_b(v)||_2||g||_{qm}} = r$. $g$ represents the incoming error gradient. In the bottom half, we add an identity skip connection to $\rho$. Assuming both $Iv$ and $\rho_b(v)$ as well as $g$ and $g\mathcal{J}_v^{\rho_b}$ are uncorrelated and thus orthogonal, the Pythagorean theorem ensures that the growth of the GSC is severely reduced.

\begin{figure}[h]
\adjincludegraphics[width=\textwidth]{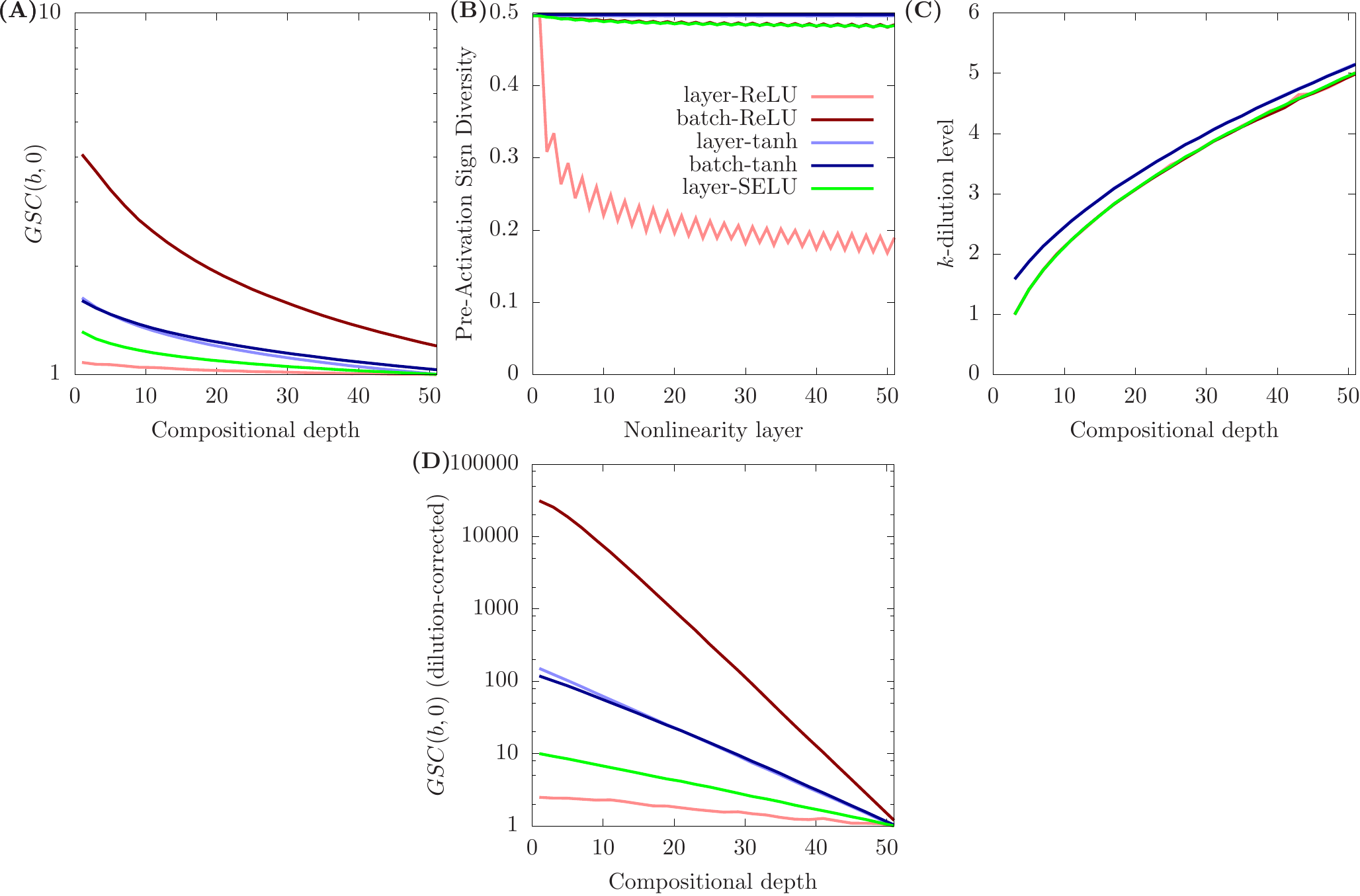}
\caption{Key metrics for ResNet architectures at in their randomly initialized state evaluated on Gaussian noise. In A and C, only values obtained between blocks are plotted. The x axis shows depth in terms of the number of linear layers counted from the input. The x axis in B counts nonlinearity layers, starting from the input. In C, batch-tanh shadows layer-tanh and SeLU shadows ReLU and layer-ReLU.} \label{GSCresfigure}
\end{figure}

To validate our theory, we repeated the experiments in figure \ref{GSCfigure} with 5 ResNet architectures: layer-ReLU, batch-ReLU, layer-tanh, batch-tanh and layer-SELU. Each residual block is bypassed by an identity skip connection and composed of 2 sub-blocks of 3 layers each: first a normalization layer, then a nonlinearity layer, and then a linear layer, similar to \cite{canonicalResNet,wideResNet}. For further details, see section \ref{detailssection}. Comparing figure \ref{GSCresfigure}A to figure \ref{GSCfigure}A, we find the gradient growth is indeed much lower for ResNet compared to corresponding vanilla networks, with much of it taking place in the lower layers. In figure \ref{GSCresfigure}B we find that the growth of domain bias for layer-ReLU, as measured by pre-activation sign diversity, is also significantly slowed.

We then went on to check whether the gradient reduction experienced is in line with theorem \ref{dilutionTheorem}. We measured the $k$-dilution level $k_b$ induced by the skip block as well as $r_b$, the growth rate of $GSC(b,0)$, at each individual block $b$. We then replaced the growth rate with $1 + (k_b^2+1)(r_b - 1)$, obtaining new GSC curves which are shown in figure \ref{GSCresfigure}D. Indeed, the GSC of the exploding architectures now again grows almost linearly in log space, with the exception of batch-ReLU in the lowest few layers. The explosion rates closely track those in figure \ref{GSCfigure}A, being only slightly higher. This confirms that the estimate of the magnitude of gradient reduction from theorem \ref{dilutionTheorem} is accurate in practical architectures. The $k$-dilution levels are shown in figure \ref{GSCresfigure}C. They grows as $\approx \sqrt{B-b}$ as the skip block is the accumulation of approximately $B-b$ uncorrelated residual blocks of equal size.

We then repeated the CIFAR10 experiments shown in figure \ref{dyna} with our 5 ResNet architectures. The results are shown in figure \ref{dynaRes}. As expected, in general, ResNet enables higher relative update sizes, achieves lower error, a higher effective depth and is less ``robust'' to Taylor approximation than corresponding vanilla networks. The only exception to this trend is the layer-SELU ResNet when compared to the SELU vanilla network, which already has a relatively slowly exploding gradient to begin with. Note that the severe reduction of the GSC persists throughout training (figure \ref{dynaRes}C). Also see table \ref{CIFARtable} to compare final error values. Note that in order to make the effective depth results in figure \ref{dynaRes}D comparable to those in figure \ref{dyna}D, we applied the residual trick to ResNet. We let the initial function $i$ encompass not just the skip block $s$, but also the initial residual block function $\rho(\theta^{(0)})$. In fact, we broke down each layer in the residual block into its initial and residual function. See section \ref{depthComputationSection} for more details. Note that our effective depth values for ResNet are much higher than those of \cite{effectiveDepth}. This is because we use a much more conservative estimate of this intractable quantity for both ResNet and vanilla networks.

Gradient reduction is achieved not just by identity skip connections but, as theorem \ref{dilutionTheorem} suggests, also by skip blocks that multiply the incoming value with e.g. a Gaussian random matrix. Using Gaussian skip connections, the amount of gradient reduction achieved in practice is not quite as great (table \ref{gaussianTable}).

\cite{effectiveDepth} argues that deep ResNets behave like an ensemble of relatively shallow networks. We argue that comparable vanilla networks often behave like ensembles of even shallower networks. \cite{resNetRefinement} argues that deep ResNets are robust to lesioning. We argue that comparable vanilla networks are often even more robust to depth reduction when considering the first order Taylor expansion.

\subsection{The limits of dilution}

$k$-dilution has its limits. Any $k$-diluted function with large $k$ is close to a linear function. Hence, we can view $k$-dilution as another form of pseudo-linearity that can damage representational capacity. It also turns out that at least in the randomly initialized state, dilution only disappears slowly as diluted functions are composed. If the diluting linear functions $s_b$ are identity functions, this corresponds to feature refinement as postulated by \cite{resNetRefinement}.

\begin{theorem} \label{dilutionComposition}
Under certain conditions, the composition of $B$ randomly initialized blocks that are $k_b$-diluted in expectation respectively is $\Big(\big(\prod_l(1+\frac{1}{k_b^2})\big)-1\Big)^{-\frac{1}{2}}$-diluted in expectation. (See section \ref{theo3p5proof} for details.)
\end{theorem} 

More simply, assume all the $k_b$ are equal to some $k$. Ignoring higher-order terms, the composition is $\frac{1}{\sqrt{B}}k$-diluted. Under the conditions of theorem \ref{dilutionComposition}, the flipside of an $O(k^2)$ reduction in gradient via dilution is thus the requirement of $O(k^2)$ blocks to eliminate that dilution. This indicates that the overall amount of gradient reduction achievable through dilution without incurring catastrophic pseudo-linearity is limited.

\subsection{Choosing dilution levels}

The power of our theory lies in exposing the GSC-reducing effect of skip connections for general neural network architectures. As far as we know, all comparable previous works (e.g. \cite{meanFieldResNet,shattering}) demonstrated similar effects only for specific architectures. Our argument is not that certain ResNets achieve a certain level of GSC reduction, but that ResNet users have the power to choose the level of GSC reduction by controlling the amount of dilution. While the level of dilution increases as we go deeper in the style of ResNet architecture we used for experiments in this section, this need not be so. 

The skip block $s$ and residual block $\rho$ can be scaled with constants to achieve arbitrary, desired levels of dilution \citep{inceptionv4,layerSplitting2}. Alternatively, instead of putting all normalization layers in the residual blocks, we could insert them between blocks. This would keep the dilution level constant. 

\subsection{On the relationship of dilution, linear approximation error and the standard deviation of absolute Jacobian eigenvalues} \label{relationshipSection}

The insights presented in this section cast a new light on earlier results. For example, the concept of linear approximation error as shown in figure \ref{GSCfigure}F is similar to the concept of dilution. Therefore we might expect that a low linear approximation error would be associated with a low gradient growth. This is precisely what we find. The explosion rates from figure \ref{GSCfigure}A display a similar magnitude as the linear approximation errors in figure \ref{GSCfigure}F. This also explains how the tanh architecture avoids exploding gradients - via extreme pseudo-linearity, as depicted in figure \ref{pseudoLin}C. tanh also does not perform well (table \ref{CIFARtable} / figure \ref{dynaCollapse}).

Conversely, we can interpret dilution in terms of the linear approximation error. If we view the skip block as the signal and the residual block as the ``noise'', then increasing the dilution corresponds to an increase in the signal relative to the noise. Specifically, increasing dilution has a squared effect on the signal-to-noise ratio, which suggests a squared increase in the number of blocks is needed to bring the signal-to-noise ratio back to a given level. This leads us back to theorem \ref{dilutionComposition}.

In section \ref{widerViewsection}, we pointed to a reduction in the standard deviation of absolute singular values of the layer-wise Jacobian as a strategy for reducing gradient growth. Dilution with the identity or an orthogonal matrix, and to a lesser extent dilution with a Gaussian random matrix, achieves exactly that. Furthermore, we note that in theorem \ref{bigGradTheorem}, we have $\epsilon = O(\sigma_{|s|}^2)$ as $\sigma_{|s|} \rightarrow 0$. Of course, $k$-dilution with the identity or an orthogonal matrix leads to a $k$-fold in reduction $\sigma_{|s|}$. So theorem \ref{bigGradTheorem} suggests that a $k$-fold reduction in $\sigma_{|s|}$ may lead to a $O(k^2)$-fold reduction in gradient growth, which leads us back to theorem \ref{dilutionTheorem}.

\subsection{The orthogonal initial state}

Applying the residual trick to ResNet reveals several insights. The difference between ResNet and vanilla networks in terms of skip connections is somewhat superficial, because both ResNet and vanilla networks can be expressed as residual networks in the framework of equation \ref{residualForm}. Also, both ResNet and vanilla networks have nonlinear initial functions, because $\rho_b(\theta_b^{(0)})$ is initially nonzero and nonlinear. However, there is one key difference. The initial functions of ResNet are closer to a linear transformation and indeed closer to an orthogonal transformation because they are composed of a nonlinear function $\rho_b(\theta_b^{(0)})$ that is significantly diluted by what is generally chosen to be an orthogonal transformation $s_b$. Therefore, ResNet, while being conceptually more complex, is mathematically simpler.

We have shown how ResNets achieve a reduced gradient via $k$-diluting the initial function. And just as with effective depth, the residual trick allows us to generalize this notion to arbitrary networks.

\pagebreak

\paragraph{Definition 5.} We say a residual network $f(\theta)$ has an {\it `orthogonal initial state' (OIS)} if each initial function $i_l$ is a multiplication with an orthogonal matrix or a slice / multiple thereof.\\

Any network that is trained from an (approximate) OIS can benefit from reduced gradients via dilution to the extent to which initial and residual function are uncorrelated. ResNet is a style of architecture that achieves this, but it is far from being the only one. \cite{shattering} introduced the `looks-linear initialization' (LLI) for ReLU networks, where initial weights are set in a clever way to bypass the nonlinear effect of the ReLU layer. We detail this initialization scheme in section \ref{looklineardetails}. A plain ReLU network with weights set by LLI achieves not only an approximate OIS, but outperformed ResNet in the experiments of \cite{shattering}. In table \ref{CIFARtable}, we show that applying LLI to our ReLU architecture causes it to outperform ResNet in our CIFAR10 experiments as well. In figure \ref{dynaLL}C, we find that indeed LLI reduces the gradient growth of batch-ReLU drastically not just in the initialized state, but throughout training even as the residual functions grow beyond the size achieved under Gaussian initialization (compare figure \ref{dynaLL}E to \ref{dyna}E and \ref{dynaCollapse}E).

DiracNet \citep{diracnet} also achieves an approximate OIS. A simpler but much less powerful strategy is to initialize weight matrices as orthogonal matrices instead of Gaussian matrices. This reduces the gradient growth in the initialized state somewhat (table \ref{gaussianTable}).

\subsection{The power of initial vs residual dilution}

An orthogonal initial state is not enough to attain high performance. Trivially, an orthogonal linear network without nonlinearity or normalization layers achieves an orthogonal initial state, but does not attain high performance. Clearly, we need to combine orthogonal initial functions with sufficiently {\it non}-linear residual functions.

The ensemble view of deep networks detailed in section \ref{backgroundSection} reveals the power of this approach. With high probability, the input to an ensemble member must pass through a significant number of initial function to reach the prediction layer. Therefore, having non-orthogonal initial functions is akin to taking a shallow network and adding additional, {\it untrainable} non-orthogonal layers to it. This has obvious downsides such as a collapsing domain and / or exploding gradient, and an increasingly unfavorable eigenspectrum of the Jacobian \citep{orthogonalInitialization}. One would ordinarily not make the choice to insert such untrainable layers. While there has been some success with convolutional networks where lower layers are not trained (e.g. \cite{randomConv1,randomConv2}), it is not clear whether such networks are capable of outperforming other networks where such layers are trained.

Conversely, using non-linear residual functions means that the input to an ensemble member passes through a significant number of trainable, composed, nonlinear residual functions to reach the prediction layer. This is precisely what ensures the representational capacity.

While tools that dilute the initial function such as skip connections or LLI do not resolve the tension between exploding gradients and collapsing domains, they reduce the pathology by specifically avoiding untrainable, and thus potentially unnecessary non-orthogonality contained in the initial functions. To further validate this idea, we compared these strategies against another that dilutes both initial and residual functions - leaky ReLU.

Leaky ReLU \citep{leakyRelu} is a generalization of ReLU. Instead of assigning a zero value to all negative pre-activations, leaky ReLU multiplies them with the `leakage parameter' $c$. \cite{leakyRelu} set $c$ set to 0.01, but this is not a strict requirement. In fact, when $c=1$, leaky ReLU becomes the identity function. If $c=0$, we recover ReLU. Therefore, by varying $c$ we can interpolate between the linear identity function and the nonlinear ReLU function. The closer $c$ is to 1, the more ReLU is diluted. When $c=1$, the leaky ReLU network achieves an initial linear state. In contrast to ResNet, the dilution of the ReLU layer affects the signal that passes through the residual weight matrices.

\begin{figure}
\centering
\includegraphics[width=0.7\textwidth]{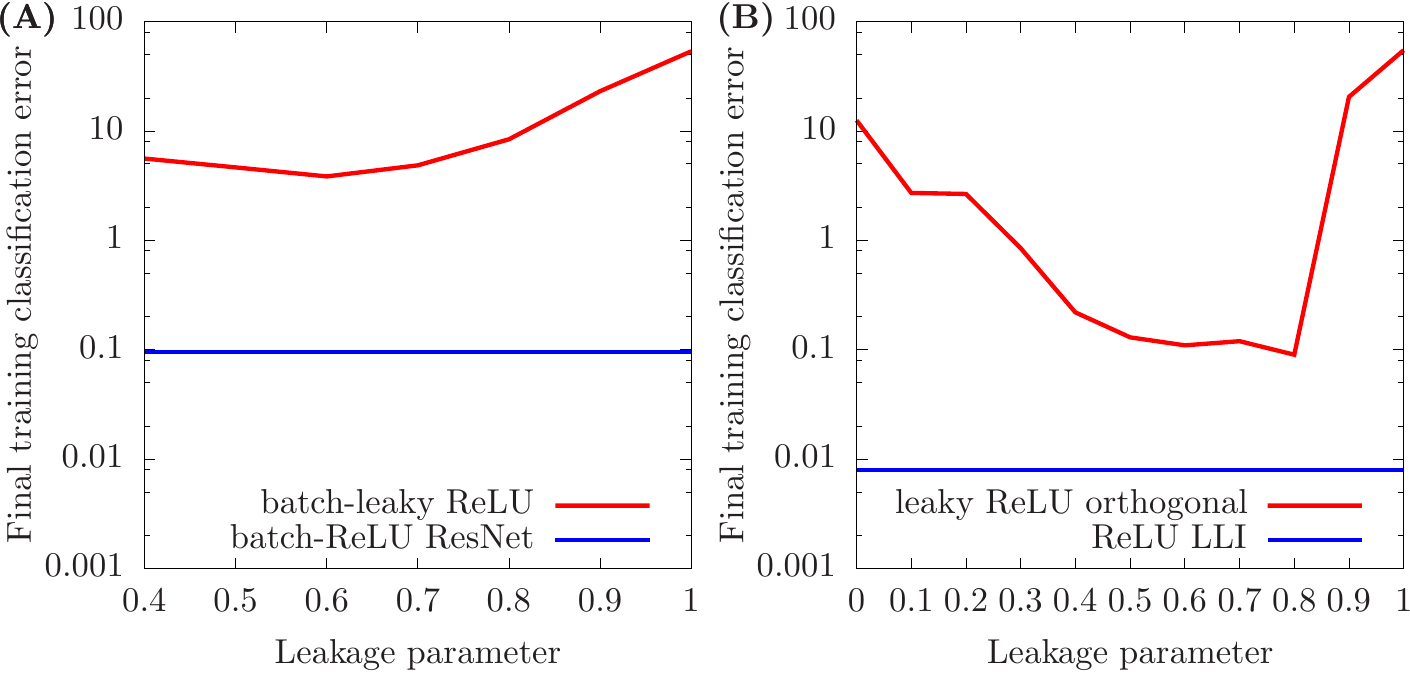}
\caption{Comparing different dilution styles via their performance on CIFAR10.} \label{OIS}
\end{figure}

We repeat our CIFAR10 experiments with the leaky-ReLU and batch-leaky ReLU architectures. The latter is comparable to batch-ReLU ResNet, the former to LLI ReLU. To make the comparison to LLI ReLU even more faithful, we initialized the batch-leaky ReLU network with orthogonal weight matrices. Therefore, when $c=1$, the leaky ReLU architecture achieves an OIS. The results are shown in figure \ref{OIS} and table \ref{CIFARtable}. While varying the leakage parameter can have a positive effect on performance, as expected, initial-only dilution schemes perform much better.

The big question is now: What is the purpose of {\it not} training a network from an orthogonal initial state? We are not aware of such a purpose. Since networks with orthogonal initial functions are mathematically simpler than other networks, we argue they should be the default choice. Using non-orthogonality in the initial function, we believe, is what requires explicit justification.

\cite{shattering} asks in the title: {\it If ResNet is the answer, then what is the question?} We argue that a better question would be: {\it Is there a question to which vanilla networks are the answer?}

\section{Related work} \label{relatedWorksection}

In this paper, we have discussed exploding gradients and collapsing domains. In this section, we review related metrics and concepts from literature.

Our work bears similarity to a recent line of research studying deep networks using mean field theory \citep{meanFieldOrig,meanFieldExplosion,meanFieldResNet, widthHacking}. The authors study infinitely wide and deep networks in their randomly initialized state. They identify two distinct regimes, order and chaos, based on whether the correlation between two forward activation vectors corresponding to two different datapoints converges exponentially to one (`order'), exponentially to a value less than one (`chaos') or sub-exponentially (`edge of chaos'), as the vectors are propagated towards infinite depth. They show that for MLPs where the forward activation vector length converges, order corresponds to gradient vanishing according to the metric of e.g. gradient vector length. If the network is also a tanh MLP, chaos corresponds to gradient explosion according to the same metrics. They show how to use mean field theory as a powerful and convenient tool for the static analysis of network architectures and obtain a range of interesting results. Our work differs from and extends this line of work in several ways, as we discuss in detail in section \ref{meanFieldSection}.

Recently, \cite{layerSplitting1,antisymmResNet1,antisymmResNet2, layerSplitting2} introduced the concept of stability according to dynamical systems theory for ResNet architectures. A central claim is that in architectures that achieve such stability, both forward activations and gradients (and hence the GSC) are bounded as depth goes to infinity. These papers derive a range of valuable strategies such as deepening a ResNet gradually by duplicating residual blocks and achieving effective regularization by tying weights in consecutive blocks. In our work, we showed how dilution can suppress gradient growth drastically (theorem \ref{dilutionTheorem}) and how dilution can disappear very slowly with increasing depth (theorem \ref{dilutionComposition}). We are not convinced that the strategies these papers introduce offer significant additional benefit over general dilution in terms of reducing gradient growth. We provide experimental results and further discussion in section \ref{dynamicalSection}.

We build on the work of \cite{shattering}, who introduced the concept of gradient shattering. This states that in deep networks, gradients with respect to nearby points become more and more uncorrelated with depth. This is very similar to saying that the gradient is only informative in a smaller and smaller region around the point at which it is taken. This is precisely what happens when gradients explode and also, as we argue in section \ref{collapsingSection}, under domain bias. Therefore, the exploding gradient problem and domain bias problem can be viewed as a further specification of the shattering gradient problem rather than as a counter-theory or independent phenomenon. 

We extend the work of \cite{shattering} in several ways. First, they claim that the exploding gradient problem ``has been largely overcome''. We show that this is not the case, especially in the context of very deep batch-ReLU MLPs, which \cite{shattering} investigate. Second, by using effective depth we make a rigorous argument as to why exploding gradients cause training difficulty. While \cite{shattering} point out that shattering gradients interfere with theoretical guarantees that exist for specific optimization algorithms, they do not provide a general argument as to why shattering gradients are in fact a problem. Third, our analysis extends beyond ReLU networks.

We also build on the work of \cite{pathLength}. They showed that both trajectories and small perturbations, when propagated forward, can increase exponentially in size. However, they do not distinguish two important cases: (i) an explosion that is simply due to an increase in the scale of forward activations and (ii) an explosion that is due to an increase in the gradient relative to forward activations. We are careful to make this distinction and focus only on case (ii). Since this is arguably the more interesting case, we believe the insights generated in our paper are more robust.

\citet{orthogonalInitialization} and \citet{eigenspectrum} investigated another important pathology of very deep networks: the divergence of singular values in multi-layer Jacobians. As layer-wise Jacobians are multiplied, the variances of their singular values compound. This leads to the direction of the gradient being determined by the dominant eigenvectors of the multi-layer Jacobian rather than the label, which slows down training considerably. 

In their seminal paper, \cite{batchnorm} motivated batch normalization with the argument that changes to the distribution of latent representations, which they term `covariate shift', are pathological and need to be combated. This argument was then picked up by e.g. \cite{weightNormalization} and \cite{cosineNormalization} to motivate similar normalization schemes. We are not aware of any rigorous definition of the `covariate shift' concept nor do we understand why it is undesirable. After all, isn't the very point of training deep networks to have each layer change the function it computes, to which other layers co-adapt, to which then other layers co-adapt and so on? Having each layer fine-tune its weights in response to shifts in other layers seems to us to be the very mechanism by which deep networks achieve high accuracy.

A classical notion of trainability in optimization theory is the conditioning of the Hessian. This can also deteriorate with depth. Recently, \cite{whitening} introduced an architecture that combats this pathology in an effective and computationally tractable way via iterative numerical methods and matrix decomposition. Matrix decomposition has also been used by e.g. \cite{fastfoodUnitaryRNN,orthRNN} to maintain orthogonality of recurrent weight matrices. Maybe such techniques could also be used to reduce the divergence of singular values of the layer-wise Jacobians during training.

\section{Conclusion}\label{conclusionsection}

\paragraph{Summary} In this paper, we demonstrate that contrary to popular belief, many MLP architectures composed of popular layer types exhibit exploding gradients (section \ref{explodingsection}), and those that do not exhibit collapsing domains (section \ref{collapsingSection}) or extreme pseudo-linearity (section \ref{relationshipSection}). This tradeoff is caused by the discrepancy between geometric and quadratic means of the absolute singular values of layer-wise Jacobians (section \ref{widerViewsection}). Both sides of this tradeoff can cause pathologies. Exploding gradients, when defined by the GSC (section \ref{GSCsection}) cause low effective depth (section \ref{shallowsection}). Collapsing domains can cause pseudo-linearity and also low effective depth (section \ref{collapsingSection}). However, both pathologies can be avoided to a surprisingly large degree by eliminating untrainable, and thus potentially unnecessary non-orthogonality contained in the initial functions. Making the initial functions more orthogonal via e.g. skip connections leads to improved outcomes (section \ref{ResNetsection}).


\begin{figure}
\centering
\includegraphics[width=0.7\textwidth]{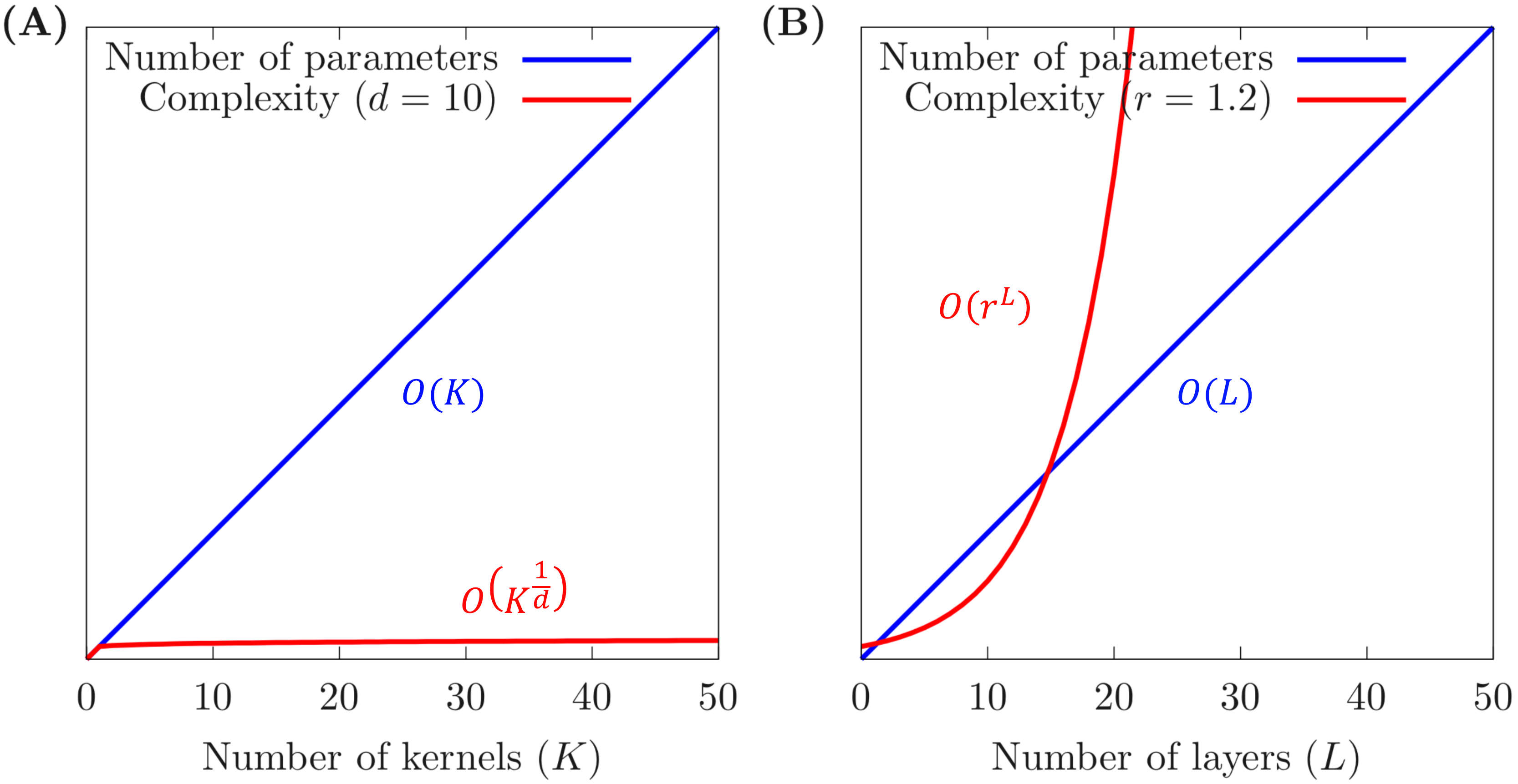}
\caption{Illustration of the functional complexity of neural networks versus classical nonlinear models as represented by Gaussian kernel machines. Note: This figure does not display experimental results.} \label{im_surfaceComplexityLabeled}
\end{figure}

The picture of deep learning that emerges throughout this paper is considerably different from classical machine learning. In classical nonlinear models such as Gaussian kernel machines, we experience the curse of dimensionality, where the complexity of the function computed by the model grows as $O(K^\frac{1}{d})$, where $K$ is the number of kernels and $d$ is the dimentionality of the data. Conversely, the complexity of the function computed by neural networks increases exponentially with depth, independently of the dimensionality of the data, assuming that the network exhibits e.g. a constant rate of gradient explosion. A visual high-level summary of this paper is shown in figure \ref{im_surfaceComplexityLabeled}.


\paragraph{Practical Recommendations}

\begin{itemize}
\item {\bf Train from an orthogonal initial state,} i.e. initialize the network such that it is a series of orthogonal linear transformations. This can greatly reduce the growth of the GSC and domain collapse not just in the initial state, but also as training progresses. It can prevent the forward activations from having to pass through unnecessary non-orthogonal transformations. Even if a perfectly orthogonal initial state is not achievable, an architecture that approximates this such as ResNet can still confer significant benefit.
\item When not training from an orthogonal initial state, {\bf avoid low effective depth}. A low effective depth signifies that the network is composed of an ensemble of networks significantly shallower than the full network. If the initial functions are not orthogonal, the values computed by these ensemble members have to pass through what may be unnecessary and harmful untrainable non-orthogonal transformations. Low effective depth may be caused by, for example, exploding gradients or a collapsing domain.
\item {\bf Avoid pseudo-linearity.} For the representational capacity of a network to grow with depth, linear layers must be separated by nonlinearities. If those nonlinearities can be approximated by linear functions, they are ineffective. Pseudo-linearity can be caused by, for example, a collapsing domain or excessive dilution.
\item {\bf Keep in mind that skip connections help in general, but other techniques do not} Diluting a nonlinear residual function with an uncorrelated linear initial function can greatly help with the pathologies described in this paper. Techniques such as normalization layers, careful initialization of weights or SELU nonlinearities can prevent the explosion or vanishing of forward activations. Adam, RMSprop or vSGD can improve performance even if forward activations explode or vanish. While those are important functionalities, these techniques in general neither help address gradient explosion relative to forward activations as indicated by the GSC nor the collapsing domain problem. 
\item {\bf As the GSC grows, adjust the step size.} If it turns out that some amount of growth of the GSC is unavoidable or desirable, weights in lower layers could benefit from experiencing a lower relative change during each update. Optimization algorithms such as RMSprop or Adam may partially address this. 
\item {\bf Control dilution level to control network properties.} Skip connections, normalization layers and scaling constants can be placed in a network to trade off gradient growth and representational capacity. Theorem \ref{dilutionTheorem} can be used for a static estimate of the amount of gradient reduction achieved. Similarly, theorem \ref{dilutionComposition} can be used for a static estimate of the overall dilution of the network.
\item {\bf Great compositional depth may not be optimal.} Networks with more than 1000 layers have recently been trained \citep{resNet}. \cite{antisymmResNet1} gave a formalism for training arbitrarily deep networks. However, ever larger amounts of dilution are required to prevent gradient explosion \citep{inceptionv4}. This may ultimately lead to an effective depth much lower than the compositional depth and individual layers that have a very small impact on learning outcomes, because functions they represent are very close to linear functions. If there is a fixed parameter budget, it may be better spent on width than extreme depth \citep{wideResNet}.
\end{itemize}

\paragraph{Implications for deep learning research}

\begin{itemize}
\item {\bf Exploding gradients matter.} They are not just a numerical quirk to be overcome by rescaling but are indicative of an inherently difficult optimization problem that cannot be solved by a simple modification to a stock algorithm.
\item {\bf GSC is an effective benchmark for gradient explosion.} For the first time, we established a rigorous link between a metric for exploding gradients and training difficulty. The GSC is also robust to network rescaling, layer width and individual layers.
\item {\bf Any neural network is a residual network.} The residual trick allows the application of ResNet-specific tools such as the popular theory of effective depth to arbitrary networks.
\item {\bf Step size matters when studying the behavior of networks.} We found that using different step sizes for different layers had a profound impact on the training success of various architectures. Many studies that investigate fundamental properties of deep networks either do not consider layerwise step sizes (e.g. \cite{meanFieldExplosion}) or do not even consider different global step sizes (e.g. \cite{sharpMinima}). This can lead to inaccurate conclusions.
\end{itemize}

We provide continued discussion in section \ref{discussionSection}.

\begin{table}[h]
\centering
{
\footnotesize
\begin{tabular}{cccccccc}
Nonlinearity&Normalization&Matrix type&Skip type&Width&$GSC(L+1,0)$&St. Dev.&Sign Div.\\ \hline\hline
ReLU&none&Gaussian&none&100&1.52&0.22&{\color{red} 0.030}\\
ReLU&layer&Gaussian&none&100&1.16&0.096&{\color{red} 0.029}\\
ReLU&batch&Gaussian&none&100&{\color{red} 5728}&1.00&0.41\\
tanh&none&Gaussian&none&100&1.26&{\color{red} 0.096}&0.50\\
tanh&layer&Gaussian&none&100&{\color{red} 72.2}&1.00&0.50\\
tanh&batch&Gaussian&none&100&{\color{red} 93.6}&1.00&0.50\\
SELU&none&Gaussian&none&100&{\color{red} 6.36}&0.97&0.42\\
ReLU&batch&Gaussian&none&200&{\color{red} 5556}&1.00&0.42\\
ReLU&batch&Gaussian&none&100/200&{\color{red} 5527}&1.00&0.41\\
SELU&none&Gaussian&none&200&{\color{red} 5.86}&0.99&0.45\\
SELU&none&Gaussian&none&100/200&{\color{red} 6.09}&0.98&0.43\\
ReLU&none&orthogonal&none&100&1.29&0.20&{\color{red} 0.03}\\
ReLU&layer&orthogonal&none&100&1.00&0.10&{\color{red} 0.03}\\
ReLU&batch&orthogonal&none&100&{\color{red} 5014}&1.00&0.42\\
tanh&none&orthogonal&none&100&1.18&{\color{red} 0.10}&0.50\\
tanh&layer&orthogonal&none&100&{\color{red} 56.3}&1.00&0.50\\
tanh&batch&orthogonal&none&100&{\color{red} 54.6}&1.00&0.50\\
SELU&none&orthogonal&none&100&{\color{red} 5.47}&1.00&0.49\\
ReLU&none&looks-linear&none&100&1.00&1.00&0.50\\
ReLU&layer&looks-linear&none&100&1.00&1.00&0.50\\
ReLU&batch&looks-linear&none&100&1.00&1.00&0.50\\
ReLU&layer&Gaussian&identity&100&1.08&0.56&0.19\\
ReLU&batch&Gaussian&identity&100&4.00&1.00&0.48\\
tanh&layer&Gaussian&identity&100&1.63&1.00&0.50\\
tanh&batch&Gaussian&identity&100&1.57&1.00&0.50\\
SELU&layer&Gaussian&identity&100&1.31&0.99&0.48\\
ReLU&layer&Gaussian&Gaussian&100&1.17&0.56&0.18\\
ReLU&batch&Gaussian&Gaussian&100&4.50&1.00&0.48\\
tanh&layer&Gaussian&Gaussian&100&1.97&1.00&0.50\\
tanh&batch&Gaussian&Gaussian&100&1.71&1.00&0.50\\
SELU&layer&Gaussian&Gaussian&100&1.53&9.97&0.48
\end{tabular}
}
\caption{Key metrics for architectures in their randomly initialized state evaluated on Gaussian noise. In the `Normalization' column, `layer' refers to layer normalization, `batch' refers to batch normalization and `none' refers to an absence of a normalization layer. In the `Matrix type' column, `Gaussian' refers to matrices where each entry is drawn independently from a Gaussian distribution. `orthogonal' refers to a uniformly random orthogonal matrix and `looks-linear' refers to the initialization scheme proposed by \cite{shattering} and expounded in section \ref{looklineardetails}. In the `Skip type' column, `identity' refers to identity skip connections and `Gaussian' refers to skip connections that multiply the incoming value with a matrix where each entry is drawn independently from a Gaussian distribution. `none' refers to an absence of skip connections. In the `Width' column, `100/200' refers to linear layers having widths alternating between 100 and 200. `St. Dev.' refers to pre-activation standard deviation at the highest nonlinearity layer. `Sign Div.' refers to pre-activation sign diversity at the highest nonlinearity layer. For further details, see section \ref{detailssection}. Red values indicate gradient explosion or pseudo-linearity.}
\label{gaussianTable}
\end{table}
\begin{table}[h]
\centering
{
\footnotesize
\begin{tabular}{cccccc}
Nonlinearity&Norm.&Matrix type&Skip type&Error (custom s.s.)&Error (single s.s.)\\ \hline\hline
ReLU&none&Gaussian&none&31.48\%&{\bf 19.24\%}\\
ReLU&layer&Gaussian&none&42.48\%&{\bf 21.23\%}\\
ReLU&batch&Gaussian&none&{\bf 34.83\%}&76.65\%\\
tanh&none&Gaussian&none&23.42\%&{\bf 16.22\%}\\
tanh&layer&Gaussian&none&{\bf 1.92\%}&17.5\%\\
tanh&batch&Gaussian&none&{\bf 12.31\%}&23.8\%\\
SELU&none&Gaussian&none&{\bf 0.24\%}&1.78\%\\
ReLU&none&looks-linear&none&{\bf 0.002\%}&0.008\%\\
ReLU&layer&looks-linear&none&{\bf 0.77\%}&1.2\%\\
ReLU&batch&looks-linear&none&0.38\%&{\bf 0.19\%}\\
tanh&layer&Gaussian&id&0.35\%&{\bf 0.27\%}\\
tanh&batch&Gaussian&id&{\bf 0.13\%}&0.24\%\\
ReLU&layer&Gaussian&id&2.09\%&{\bf 1.49\%}\\
ReLU&batch&Gaussian&id&{\bf 0.06\%}&0.096\%\\
SELU&layer&Gaussian&id&{\bf 1.55\%}&{\bf 1.55\%}\\
leaky ReLU ($c=0.4$)&batch&Gaussian&none&-&5.6\%\\
leaky ReLU ($c=0.5$)&batch&Gaussian&none&-&4.64\%\\
leaky ReLU ($c=0.6$)&batch&Gaussian&none&-&3.84\%\\
leaky ReLU ($c=0.7$)&batch&Gaussian&none&-&4.86\%\\
leaky ReLU ($c=0.8$)&batch&Gaussian&none&-&8.41\%\\
leaky ReLU ($c=0.9$)&batch&Gaussian&none&-&23.14\%\\
none&batch&Gaussian&none&-&53.55\%\\
ReLU&none&orthogonal&none&-&12.51\%\\
leaky ReLU ($c=0.1$)&none&orthogonal&none&-&2.72\%\\
leaky ReLU ($c=0.2$)&none&orthogonal&none&-&2.66\%\\
leaky ReLU ($c=0.3$)&none&orthogonal&none&-&0.85\%\\
leaky ReLU ($c=0.4$)&none&orthogonal&none&-&0.22\%\\
leaky ReLU ($c=0.5$)&none&orthogonal&none&-&0.13\%\\
leaky ReLU ($c=0.6$)&none&orthogonal&none&-&0.11\%\\
leaky ReLU ($c=0.7$)&none&orthogonal&none&-&0.12\%\\
leaky ReLU ($c=0.8$)&none&orthogonal&none&-&0.09\%\\
leaky ReLU ($c=0.9$)&none&orthogonal&none&-&20.51\%\\
none&none&orthogonal&none&-&54.64\%
\end{tabular}
}
\caption{Training classificaion error for architectures trained on CIFAR10. In the `Normalization' column, `layer' refers to layer normalization, `batch' refers to batch normalization and `none' refers to an absence of a normalization layer. In the `Matrix type' column, `Gaussian' refers to matrices where each entry is drawn independently from a Gaussian distribution. `looks-linear' refers to the looks-linear initialization scheme proposed by \cite{shattering} and expounded in section \ref{looklineardetails}. `orthogonal' refers to uniformly random orthogonal matrices. In the `Skip type' column, `identity' refers to identity skip connections and `none' refers to an absence of skip connections. In the two rightmost columns, we show the training classification error achieved when using a single step size and when using a custom step size for each layer, whenever this experiment was conducted. If two error values are given, the lower one is shown in bold. For further details, see section \ref{detailssection}. For a detailed breakdown of these results, see figures \ref{dyna}, \ref{dynaCollapse}, \ref{dynaRes} and \ref{dynaLL}.}
\label{CIFARtable}
\end{table}

\FloatBarrier

\begin{figure}[h]
\adjincludegraphics[width=\textwidth]{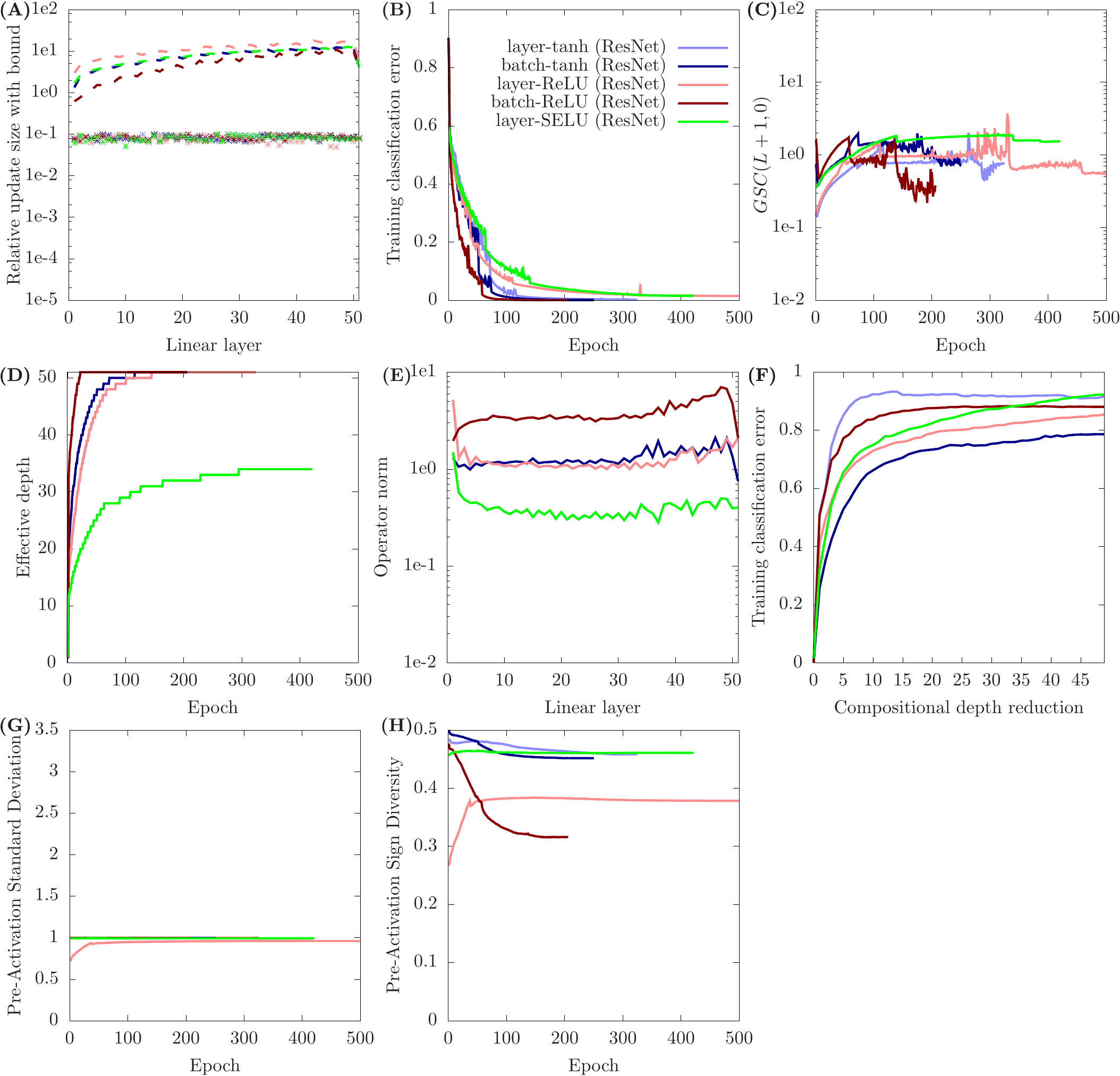}
\caption{Key metrics for ResNet architectures trained on CIFAR10. The top left graph shows the estimated optimal relative update size in each layer according to the algorithm described in section \ref{stepSizeSection}. Remaining graphs show results obtained from training with either a custom step sizes or a single step size, whichever achieved a lower error (see table \ref{CIFARtable}). The top two rows are equivalent to graphs in figure \ref{dyna}. The bottom row shows pre-activation standard deviation and pre-activation sign diversity (see section \ref{gaussianProtocol} for definition) at the highest nonlinearity layer as training progresses.} \label{dynaRes}
\end{figure}

\begin{figure}[h]
\adjincludegraphics[width=\textwidth]{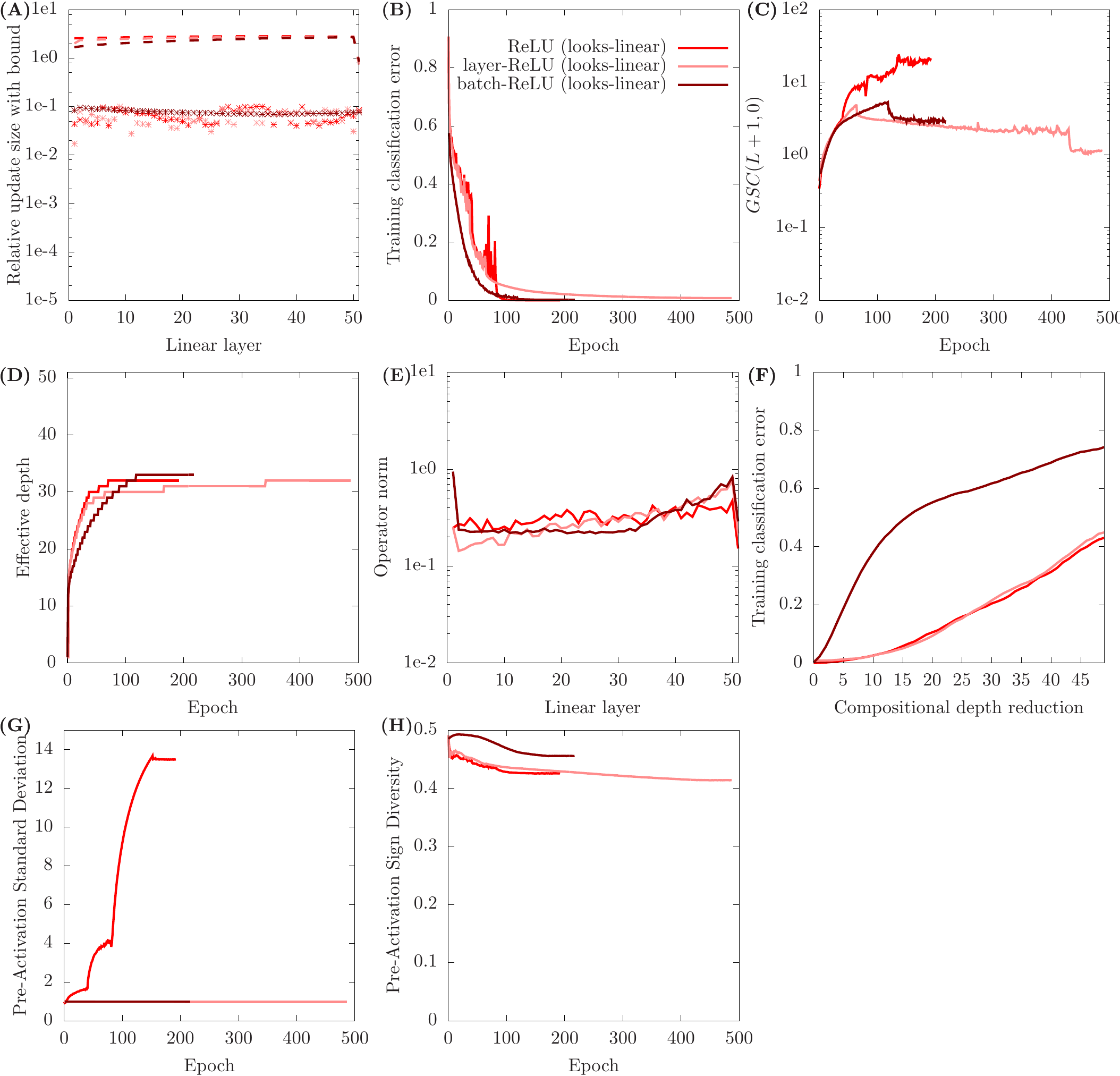}
\caption{Key metrics for ReLU-based architectures with looks-linear initialization trained on CIFAR10. The top left graph shows the estimated optimal relative update size in each layer according to the algorithm described in section \ref{stepSizeSection}. Remaining graphs show results obtained from training with either custom step sizes or a single step size, whichever achieved a lower error (see table \ref{CIFARtable}). The top two rows are equivalent to graphs in figure \ref{dyna}. The bottom row shows pre-activation standard deviation and pre-activation sign diversity (see section \ref{gaussianProtocol} for definition) at the highest nonlinearity layer as training progresses.} \label{dynaLL}
\end{figure}

\FloatBarrier

\newpage
\appendix

\section{Further discussion} \label{discussionSection}

\subsection{Mean field theory - exploding gradients / collapsing domain vs order / chaos} \label{meanFieldSection}

Our work bears similarity to a recent line of research studying deep networks using mean field theory \citep{meanFieldOrig,meanFieldExplosion,meanFieldResNet, widthHacking}. The authors study infinitely wide and deep networks in their randomly initialized state. They identify two distinct regimes, order and chaos, based on whether the correlation between two forward activation vectors corresponding to two different datapoints converges exponentially to one (`order'), exponentially to a value less than one (`chaos') or sub-exponentially (`edge of chaos'), as the vectors are propagated towards infinite depth. For MLPs where the forward activation vector length converges, order corresponds to gradient vanishing according to the metric of e.g. gradient vector length. If the network is also a tanh MLP, chaos corresponds to gradient explosion according to the same metrics. They show how to use mean field theory as a powerful and convenient tool for the static analysis of network architectures and obtain a range of interesting results. 

There are three core similarities between our and their work. Firstly, they discuss the exploding / vanishing gradient dichotomy. Second, the concept of order is very similar to an increasing domain bias. Hence, they show a gradient explosion / domain bias dichotomy for tanh MLPs with stable forward activations. Third, both works rely, at least in part, on the emerging behavior of networks in their randomly initialized state. 

We extend their line of work in several ways. Firstly, we argue that the GSC is a better metric for determining the presence of pathological exploding or vanishing gradients than the quadratic mean of gradient vector entries, which is their metric of choice. Using the GSC, we obtain very different regions of gradient explosion, vanishing and stability for popular architectures. For example, for a constant width ReLU MLP with no biases, using the quadratic mean of gradient vector entries, vanishing is obtained for $\sigma_w < \sqrt{2}$, stability for $\sigma_w = \sqrt{2}$ and explosion for $\sigma_w > \sqrt{2}$. ($\sigma_w$ is defined as the standard deviation of weight matrix entries times the square root of the width, as in \cite{meanFieldOrig}.) For a constant width ReLU MLP with no biases, using the GSC, stability is inevitable. In such networks, the correlation of two forward activation vectors converges sub-exponentially for all weight scales. Hence, such networks are on the edge of chaos for all weight scales, which matches the gradient behavior when considering the GSC. Therefore, the GSC allows us to discard the assumptions of a converging forward activation vector length and still obtain a correspondence between gradient and correlation behavior.

\cite{widthHacking} propose to combat the exploding gradient problem by varying the width of intermediate layers. Such variation can indeed reduce the quadratic mean of gradient vector entries. However, our analysis in section \ref{depthexperimentssection} suggests that this technique is not effective in reducing the growth of the GSC. While we argue that an exploding GSC must cause training difficulties, to out knowledge, no such argument exists for a an exploding quadratic mean of gradient vector entries. In fact, our results suggest that width variation is ineffective at combating gradient pathology.

A second extension is that we show how both gradient explosion and domain bias can directly harm training. Neither is obvious. Gradient explosion might be a numerical quirk to be overcome by rescaling. Correlation information is a rather small part of the information present in the data, so losing that information via domain bias / order might be irrelevant. As a simple example, consider k-means. Performing k-means on an arbitrary dataset yields the same result as first adding a large constant to the data and then performing k-means, even though the addition destroys correlation information.

Thirdly, we demonstrate the importance of using different step sizes for different layers when comparing architectures. While \cite{meanFieldExplosion} show experimentally that architectures on the edge of chaos perform best at great depths, we obtain somewhat contrary evidence. Our two best performing vanilla architectures, SELU and layer-tanh, are both inside the chaotic regime whereas ReLU, layer-ReLU and tanh, which are all on the edge of chaos, exhibit a higher training classification error. Our chaotic architectures avoid pseudo-linearity and domain bias. The difference between our experiments and those in \cite{meanFieldExplosion} is that we allowed the step size to vary between layers. This had a significant impact, as can be seen in table 2.

Fourthly, \cite{meanFieldResNet} show that pathologies such as gradient explosion that arise in vanilla networks are reduced in specific ResNet architectures. We extend this finding to general ResNet architectures.

\subsection{ResNet from a dynamical systems view} \label{dynamicalSection}

Recently, \cite{layerSplitting1,antisymmResNet1,antisymmResNet2, layerSplitting2} proposed ResNet architectures inspired by dynamical systems and numerical methods for ordinary differential equations. A central claim is that these architectures achieve bounded forward activations and gradients (and hence GSC) as depth goes to infinity. They propose four practical strategies for building ResNet architectures: (a) ensuring that residual and skip blocks compute vectors orthogonal to each other by using e.g. skew-symmetric weight matrices (b) ensuring that the Jacobian of the residual block has eigenvalues with negative real part by using e.g. weight matrices factorized as $-C^TC$ (c) scaling each residual block by $1/B$ where $B$ is the number of blocks in the network and (d) regularizing weights in successive blocks to be similar via a fusion penalty.

\begin{table}[h]
\centering
{
\footnotesize
\begin{tabular}{ccc}
Architecture&$GSC(L+1,0)$ (base 10 log)&$GSC(L+1,0)$ dilution-corrected (base 10 log)\\ \hline \hline
batch-ReLU (i)&0.337&4.23\\
batch-ReLU (ii)&0.329&4.06\\
batch-ReLU (iii)&6.164&68.37\\
batch-ReLU (iv)&0.313&7.22\\
layer-tanh (i)&0.136&2.17\\
layer-tanh (ii)&0.114&1.91\\
layer-tanh (iii)&3.325&5.46\\
layer-tanh (iv)&0.143&2.31\\
\end{tabular}
}
\caption{Key metrics for architectures derived from dynamical systems theory.}
\label{GSCdynamic}
\end{table}

We evaluated those strategies empirically. In table \ref{GSCdynamic}, we show the value of $GSC(L+1,0)$ for 8 different architectures in their initialized state applied to Gaussian noise. All architectures use residual blocks containing a single normalization layer, a single nonlinearity layer and a single fully-connected linear layer. We initialize the linear layer in four different ways: (i) Gaussian initialization, (ii) skew-symmetric initialization, (iii) initialization as -$C^TC$ where C is Gaussian initialized and (iv) Gaussian initialization where weight matrices in successive blocks have correlation 0.5. Initializations (ii), (iii) and (iv) mimic strategies (a), (b) and (d) respectively. To enable the comparison of the four initialization styles, we normalize each weight matrix to have a unit qm norm. We study all four initializations for both batch-ReLU and layer-tanh. See section \ref{detailssection} for details.

Initialization (ii) reduces the gradient slightly relative to initialization (i). This is expected given theorem \ref{dilutionTheorem}. One of the key assumptions is that skip and residual block are orthogonal in expectation. While initialization (i) achieves this, under (ii), the two functions are orthogonal not just in expectation, but with probability 1. 

Initialization (iii) has gradients that grow much faster than initialization (i). On the one hand, this is surprising as \cite{antisymmResNet1} state that eigenvalues with negative real parts in the residual block Jacobian slow gradient growth. On the other hand, it is not surprising because introducing correlation between the residual and skip blocks breaks the conditions of theorem \ref{dilutionTheorem}. 

Initialization (iv) performs comparably to initialization (i) in reducing gradient growth, but requires a larger amount of dilution to achieve this result. Again, introducing correlation between successive blocks and thus between skip and residual blocks breaks the conditions of theorem \ref{dilutionTheorem} and weakens the power of dilution.

While we did not investigate the exact architectures proposed in \cite{antisymmResNet1,antisymmResNet2}, our results show that more theoretical and empirical evaluation is necessary to determine whether architectures based on (a), (b) and (d) are indeed capable of significantly improved forward activation and gradient stability. Of course, those architectures might still confer benefits in terms of e.g. inductive bias or regularization.

Finally, strategy (c), the scaling of either residual and/or skip blocks with constants is a technique already widely used in regular ResNets. In fact, our study suggests that in order to bound the GSC at arbitrary depth in a regular ResNet, it is sufficient to downscale each residual blocks by only $\frac{1}{\sqrt{B}}$ instead of $\frac{1}{B}$ as the dynamical systems papers suggest. 

\subsection{Exploding and vanishing gradients in RNNs} \label{RNNsection}

Exploding and vanishing gradients have been studied more extensively in the context of RNNs (e.g. \cite{recurrentPascanu,recurrentBengio}). It is important to note that the problem as it arises in RNNs is similar but also different from the exploding gradient problem in feedforward networks. The goal in RNNs is often to absorb information early on and store that information through many time steps and sometimes indefinitely. In the classical RNN architecture, signals acquired early would be subjected to a non-orthogonal transformation at every time step which leads to all the negative consequences described in this paper. LSTMs \citep{LSTM} and GRUs \citep{GRU}, which are the most popular solutions to exploding / vanishing gradients in RNNs, are capable of simply leaving each neuron that is considered part of the latent state completely unmodified from time step to time step by gating the incoming signal unless new information is received that is pertinent to that specific neuron. This solution does not apply in feedforward networks, because it is the very goal of each layer to modify the signal productively. Hence, managing exploding gradients in feedforward networks is arguably more difficult.

Nevertheless, there is similarity between LSTM and the orthogonal initial state because both eliminate non-orthogonality ``as much as possible''. LSTM can eliminate non-orthogonality completely from time step to time step whereas in the orthogonal initial state, non-orthogonality is eliminated only from the initial function. Again, viewing feedforward networks as ensembles of shallower networks, orthogonal initial functions ensure that information extracted from each ensemble member does not have to pass through non-orthogonal transformations without clear reason. This is precisely what LSTM attempts to achieve.

\subsection{Open research questions and future work} \label{futureWorkSection}

\paragraph{Biases, convolutional and recurrent layers}

In this paper, we focus our analysis on MLPs without trainable bias and variance parameters. Theorem \ref{depthTheorem}, in its formulation, applies only to such MLPs. The other theorems use conditions that are potentially harder to fulfill, even approximately, in non-MLP architectures. Our experimental evaluation is limited to MLPs.

We think that results very similar to those presented in this paper are acheivable for other types of neural networks, such as those containing trainable biases, convolutional layers or recurrent layers, although we suspect the gap between theory and practice may increase.

Analysis of deep gradients has so far focused on MLPs (e.g. \cite{shattering,meanFieldExplosion,meanFieldResNet, orthogonalInitialization}), so a principled extension of these results to other network types would break new and important ground.

\paragraph{Understanding collapsing domains}

It is difficult to assess or measure the degree to which the domain collapses in a given network. Neither entropy nor exponential entropy can be computed directly. How should we evaluate domains that are composed of sub-spaces that have varying intrinsic dimensionality? 

A domain can collapse in many different ways. For example, in a deep linear, Gaussian-initialized network, the domain collapses onto the line through the principal eigenvector of the product of weight matrices, but never onto a single point. In a ReLU network, the domain collapses onto a ray from the origin. In layer-ReLU, the normalization operation then collapses the domain onto a single point. In a tanh network with very large weights, each tanh layer collapses the domain onto the corners of the hypercube. In what other ways can a collapsing domain manifest? How can those manifestations harm training?

\paragraph{What gradient scale is best?} $GSC(1,L+1)$ indicates the relative responsiveness of the prediction layer with respect to changes in the input layer. Of course, the goal in deep learning, at least within a prediction setting, is to model some ground truth function $f_\text{truth}$ that maps data inputs to true labels. That function has itself a GSC at each input location $X$ that measures the relative responsiveness of $f_\text{truth}(X)$ to changes in $X$. If the network is to perfectly represent the ground truth function, the GSCs would also have to match up. If, on the other hand, the GSC of the network differs significantly from that of $f_\text{truth}$, the network is not fitting $f_\text{truth}$ well. This suggests that in fact, the ``best'' value of the GSC is one that matches that of the ground truth. If the GSC of the network is too low, we may experience underfitting. If the GSC of the network is too high, we may experience overfitting.


\paragraph{How to achieve the ``right'' gradient?}

To model the ground truth function, we may not just want to consider the overall magnitude of the GSC across the dataset, but to enable the network to have gradients of different magnitudes from one data input to the next; or to learn highly structured gradients. For example, given an image of a dog standing in a meadow, we might desire a high gradient with respect to pixels signifying e.g. facial features of the dog but a low gradient with respect to pixels that make up the meadow, and a uniformly low gradient given an image of a meadow. Such gradients would be very valuable not just in modelling real world functions more accurately and improving generalization, but in making the output of neural networks more explainable and avoiding susceptibility to attacks with adversarial inputs.

\paragraph{What is the relationship between compositional depth, effective depth, linear approximation error, dilution, gradient scale and representational capacity?}

Throughout this paper, we have discussed various metrics that can influence the performance of deep networks. We proved and discussed many relationships between these metrics. However, there are still many open questions regarding how these concepts interrelate. Is effective depth truly a better tool for measuring ``depth'' than compositional depth? Does depth provide additional modeling benefits beyond its power to exponentially increase gradient scale? Is there a reason to prefer a deeper network if its gradient scale is the same as a shallower network? Is there a reason to prefer a network with higher linear approximation error if its gradient scale is the same as that of a network with lower linear approximation error? Does dilution bring about harms or benefits independently of its impact on gradient scale?

\paragraph{How far does the orthogonal initial state take us?} An orthogonal initial state reduces gradients via dilution, which allows for relatively larger updates, which enables increased growth of residual functions, which allows for greater effective depth. However, as residual functions grow, dilution decreases, so the gradient increases, so updates must shrink, so the growth of residual functions slows, so the growth of effective depth slows.

In other words, for the network to become deeper, it needs to be shallow.

Therefore, while training from an orthogonal initial state can increase effective depth, we expect this effect to be limited. Additional techniques could be required to learn functions which require a compositional representation beyond this limit.

\section{Further terminology, notation and conventions} \label{terminologySection}

\begin{itemize}
\item $x$ and $y$ are generally used to refer to the components of a datapoint. Then, we have $(x,y) \in D$.
\item $X$ generally refers to a vector of dimensionality $d$, i.e. the same dimensionality as the $x$ component of datapoints. Similarly, $Y$ refers to an element of the domain of possible labels. We refer to $x$ and $X$ as `data input' .
\item $F_l$ refers to a vector of dimensionality $d_l$, i.e. the same dimensionality as $f_l$.
\item  We write $f_l(\theta, x)$ as a short form of $f_l(\theta_l, f_{l+1}(..(f_L(\theta_L, x))..))$. Sometimes, we omit $x$ and / or $\theta$. In that case, $x$ and / or $\theta$ remain implicit. $f_l(\theta, X)$ is an analogous short form. 
\item We write $f_l(\theta, f_k)$ as a short form of $f_l(\theta_l, f_{l+1}(..f_{k-1}(\theta_{k-1},f_{k})..))$. Sometimes, we omit $\theta$. In that case, $\theta$ remains implicit. $f_l(\theta, F_k)$ is an analogous short form.
\item We use $f_{L+1}$, $i_{L+1}$ and $F_{L+1}$ interchangeably with $x$ or $X$.
\item We say a random vector is `radially symmetric' if its length is independent of its orientation and its orientation is uniformly distributed.
\item We say a random matrix is `Gaussian initialized' if its entries are drawn independently from a mean zero Gaussian distribution.
\item We say an $m*n$ random matrix is `orthogonally initialized' if it is a fixed multiple of an $m*n$ submatrix of a $\max(m,n)*\max(m,n)$ uniformly random orthogonal matrix.
\item We use parentheses $()$ to denote vector and matrix elements, i.e. $A(3,4)$ is the fourth element in the third row of the matrix $A$.
\item Throughout sections \ref{propProofSection} and \ref{theoProofSection}, we assume implicitly that the GSC is defined and thus that neural networks are differentiable. All results can be trivially extended to cover networks that are almost surely differentiable and directionally differentiable everywhere, which includes SELU and ReLU networks. 
\item We discuss the conditions that arise in theoretical results only in the context of MLPs. Note that several of our theoretical results apply to varying degrees to non-MLPs. We will not discuss the degree of applicability.
\end{itemize}

\section{Effective depth: details} \label{depthDetailsSection}

\subsection{Formal definition} \label{depthDefinitionSection}

Let a `gradient-based algorithm' for training a mutable parameter vector $\theta$ from an initial value $\theta^{(0)}$ for a network $f$ be defined as a black box that is able to query the gradient $\frac{df(\theta,X,Y)}{d\theta}$ at arbitrary query points $(X,Y)$ but only at the current value of the mutable parameter vector $\theta$. It is able to generate updates $\Delta \theta$ which are added to the mutable parameter vector $\theta$. Let the sequence of updates be denoted as $\Delta \theta^{(1)}, \Delta \theta^{(2)}, ..$. We define the successive states of $\theta$ recursively as $\theta^{(t)} = \theta^{(t-1)} + \Delta \theta^{(t)}$. For simplicity, assume the algorithm is deterministic. 

In a residual network defined according to equation \ref{residualForm}, we can write the gradient with respect to a parameter sub-vector as $\frac{df(\theta,X,Y)}{d\theta_l} = \frac{df_0}{df_1}\frac{df_1}{df_2} .. \frac{df_{l-1}}{df_l}\frac{df_l}{d\theta_l} = \frac{df_0}{df_1}(\frac{di_1}{df_2} + \frac{dr_1}{df_2}) .. (\frac{di_{l-1}}{df_l} + \frac{dr_{l-1}}{df_l})\frac{df_l}{d\theta_l}$. Multiplying this out, we obtain $2^{l-1}$ terms. We call a term `$\lambda$-residual' if it contains $\lambda$ or more Jacobians of residual functions, as opposed to Jacobians of initial functions. Let $res^\lambda_l(f,\theta,X,Y)$ be the sum of all $\lambda$-residual terms in $\frac{df(\theta,X,Y)}{d\theta_l}$.

Now consider two scenarios. In scenario (1), when the algorithm queries the gradient, it receives $\{\frac{df(\theta,X,Y)}{d\theta_1},\frac{df(\theta,X,Y)}{d\theta_2},..,\frac{df(\theta,X,Y)}{d\theta_L}\}$ i.e. the ``regular'' gradient. In scenario (2), it receives $\{\frac{df(\theta,X,Y)}{d\theta_1} - res^\lambda_1(f,\theta,X,Y),\frac{df(\theta,X,Y)}{d\theta_2} - res^\lambda_2(f,\theta,X,Y),..,\frac{df(\theta,X,Y)}{d\theta_L} - res^\lambda_L(f,\theta,X,Y)\}$, i.e. a version of the gradient where all $\lambda$-residual terms are removed. Let the parameter vector attain states $\theta^{(1)}$, $\theta^{(2)}$, .. in scenario (1) and $\theta^{(1,\lambda)}$, $\theta^{(2,\lambda)}$, .. in scenario (2). Then we say the `$\lambda$-contribution' at time $t$ is $\theta^{(t)}-\theta^{(t,\lambda)}$. Finally, we say the `effective depth at time $t$ with threshold $h$' is the largest $\lambda$ such that there exists an $l$ with $||\theta_l^{(t)}-\theta_l^{(t,\lambda)}||_2 \ge h$, plus one. We add one because we include the residual function at layer $l$, which is co-adapting to the residual Jacobians contained in the gradient term.

There is no objectively correct value for the threshold $h$. In practice, we find that the $\lambda$-contribution decreases quickly when $\lambda$ is increased beyond a certain point. Hence, the exact value of $h$ is not important when comparing different networks by effective depth.

The impact that the shift $\theta_l^{(t)}-\theta_l^{(t,\lambda)}$ has on the output of the network is influenced by the scale of $\theta_l^{(t)}$ as well as $GSC(l,0)$. If those values vary enormously between layers or architectures, it may be advisable to set different thresholds for different layers or architectures, though we did not find this necessary.

\subsection{Computational estimate} \label{depthComputationSection}

Unfortunately, computing the effective depth measure is intractable as it would require computing exponentially many gradient terms. In this section, we explain how we estimate effective depth in our experiments. 

In this paper, we train networks only by stochastic gradient descent with either a single step size for all layers or a custom step size for each layer. Our algorithm for computing effective depth assumes this training algorithm.

\paragraph{Vanilla networks} Assume that the network is expressed as a residual network as in equation \ref{residualForm}. Let $B_t$ be the batch size for the $t$'th update, let $c_l^{(t)}$ be the step size used at layer $l$ for the $t$'th update and let $((X^{(t,1)},Y^{(t,1)}),(X^{(t,2)},Y^{(t,2)}), .., (X^{(t,B_t)},Y^{(t,B_t)}))$ be the batch of query points used to compute the $t$'th update. Then SGD computes for all $l$

\begin{eqnarray*}
\Delta \theta_l^{(t)} &=& c_l^{(t)} \sum_{b=1}^{B_t}  \frac{df(\theta^{(t-1)},X^{(t,b)},Y^{(t,b)})}{d\theta^{(t-1)}_l}\\
\theta_l^{(t)} &=& \theta_l^{(t-1)} + \Delta \theta_l^{(t)}
\end{eqnarray*}

\begin{algorithm}
\CommentSty{\color{blue}}
\SetKwInOut{Input}{input}
$arr := [1]$\;
\For{$k=0$ \KwTo $l-1$}
{
	$size = size(arr)$\;
	$arr.\text{push\_back}(arr[size-1]*||r_k||_{op})$\;
	\For{$i=size-1$ \KwTo $1$}
	{
		$arr[i] = arr[i] * \frac{||\frac{df}{df_{k+1}}||_2}{||\frac{df}{df_{k}}||_2} + arr[i-1]*||r_k||_{op}$\;
	}
	$arr[0] = arr[0] *  \frac{||\frac{df}{df_{k+1}}||_2}{||\frac{df}{df_{k}}||_2}$\;
}
$out = 0$\;
\For{$i=\lambda$ \KwTo $size(arr) - 1$}
{
	$out = out + arr[i]$\;
}
\Return $out*c_l^{(t)}*||f_{l+1}||_2$\;
\caption{Estimating the $\lambda$-contribution at some layer $l$ for a given batch and query point.} \label{contribVanilla}
\end{algorithm}

For any update $t$ and query point $b$, we estimate its $\lambda$-contribution at layer $l$ as in algorithm \ref{contribVanilla}. For unparametrized layers, $||r_k||_{op}$ is set to zero. For linear layers, it is the operator norm of the residual weight matrix. The final estimate of the length of the $\lambda$-contribution at layer $l$ for the entire training period is then simply the sum of the lengths of the estimated $\lambda$-contributions over all time points and query points.

The core assumption here is that applying the Jacobian of an initial or residual function of a given layer will increase the lengths of all terms approximately equally, no matter how many residual Jacobians they contain. In other words, we assume that in $\lambda$-residual terms, the large singular values of layer-wise Jacobians do not compound disproportionately compared to other terms. This is similar to the core assumption in theorem \ref{depthTheorem} in section \ref{theo1proof}.

We conservatively bound how a Jacobian of an initial function will increase the length of a term with the impact of the Jacobian of the entire layer, i.e. $ \frac{||\frac{df}{df_{k+1}}||_2}{||\frac{df}{df_{k}}||_2}$. We use $||r_k||_{op}$ as a conservative estimate on how a residual Jacobian will increase the length of a term. 

We use the sum of the lengths of all $\lambda$-residual terms in a batch as a conservative bound of the length of the $\lambda$-contribution of the batch. In essence, we assume that all $\lambda$-residual terms have the same orientation.

Finally, we use the sum of the lengths of the $\lambda$-contributions within each update as an estimate of the length of the total $\lambda$-contribution of the entire training period. On the one hand, this is conservative as we implicitly assume that the $\lambda$-contributions of each batch have the same orientation. On the other hand, we ignore indirect effects that $\lambda$-contributions in early batches have on the trajectory of the parameter value and hence on $\lambda$-contributions of later batches. Since we are ultimately interested in effective depth, we can ignore these second-order effects as they are negligible when the total $\lambda$-contribution is close to a small threshold $h$.

Overall, we expect that our estimate of the effective depth (e.g. figure \ref{dyna}D) is larger than its actual value. This hypothesis is bolstered by the robustness of some of our trained networks to Taylor expansion (see figure \ref{dyna}F).

\paragraph{ResNet}

For ResNet architectures, we need to tweak our estimate of effective depth to take into account skip connections. In algorithm \ref{contribResNet}, we detail how the variable $arr$ is modified as it crosses a block. We write $f_n(f_m) = s_n(f_m) + \rho_n(f_m)$, where $f_n$ is a layer at which some skip connection terminates, $f_m$ is the layer at which the skip connection begins, $s_n$ is the function computed by the skip block and $\rho_n(f_m) = \rho_n(f_{n+1}(..f_{m-1}(f_m)..))$ is the function computed by the residual block. We write $f_k = i_k + r_k$ for $n+1 \le k \le m-1$ and $\rho_n = i_n + r_n$, i.e. we break down each layer in the residual block into an initial function and a residual function.

\begin{algorithm}
\CommentSty{\color{blue}}
\SetKwInOut{Input}{input}
$arr_\text{copy} = arr$\;
\For{$k=n$ \KwTo $m-1$}
{
	$size = size(arr)$\;
	$arr.\text{push\_back}(arr[size-1]*||r_k||_{op})$\;
	\For{$i=size-1$ \KwTo $1$}
	{
		$arr[i] = arr[i] * \frac{||\frac{df}{d\rho_n}\frac{d\rho_n}{df_{k+1}}||_2}{||\frac{df}{df_{k}}||_2} + arr[i-1]*||r_k||_{op}$\;
	}
	$arr[0] = arr[0] *  \frac{||\frac{df}{d\rho_n}\frac{d\rho_n}{df_{k+1}}||_2}{||\frac{df}{df_{k}}||_2}$\;
}
\For{$i=0$ \KwTo $size(arr_\text{copy}) - 1$}
{
	$arr[i] = arr[i] + arr_\text{copy}[i]*\frac{||\frac{df}{df_m}||_2 - ||\frac{df}{d\rho_n}\frac{d\rho_n}{df_m}||_2}{||\frac{df}{df_n}||_2}$\; \label{l11}
}
\caption{Modifying the $arr$ variable as it crosses a block. This algorithm is nested inside algorithm \ref{contribVanilla}.} \label{contribResNet}
\end{algorithm}

In line \ref{l11}, the combined effect of the skip connection and the initial functions of the residual block is approximated by the effect of the entire block, i.e. $\frac{||\frac{df}{df_m}||_2}{||\frac{df}{df_n}||_2}$. In the same line, we must subtract the impact of the initial functions accumulated while passing through the residual block, i.e. $\frac{- ||\frac{df}{d\rho_n}\frac{d\rho_n}{df_m}||_2}{||\frac{df}{df_n}||_2}$. The impact of the residual functions in the block is unaffected by the skip connection and bounded by the operator norm, as before.

\subsection{Discussion}

The effective depth measure has several limitations. 

One can train a linear MLP to have effective depth much larger than 1, but the result will still be equivalent to a depth 1 network.

Consider the following training algorithm: first randomly re-sample the weights, then apply gradient descent. Clearly, this algorithm is equivalent to just running gradient descent in any meaningful sense. The re-sampling step nonetheless blows up the residual functions so as to significantly increase effective depth. 

The effective depth measure is very susceptible to the initial step size. In our experiments, we found that starting off with unnecessarily large step sizes, even if those step sizes were later reduced, lead to worse outcomes. However, because of the inflating impact on the residual function, the effective depth would be much higher nonetheless.

Effective depth may change depending on how layers are defined. In a ReLU MLP, for example, instead of considering a linear transformation and the following ReLU operation as different layers, we may define them to be part of the same layer. While the function computed by the network and the course of gradient-based training do not depend on such redefinition, effective depth can be susceptible to such changes.

\section{Propositions and proofs} \label{propProofSection}

\subsection{Proposition \ref{equivalentprop}} \label{prop1proof}

\begin{repproposition}{equivalentprop}

Given:

\begin{itemize}
\item a neural network $f$ of nominal depth $L$
\item an initial parameter value $\theta^{(0)}$
\item a mutable parameter value $\theta$ that can take values in some closed, bounded domain $\Theta$
\item a finite dataset $D$ of datapoints $(x,y)$
\item a closed, bounded domain $\mathcal{D}$ of possible query points $(X,Y)$
\item a function $||.||$ from matrices to the reals that has $c||.|| = ||c.||$ and $||.|| \ge 0$
\item some deterministic algorithm that is able to query gradients of $f$ at the current parameter value and at query points in $\mathcal{D}$ and that is able to apply updates $\Delta \theta$ to the parameter value
\item constant $r'$
\end{itemize}

Assume:

\begin{itemize}
\item Running the algorithm on $f$ with $\theta$ initialized to $\theta^{(0)}$ for a certain number of updates $T$ causes $\theta$ to attain a value $\hat{\theta}$ at which $f$ attains some error value $E_{\text{final}}$ on $D$.
\item At every triplet $(\theta,X,Y) \in \Theta \times \mathcal{D}$, we have $||\mathcal{J}^l_k|| \neq 0$ for all $0 \le l \le k \le L$ and $||\mathcal{T}^l_k|| \neq 0$ for all $0 \le l \le k \le L$ where $\theta_k$ is non-empty.
\end{itemize}

Then we can specify some other neural network $f'$ and some other initial parameter value $\theta'^{(0)}$ such that the following claims hold:

\begin{enumerate}
\item $f'$ has nominal depth $L$ and the same compositional depth as $f$.
\item The algorithm can be used to compute $T$ updates by querying gradients of $f'$ at the current parameter value and at query points in $\mathcal{D}$ which cause $\theta$ to attain a value $\hat{\theta}'$ where $f'$ attains error $E_{\text{final}}$ on $D$ and makes the same predictions as $f(\hat{\theta})$ on $D$.
\item At every triplet $(\theta,X,Y) \in \Theta \times \mathcal{D}$, we have $||\mathcal{T'}^l_k|| \ge r'^{k-l}$ for all $0 \le l \le k \le L$ where $\theta_k$ is non-empty and $||\mathcal{J'}^l_k|| \ge r'^{k-l}$ for all $0 \le l \le k \le L$ except $(k,l) = (1,0)$.
\end{enumerate}

\end{repproposition}

\begin{JMLRproof}
Throughout this proof, for simplicity, we assume all $\theta_k$ are non-empty. The case where some $\theta_k$ are empty follows trivially. Since $\Theta$ and $\mathcal{D}$ are closed and bounded, so is $\Theta \times \mathcal{D}$. Therefore for all $0 \le l \le k \le L$, both $||\mathcal{J}^l_k||$ and $||\mathcal{T}^l_k||$ attain their infimum on that domain if it exists. $||.||$ is non-negative, so the infimum exists. $||.||$ is non-zero on the domain, so the infimum, and therefore the minimum, is positive. Since $f$ has finite depth, there is an $r > 0$ such that for all tuplets $(\theta, X, Y, k, l)$, we have $||\mathcal{J}^l_k|| \ge r^{k-l}$ and $||\mathcal{T}^l_k|| \ge r^{k-l}$. Let $R = \frac{r'}{r}$.

Now, we define $f'$ via its layer functions.

\begin{eqnarray*}
f'_0 &=& f_0\\
f'_1(\theta_1,F_2) &=& f_1(R\theta_1,R^2F_2)\\
f'_l(\theta_l,F_{l+1}) &=& R^{-l}f_l(R^l\theta_l,R^{l+1}F_{l+1}) \text{ for } 2 \le l \le L-1\\
f'_L(\theta_L,X) &=& R^{-L}f_L(R^L\theta_L,X)
\end{eqnarray*}

$f$ and $f'$ clearly have the same nominal and compositional depth, so claim (1) holds. Given any vector $v$ with $L$ sub-vectors, define the transformation $R(v)$ as $R(v)_l = R^lv_l$. Finally, we set $\theta'^{(0)} := R^{-1}(\theta^{(0)})$. 

We use the algorithm to train $f'$ as follows. Whenever the algorithm queries some gradient value $\frac{df'}{d\theta}$, we instead submit to it the value $R^{-1}(\frac{df'}{d\theta})$. Whenever the algorithm wants to apply an update $\Delta \theta$ to the parameter, we instead apply $R^{-1}(\Delta \theta)$. Let $S'^{(t)}$, $0 \le t \le T$ be the state of the system after applying $t$ updates to $\theta$ under this training procedure. Let $S^{(t)}$, $0 \le t \le T$ be the state of the system after applying $t$ updates to $\theta$ when the algorithm is run on $f$. Then the following invariances hold.

\begin{enumerate}[label=\Alph*]
\item $\theta'^{(t)} = R^{-1}(\theta^{(t)})$, where $\theta'^{(t)}$ is the value of $\theta$ under $S'^{(t)}$ and $\theta^{(t)}$ is the value of $\theta$ under $S^{(t)}$.
\item $f$ makes the same predictions and attains the same error on $D$ under $S^{(t)}$ as $f'$ under $S'^{(t)}$.
\item Any state the algorithm maintains is equal under both $S^{(t)}$ and $S'^{(t)}$.
\end{enumerate}

We will show these by induction. At time $t=0$, we have $\theta'^{(0)} = R^{-1}(\theta^{(0)})$ as chosen, so (A) holds. It is easy to check that (B) follows from (A). Since the algorithm has thus far not made any queries and thus received no external inputs, (C) also holds.

Now for the induction step. Assuming that $\theta'^{(t)} = R^{-1}(\theta^{(t)})$, it is easy to check that $\frac{df'(\theta'^{(t)})}{d\theta} = R(\frac{df(\theta^{(t)})}{d\theta})$. Therefore, whenever the algorithm queries a gradient of $f'$, it will receive $R^{-1}(\frac{df'(\theta'^{(t)})}{d\theta}) = R^{-1}(R(\frac{df(\theta^{(t)})}{d\theta})) = \frac{df(\theta^{(t)})}{d\theta}$. Therefore, the algorithm receives the same inputs under both $S^{(t)}$ and $S'^{(t)}$. Since the internal state of the algorithm is also the same, and the algorithm is deterministic, the update returned by the algorithm is also the same and so is the internal state after the update is returned, which completes the induction step for (C). Because the algorithm returns the same update in both cases, after the prescribed post-processing of the update under $f'$, we have $\Delta \theta'^{(t)}= R^{-1}(\Delta \theta^{(t)})$. Therefore $\theta'^{(t+1)} = \theta'^{(t)} + \Delta \theta'^{(t)} = R^{-1}(\theta^{(t)}) + R^{-1}(\Delta \theta^{(t)}) = R^{-1}(\theta^{(t)} + \Delta \theta^{(t)}) = R^{-1}(\theta^{(t+1)})$. This completes the induction step for (A) and again, (B) follows easily from (A).

(B) implies directly that claim (2) holds. Finally, for any tuplet $(\theta,X,Y,k,l)$, we have $||\mathcal{T'}^l_k|| = ||\frac{df'_l(\theta'^{(t)})}{d\theta_k}|| = ||R^{k-l}\frac{df_l(\theta^{(t)})}{d\theta_k}|| \ge R^{k-l}r^{k-l} = r'^{k-l}$ and unless $(k,l) = (1,0)$ we have $||\mathcal{J'}^l_k|| = ||\frac{df'_l(\theta'^{(t)})}{df'_k}|| = ||R^{k-l}\frac{df_l(\theta^{(t)})}{df_k}|| \ge R^{k-l}r^{k-l} = r'^{k-l}$. Therefore, claim (3) also holds, which completes the proof.

\end{JMLRproof}

Notes:

\begin{itemize}
\item The condition that the Jacobians of $f$ always have non-zero norms may be unrealistic. For practical purposes, it is enough to have Jacobians that have non-zero norms almost always. We can then choose an arbitrary threshold and obtain a network $f'$ that has exploding Jacobians at $\theta'$ whenever $f$ has Jacobians of size above this threshold at $R(\theta')$. (The choice of $R$ depends on this threshold.)
\item Claim (3) of the proposition does not include the case $(k,l) = (1,0)$ and it does not include Jacobians with respect to the input $X$. These Jacobians have to be the same between $f$ and $f'$ if we require $f'$ to have the same error and predictions as $f$. However, if we are ok with multiplicatively scaled errors and predictions, claim (3) can be extended to cover those two cases. Scaled training errors and predictions are generally not a problem in e.g. classification.
\item Note that not only does the algorithm achieve the same predictions in the same number of updates for both $f$ and $f'$, but the computation conducted by the algorithm is also identical, so $f'$ is as ``easy to train'' as $f$ no matter how we choose to quantify this as long as we know to apply the scaling transformation.
\item There are no constraints on the explosion rate $r'$. If we can successfully train a network with some explosion rate, we can successfully train an equivalent network with an arbitrary explosion rate.
\item $f'$ is very similar to $f$, so this proposition can be used to construct trainable networks with exploding Jacobians of any type and depth as long as there exists some trainable network of that type and depth.
\item The proposition can be easily extended to non-deterministic algorithms by using distributions and expectations.
\end{itemize}

\subsection{Proposition \ref{vectorEffect}} \label{propb21proof}

\begin{repproposition}{vectorEffect}
Let $A$ be an $m \times n$ matrix and $u$ a uniformly distributed, $n-dimensional$ unit length vector. Then $||A||_{qm} = \mathbb{Q}_{u}||Au||_2$. 
\end{repproposition}

\begin{JMLRproof}
We use $L\Sigma R^T$ to denote the singular value decomposition and $s_i$ to denote singular values.

\begin{eqnarray*}
&&\mathbb{Q}_{u}||Au||_2\\
&=& \mathbb{Q}_{u}||L\Sigma R^Tu||_2\\
&=& \mathbb{Q}_{u}||\Sigma u||_2\\
&=&\mathbb{Q}_{u}\sqrt{\sum_{i=1}^{\min(m,n)} s_i^2u(i)^2}\\
&=&\sqrt{\mathbb{E}_{u}\sum_{i=1}^{\min(m,n)} s_i^2u(i)^2}\\
&=&\sqrt{\sum_{i=1}^{\min(m,n)} s_i^2\mathbb{E}_{u}u(i)^2}\\
&=&\sqrt{\sum_{i=1}^{\min(m,n)} s_i^2\mathbb{E}_{u}(\frac{1}{n}\sum_{j=1}^n u(j)^2)}\\
&=&\sqrt{\sum_{i=1}^{\min(m,n)} s_i^2\mathbb{E}_{u}(\frac{1}{n})}\\
&=&\sqrt{\frac{\sum_{i=1}^{\min(m,n)} s_i^2}{n}}\\
&=& ||A||_{qm}
\end{eqnarray*}

\end{JMLRproof}

\subsection{Proposition \ref{L2relation}} \label{propb22proof}

\begin{repproposition}{L2relation}
Let $A$ be an $m \times n$ matrix. Then $||A||_{qm} = \frac{1}{\sqrt{n}}||A||_2$.
\end{repproposition}

\begin{JMLRproof}
Let $u$ be a uniformly distributed, $n$-dimensional unit length vector.

\begin{eqnarray*}
&&||A||_{qm}\\
&=&\mathbb{Q}_{u}||Au||_2\\
&=&\mathbb{Q}_{u}\sqrt{\sum_{i=1}^m (\sum_{j=1}^n A(i,j)u(j))^2}\\
&=&\sqrt{\mathbb{E}_{u}\sum_{i=1}^m (\sum_{j=1}^n A(i,j)u(j))^2}\\
&=&\sqrt{\mathbb{E}_{u}\big(\sum_{i=1}^m \sum_{j=1}^n A(i,j)^2u(j)^2 + 2\sum_{i=1}^m \sum_{j=1}^n\sum_{k=1,k\neq j}^n A(i,j)u(j)A(i,k)u(k)\big)}\\
&=&\sqrt{\sum_{i=1}^m \sum_{j=1}^n A(i,j)^2\mathbb{E}_{u}u(j)^2 + 2\sum_{i=1}^m \sum_{j=1}^n\sum_{k=1,k\neq j}^n A(i,j)A(i,k)\mathbb{E}_{u}u(j)u(k)}\\
&=&\sqrt{\sum_{i=1}^m \sum_{j=1}^n A(i,j)^2\mathbb{E}_{u}u(j)^2}\\
&=&\sqrt{\sum_{i=1}^m \sum_{j=1}^n A(i,j)^2\mathbb{E}_{u}(\frac{1}{n}\sum_{k=1}^n u(k)^2)}\\
&=&\frac{1}{\sqrt{n}}\sqrt{\sum_{i=1}^m \sum_{j=1}^n A(i,j)^2}\\
&=&\frac{1}{\sqrt{n}}||A||_2
\end{eqnarray*}

\end{JMLRproof}

\subsection{Proposition \ref{relrel}} \label{prop2proof}

\begin{repproposition}{relrel}

Let $u$ be a uniformly distributed, $d_k$-dimensional unit length vector. Then \begin{equation*}GSC(k,l) = \lim_{\epsilon \to 0} \mathbb{Q}_u\frac{\frac{||f_l(f_k+\epsilon u) - f_l(f_k)||_2}{||f_l(f_k)||_2}}{\frac{||f_k+\epsilon u - f_k||_2}{||f_k||_2}}\end{equation*}
\end{repproposition}

\begin{JMLRproof}

\begin{eqnarray*}
&& \lim_{\epsilon \to 0} \mathbb{Q}_{u}\frac{\frac{||f_l(f_k+\epsilon u) - f_l(f_k)||_2}{||f_l(f_k)||_2}}{\frac{||f_k+\epsilon u - f_k||_2}{||f_k||_2}}\\
&=&\lim_{\epsilon \to 0} \mathbb{Q}_{u}\frac{||f_l(f_k+\epsilon u) - f_l(f_k)||_2||f_k||_2}{\epsilon||f_l(f_k)||_2}\\
&=&\lim_{\epsilon \to 0} \frac{||f_k||_2}{\epsilon||f_l(f_k)||_2}\mathbb{Q}_{u}||f_l(f_k+\epsilon u) - f_l(f_k)|||_2\\
&=&\lim_{\epsilon \to 0} \frac{||f_k||_2}{\epsilon||f_l(f_k)||_2}\mathbb{Q}_{u}||f_l(f_k) + \epsilon\mathcal{J}^l_ku + O(\epsilon^2) - f_l(f_k)||_2\\
&=&\lim_{\epsilon \to 0} \frac{||f_k||_2}{||f_l(f_k)||_2}\mathbb{Q}_{u}||\mathcal{J}^l_ku + \frac{O(\epsilon^2)}{e}||_2\\
&=&\frac{||f_k||_2}{||f_l(f_k)||_2}\mathbb{Q}_{u}||\mathcal{J}^l_ku||_2\\
&=& \frac{||f_k||_2}{||f_l(f_k)||_2}||\mathcal{J}^l_k||_{qm}\\
&=& GSC(k,l)
\end{eqnarray*}

\end{JMLRproof}

\pagebreak

\subsection{Proposition \ref{relrelparm}} \label{prop3proof}

\begin{repproposition}{relrelparm}

Let $u$ be a uniformly distributed, $d_kd_{k+1}$-dimensional unit length vector. Assume $f_k$ is a fully-connected linear layer without trainable bias parameters and $\theta_k$ contains the entries of the weight matrix. Then \begin{equation*}GSC(k,l)\frac{||\theta_k||_2||f_{k+1}||_2}{||f_k||_2\sqrt{d_{k+1}}} = \lim_{\epsilon \to 0} \mathbb{Q}_u \frac{\frac{||f_l(f_k(\theta_k+\epsilon u,f_{k+1})) - f_l(f_k(\theta_k,f_{k+1}))||_2}{||f_l||_2}}{\frac{||\theta_k+\epsilon u-\theta_k||_2}{||\theta_k||_2}}\end{equation*}

Further, if $\theta_k$ is random and radially symmetric, we have \begin{equation*}\mathbb{Q}^{-1}_{\theta_k}\frac{||\theta_k||_2||f_{k+1}||_2}{||f_k||_2\sqrt{d_{k+1}}} = 1\end{equation*}

\end{repproposition}

\begin{JMLRproof}
Throughout this proof, we will use $\theta_k$ to refer to both the parameter sub-vector and the weight matrix. Similarly, we will use $\epsilon u$ to refer to both a perturbation of the parameter sub-vector and of the weight matrix. We use $L\Sigma R^T$ to denote the singular value decomposition and $s_i$ to denote singular values.

\begin{eqnarray*}
&& \lim_{\epsilon \to 0}\mathbb{Q}_{u} \frac{\frac{||f_l(f_k(\theta_k+\epsilon u,f_{k+1})) - f_l(f_k(\theta_k,f_{k+1}))||_2}{||f_l||_2}}{\frac{||\theta_k+\epsilon u-\theta_k||_2}{||\theta_k||_2}}\\
&=& \lim_{\epsilon \to 0}\frac{||\theta_k||_2}{\epsilon||f_l||_2}\mathbb{Q}_{u} ||f_l(f_k(\theta_k+\epsilon u,f_{k+1})) - f_l(f_k(\theta_k,f_{k+1}))||_2\\
&=& \lim_{\epsilon \to 0}\frac{||\theta_k||_2}{\epsilon||f_l||_2}\mathbb{Q}_{u} ||f_l((\theta_k+\epsilon u)f_{k+1}) - f_l(\theta_kf_{k+1})||_2\\
&=& \lim_{\epsilon \to 0}\frac{||\theta_k||_2}{\epsilon||f_l||_2}\mathbb{Q}_{u} ||f_l(\theta_kf_{k+1}+\epsilon uf_{k+1}) - f_l(\theta_kf_{k+1})||_2\\
&=& \lim_{\epsilon \to 0}\frac{||\theta_k||_2}{\epsilon||f_l||_2}\mathbb{Q}_{u} ||f_l(\theta_kf_{k+1})+\mathcal{J}^l_k\epsilon uf_{k+1} + O(\epsilon^2) - f_l(\theta_kf_{k+1})||_2\\
&=& \lim_{\epsilon \to 0}\frac{||\theta_k||_2}{||f_l||_2}\mathbb{Q}_{u} ||\mathcal{J}^l_k uf_{k+1} + \frac{O(\epsilon^2)}{\epsilon}||_2\\
&=& \frac{||\theta_k||_2}{||f_l||_2}\mathbb{Q}_{u} ||\mathcal{J}^l_k uf_{k+1}||_2\\
&=& \frac{||\theta_k||_2}{||f_l||_2}\mathbb{Q}_{u} ||L\Sigma R^Tuf_{k+1}||_2\\
&=& \frac{||\theta_k||_2}{||f_l||_2}\mathbb{Q}_{u} ||\Sigma uf_{k+1}||_2\\
&=& \frac{||\theta_k||_2}{||f_l||_2}\mathbb{Q}_{u} \sqrt{\sum_{i=1}^{\min(d_k,d_l)}(\sum_{j=1}^{d_{k+1}} s_iu(i,j)f_{k+1}(j))^2}\\
&=& \frac{||\theta_k||_2}{||f_l||_2} \sqrt{\mathbb{E}_{u}\sum_{i=1}^{\min(d_k,d_l)}(\sum_{j=1}^{d_{k+1}} s_iu(i,j)f_{k+1}(j))^2}\\
&=& \frac{||\theta_k||_2}{||f_l||_2} \sqrt{\mathbb{E}_{u}\sum_{i=1}^{\min(d_k,d_l)}\sum_{j=1}^{d_{k+1}}\sum_{m=1}^{d_{k+1}} s_i^2u(i,j)f_{k+1}(j)u(i,m)f_{k+1}(m)}\\
&=& \frac{||\theta_k||_2}{||f_l||_2} \sqrt{\sum_{i=1}^{\min(d_k,d_l)}\sum_{j=1}^{d_{k+1}}\sum_{m=1}^{d_{k+1}} s_i^2f_{k+1}(j)f_{k+1}(m)\mathbb{E}_{u}u(i,j)u(i,m)}\\
&=& \frac{||\theta_k||_2}{||f_l||_2} \sqrt{\sum_{i=1}^{\min(d_k,d_l)}\sum_{j=1}^{d_{k+1}}\sum_{m=1}^{d_{k+1}} s_i^2f_{k+1}(j)f_{k+1}(m)\delta_{jm}\mathbb{E}_{u}u(i,j)^2}\\
&=& \frac{||\theta_k||_2}{||f_l||_2} \sqrt{\sum_{i=1}^{\min(d_k,d_l)}\sum_{j=1}^{d_{k+1}}s_i^2f_{k+1}(j)^2\mathbb{E}_{u}u(i,j)^2}\\
&=& \frac{||\theta_k||_2}{||f_l||_2} \sqrt{\frac{1}{d_kd_{k+1}}\sum_{i=1}^{\min(d_k,d_l)}\sum_{j=1}^{d_{k+1}}s_i^2f_{k+1}(j)^2}\\
&=& \frac{||\theta_k||_2}{||f_l||_2\sqrt{d_{k+1}}} \sqrt{\frac{1}{d_k}\sum_{i=1}^{\min(d_k,d_l)}s_i^2\sum_{j=1}^{d_{k+1}}f_{k+1}(j)^2}\\
&=& \frac{||\theta_k||_2}{||f_l||_2\sqrt{d_{k+1}}} ||\Sigma||_{qm}||f_{k+1}||_2\\
&=& \frac{||\theta_k||_2}{||f_l||_2\sqrt{d_{k+1}}} ||\mathcal{J}^l_k||_{qm}||f_{k+1}||_2\\
&=& GSC(k,l)\frac{||\theta_k||_2||f_{k+1}||_2}{||f_k||_2\sqrt{d_{k+1}}}
\end{eqnarray*}

Further, assume that $\theta_k$ is random and radially symmetric. Under those conditions, $\theta_k$ is the product of a random scalar length variable $\ell$ and an independent, uniformly random unit length vector $u_k$. Throughout this proof, we will also use $u_k$ to refer to the corresponding matrix. Then we have:


\begin{eqnarray*}
&& \mathbb{Q}^{-1}_{\theta_k}\frac{||\theta_k||_2||f_{k+1}||_2}{||f_k||_2\sqrt{d_{k+1}}}\\
&=& \frac{||f_{k+1}||_2}{\sqrt{d_{k+1}}} \mathbb{Q}^{-1}\frac{||\theta_k||_2}{||f_k||_2}\\
&=& \frac{||f_{k+1}||_2}{\sqrt{d_{k+1}}} \mathbb{Q}^{-1}\sqrt{\frac{||\theta_k||_2^2}{||f_k||_2^2}}\\
&=& \frac{||f_{k+1}||_2}{\sqrt{d_{k+1}}} \mathbb{Q}^{-1}\sqrt{\frac{||\theta_k||_2^2}{||\theta_kf_{k+1}||_2^2}}\\
&=& \frac{||f_{k+1}||_2}{\sqrt{d_{k+1}}} \sqrt{\mathbb{E}\frac{||\theta_kf_{k+1}||_2^2}{||\theta_k||_2^2}}^{-1}\\
&=& \frac{||f_{k+1}||_2}{\sqrt{d_{k+1}}} \sqrt{\mathbb{E}\frac{||\ell u_kf_{k+1}||_2^2}{||\ell u_k||_2^2}}^{-1}\\
&=& \frac{||f_{k+1}||_2}{\sqrt{d_{k+1}}} \sqrt{\mathbb{E}||u_kf_{k+1}||_2^2}^{-1}\\
&=& \frac{||f_{k+1}||_2}{\sqrt{d_{k+1}}} \sqrt{\mathbb{E}\sum_i(\sum_ju_k(i,j)f_{k+1}(j))^2}^{-1}\\
&=& \frac{||f_{k+1}||_2}{\sqrt{d_{k+1}}} \sqrt{\mathbb{E}\sum_{i,j,m}u_k(i,j)f_{k+1}(j)u_k(i,m)f_{k+1}(m)}^{-1}\\
&=& \frac{||f_{k+1}||_2}{\sqrt{d_{k+1}}} \sqrt{\mathbb{E}\sum_{i,j}u_k(i,j)^2f_{k+1}(j)^2}^{-1}\\
&=& \frac{||f_{k+1}||_2}{\sqrt{d_{k+1}}} \sqrt{\sum_jf_{k+1}(j)^2\sum_i\mathbb{E}u_k(i,j)^2}^{-1}\\
&=& \frac{||f_{k+1}||_2}{\sqrt{d_{k+1}}} \sqrt{\sum_jf_{k+1}(j)^2\sum_i(\mathbb{Q}u_k(i,j))^2}^{-1}\\
&=& \frac{||f_{k+1}||_2}{\sqrt{d_{k+1}}} \sqrt{\sum_jf_{k+1}(j)^2\sum_i\frac{1}{d_kd_{k+1}}}^{-1}\\
&=& \frac{||f_{k+1}||_2}{\sqrt{d_{k+1}}} \sqrt{||f_{k+1}||_2^2d_k\frac{1}{d_kd_{k+1}}}^{-1}\\
&=& 1
\end{eqnarray*}

\end{JMLRproof}

$\theta_k$ is radially symmetric, for example, if the corresponding weight matrix is either Gaussian initialized or orthogonally initialized. Therefore, the most popular initialization strategies for weight matrices are covered by this proposition.

\subsection{Proposition \ref{scalinginvariance}}\label{prop4proof}

\begin{repproposition}{scalinginvariance}

Given:

\begin{itemize}
\item some network $f$ of nominal depth $L$
\item some parameter value $\theta = (\theta_1, .., \theta_L)$
\item constants $c_2, .., c_L$ and $\gamma_1, .., \gamma_L$
\item a network $f'$ of nominal depth $L$ defined via its layer functions as follows.
\begin{eqnarray*}
f'_0 &=& f_0\\
f'_1(\theta_1,F_2) &=& f'_1(\gamma_1\theta_1,\frac{1}{c_2}F_2)\\
f'_l(\theta_l,F_{l+1}) &=& c_lf_l(\gamma_l\theta_l,\frac{1}{c_{l+1}}F_{l+1}) \text{ for } 2 \le l \le L-1\\
f'_L(\theta_L,X) &=& c_Lf_L(\gamma_L\theta_L,X)
\end{eqnarray*}
\item a parameter value $\theta' = (\theta'_1, .., \theta'_L)$ defined via $\theta'_l = \frac{1}{\gamma_l}\theta_l$
\end{itemize}

Then for all tuples $(k,l, X, Y)$ with $0 \le l \le k \le L+1$, \begin{equation*}GSC(k,l,f,\theta,X,Y) = GSC(k,l,f',\theta',X,Y)\end{equation*}
\end{repproposition}

\begin{JMLRproof}
Let $c_0 = c_1 = c_{L+1} = 1$. Then we have $||f'_l(\theta')||_2 = c_l ||f_l(\theta)||_2$ for $0 \le l \le L+1$ and we have $||\mathcal{J}'^l_k(\theta')||_{qm} = ||\frac{c_l}{c_k}\mathcal{J}^l_k(\theta)||_{qm} = \frac{c_l}{c_k}||\mathcal{J}^l_k(\theta)||_{qm}$ for $0 \le l \le k \le L+1$, so $GSC(k,l,f',\theta',X,Y) = \frac{||\mathcal{J}'^l_k(\theta')||_{qm}||f'_k(\theta')||_2}{||f'_l(\theta')||_2} = \frac{\frac{c_l}{c_k}||\mathcal{J}^l_k(\theta)||_{qm}c_k||f_k(\theta)||_2}{c_l||f_l(\theta)||_2} = GSC(k,l,f,\theta,X,Y)$ for $0 \le l \le k \le L+1$ as required.

\end{JMLRproof}

Here, we consider general multiplicative rescalings provided they do not change the predictions and error values of the network. To ensure this, each layer function must compensate for the factor introduced by the previous layer as well as for the rescaling of the parameter. Not all network transformations that are used in practice to control the scale of forward activations fall under this proposition. Changing the scale of weights in a tanh or SELU network without normalization layers is not covered, and neither is adding normalization layers to a given network. These changes can have an intricate impact on the properties of the network, as shown throughout the paper. On the other hand, changing the scale of weights in a ReLU network is covered by the proposition, as long as the product of all rescaling factors is 1. Also, changing the scale of weights in any architecture where linear layers are followed by a normalization layer is covered by the proposition.

\subsection{Proposition \ref{compositional}}\label{prop5proof}

\begin{repproposition}{compositional}

Assuming $||\mathcal{J}^l_{l+1}\mathcal{J}^{l+1}_{l+2}..\mathcal{J}^{k-1}_{k}||_{qm} \approx ||\mathcal{J}^l_{l+1}||_{qm}||\mathcal{J}^{l+1}_{l+2}||_{qm}..||\mathcal{J}^{k-1}_{k}||_{qm}$, we have $GSC(k,l) \approx GSC(k,k-1)GSC(k-1,k-2)..GSC(l+1,l)$.

\end{repproposition}

\begin{JMLRproof}
\begin{eqnarray*}
&&GSC(k,l) \\
&=&\frac{||\mathcal{J}^l_k||_{qm}||f_k||_2}{||f_l||_2}\\
&=&\frac{||\mathcal{J}^l_{l+1}\mathcal{J}^{l+1}_{l+2}..\mathcal{J}^{k-1}_{k}||_{qm}||f_k||_2}{||f_l||_2}\\
&\approx&\frac{||\mathcal{J}^l_{l+1}||_{qm}||\mathcal{J}^{l+1}_{l+2}||_{qm}..||\mathcal{J}^{k-1}_{k}||_{qm}||f_k||_2}{||f_l||_2}\\
&=&\frac{||\mathcal{J}^l_{l+1}||_{qm}||\mathcal{J}^{l+1}_{l+2}||_{qm}..||\mathcal{J}^{k-1}_{k}||_{qm}||f_k||_2}{||f_l||_2}\frac{||f_{l+1}||_2}{||f_{l+1}||_2}\frac{||f_{l+2}||_2}{||f_{l+2}||_2}..\frac{||f_{k-1}||_2}{||f_{k-1}||_2}\\
&=&\frac{||\mathcal{J}^l_{l+1}||_{qm}||f_{l+1}||_2}{||f_l||_2}\frac{||\mathcal{J}^{l+1}_{l+2}||_{qm}||f_{l+2}||_2}{||f_{l+1}||_2}..\frac{||\mathcal{J}^{k-1}_k||_{qm}||f_k||_2}{||f_{k-1}||_2}\\
&=&GSC(k,k-1)GSC(k-1,k-2)..GSC(l+1,l)
\end{eqnarray*}
\end{JMLRproof}

\section{Theorems and proofs} \label{theoProofSection}

\subsection{Theorem \ref{depthTheorem} - exploding gradients limit depth} \label{theo1proof}

See section \ref{depthDetailsSection} for the formal definition of effective depth and related concepts. We begin by introducing terminology for denoting the sum of $\lambda$-residual terms $res^\lambda$.

Consider some MLP $f$ with nominal depth $L$ and layers $f_l$, $1 \le l \le L$. Let its compositional depth be $\mathcal{N}$ and its linear layers be $f_{l_n}$, $1 \le n \le \mathcal{N}$ where $l_1 < l_2 < .. < l_\mathcal{N}$. Let each linear layer be the sum of an unparametrized initial function $i_{l_n}$ and a parametrized residual function $r_{l_n}(\theta_{l_n})$. $i_{l_n}$ represents multiplication with the initial weight matrix and is used interchangeably to denote that initial weight matrix. $r_{l_n}(\theta_{l_n})$ represents multiplication with the residual weight matrix and is used interchangeably to denote that residual weight matrix. The parameter sub-vector $\theta_{l_n}$ contains the entries of the residual weight matrix. 

Let an $N$-trace $\phi_N$ be a subset of $\{1, .., N\}$. Let $\Phi_N$ be the set of all possible $N$-traces and let $\Phi_N^\lambda$ be the set of all $N$-traces with at least $\lambda$ elements. We define the `gradient term' $G(\phi_N, f, \theta, X, Y) := J_0J_1..J_{l_{N+1}-1}$ where $J_k = \mathcal{J}^k_{k+1}$ if layer $k$ is not a linear layer, $J_k = r_{l_n}(\theta_{l_n})$ if layer $k$ corresponds to linear layer $l_n$ and $n \in \phi_N$, and $J_k = i_{l_n}$ if layer $k$ corresponds to linear layer $l_n$ and $n \not \in \phi_N$. Finally, we can denote the sum of $\lambda$-residual terms of $\frac{df}{d\theta_{l_N}}$ as $res_N^\lambda(f, \theta, X, Y) := (\sum_{\phi_{N-1} \in \Phi_{N-1}^\lambda} G(\phi_{N-1}, f, \theta, X, Y))^Tf_{l_N+1}^T$.

\begin{reptheorem}{depthTheorem}

Consider an MLP $f$ as defined above with all parameter sub-vectors initialized to $\theta_{l_n}^{(0)} = 0$. Consider some set of possible query points $\mathcal{D}$. Let each of the parameter sub-vectors be updated with a sequence of updates $\Delta \theta_{l_n}^{(t)}$ such that $\theta_{l_n}^{(t)} = \theta_{l_n}^{(t-1)} + \Delta \theta_{l_n}^{(t)}$ with $1 \le t \le T$. Let $alg$ be a fixed function and $\lambda$ a positive integer. Further assume:

\begin{enumerate}
\item $\Delta \theta_{l_n}^{(t)} = alg(\frac{df(\theta^{(t-1)},X^{(t)},Y^{(t)})}{d\theta_{l_n}})$ for some $(X^{(t)},Y^{(t)}) \in \mathcal{D}$
\item There exist constants $r > 1$ and $c > 0$ such that $\frac{||\Delta \theta_{l_n}^{(t)}||_2}{||i_{l_n}||_2} \le \frac{1}{cr^n}$ for all $n$ and $t$
\pagebreak
\item There exists a constant $c' \ge 1$ such that 
\begin{eqnarray*}
&&\frac{||alg\Big(\frac{df(\theta^{(t-1)},X^{(t)},Y^{(t)})}{d\theta_{l_N}}\Big)-alg\Big(\frac{df(\theta^{(t-1)},X^{(t)},Y^{(t)})}{d\theta_{l_N}} - res^\lambda_N(f, \theta^{(t-1)},X^{(t)},Y^{(t)})\Big)||_2}{||alg\Big(\frac{df(\theta^{(t-1)},X^{(t)},Y^{(t)})}{d\theta_{l_N}}\Big)||_2}\\
&\le& c' \sum_{\phi_{N-1} \in \Phi^\lambda_{N-1}} \prod_{n \in \phi_{N-1}}\frac{||r_{l_{n}}(\theta^{(t-1)}_{l_{n}})||_2}{||i_{l_{n}}||_2}
\end{eqnarray*}

for all $N$ and $t$.
\item $T \le \frac{ch^\frac{1}{\lambda}}{32rc'}(\ln r)^3r^{\frac{\lambda}{4}}$ for some $h \le 1$
\end{enumerate}


Then for all $N$ we have

\begin{equation*}
\frac{1}{||i_{l_N}||_2} ||\sum_{t=1}^T \Bigg( alg\Big(\frac{df(\theta^{(t-1)},X^{(t)},Y^{(t)})}{d\theta_{l_N}}\Big)
-alg\Big(\frac{df(\theta^{(t-1)},X^{(t)},Y^{(t)})}{d\theta_{l_N}} - res_N^\lambda(f, \theta^{(t-1)},X^{(t)},Y^{(t)})\Big)\Bigg)||_2\le h
\end{equation*}

\end{reptheorem}

\begin{JMLRproof}
For all $T' \le T$, we have $||\theta_{l_n}^{(T')}||_2 = ||\sum_{t=1}^{T'} \Delta \theta_{l_n}^{(t)}||_2 \le \sum_{t=1}^{T'} ||\Delta \theta_{l_n}^{(t)}||_2 \le \frac{T'}{cr^n}||i_{l_n}||_2\le \frac{T}{cr^n}||i_{l_n}||_2$. Also let $\Psi^i$ be the set of all sets of positive integers with exactly $i$ elements. Then we have

\begin{eqnarray*}
&&\frac{1}{||i_{l_N}||_2} ||\sum_{t=1}^T\Bigg(alg\Big(\frac{df(\theta^{(t-1)},X^{(t)},Y^{(t)})}{d\theta_{l_N}}\Big)-alg\Big(\frac{df(\theta^{(t-1)},X^{(t)},Y^{(t)})}{d\theta_{l_N}} - res^\lambda_N(f, \theta^{(t-1)},X^{(t)},Y^{(t)})\Big)\Bigg)||_2\\
&\le&\frac{1}{||i_{l_N}||_2}\sum_{t=1}^T ||alg\Big(\frac{df(\theta^{(t-1)},X^{(t)},Y^{(t)})}{d\theta_{l_N}}\Big)-alg\Big(\frac{df(\theta^{(t-1)},X^{(t)},Y^{(t)})}{d\theta_{l_N}} - res^\lambda_N(f, \theta^{(t-1)},X^{(t)},Y^{(t)})\Big)||_2\\
&\le& \frac{1}{||i_{l_N}||_2}\sum_{t=1}^T||alg\Big(\frac{df(\theta^{(t-1)},X^{(t)},Y^{(t)})}{d\theta_{l_N}}\Big)||_2 c' \sum_{\phi_{N-1} \in \Phi^\lambda_{N-1}} \prod_{n \in \phi_{N-1}}\frac{||r_{l_{n}}(\theta_{l_{n}}^{(t-1)})||_2}{||i_{l_{n}}||_2}\\
&=& c'\sum_{t=1}^T \frac{||\Delta \theta_{l_N}^{(t)}||_2}{||i_{l_N}||_2}  \sum_{\phi_{N-1} \in \Phi^\lambda_{N-1}} \prod_{n \in \phi_{N-1}}\frac{||r_{l_{n}}(\theta^{(t-1)}_{l_{n}})||_2}{||i_{l_{n}}||_2}\\
&\le& \frac{c'T}{cr^N}  \sum_{\phi_{N-1} \in \Phi^\lambda_{N-1}} \prod_{n \in \phi_{N-1}}\frac{||\theta^{(t-1)}_{l_{n}}||_2}{||i_{l_{n}}||_2}\\
&\le& \frac{Tc'}{cr^N}\sum_{\phi_{N-1} \in \Phi_{N-1}^\lambda}\prod_{n \in \phi_{N-1}}\frac{T}{cr^n}\\
&=& \frac{Tc'}{cr^N}\sum_{i=\lambda}^{N-1}\sum_{\phi_{N-1} \in \Phi_{N-1}^\lambda, |\phi_{N-1}|=i}\prod_{n \in \phi_{N-1}}\frac{T}{cr^n}\\
&\le& \frac{Tc'}{cr^N}\sum_{i=\lambda}^{N-1}\sum_{\psi \in \Psi^i}\prod_{n \in \psi}\frac{T}{cr^n}\\
&=& \frac{Tc'}{cr^N}\sum_{i=\lambda}^{N-1}(\frac{T}{c})^i\sum_{\psi \in \Psi^i}\prod_{n \in \psi}\frac{1}{r^n}\\
&\le& \frac{Tc'}{cr^N}\sum_{i=\lambda}^\infty(\frac{T}{c})^i\sum_{\psi \in \Psi^i}\prod_{n \in \psi}\frac{1}{r^n}\\
\end{eqnarray*}

Let $K(i,n)$ be the number of ways to choose $i$ distinct positive integers such that their sum is $n$. Clearly, $K(i, n) = 0$ for $n < \frac{i(i+1)}{2}$. For $n \ge \frac{i(i+1)}{2}$, the largest number that can be among the chosen $i$ can be $n - \frac{i(i+1)}{2} + i$ and so $K(i,n) \le (n - \frac{i(i+1)}{2} + i)^i = (n - \frac{i(i-1)}{2})^i$. So we have

\begin{eqnarray*}
&&\frac{1}{||i_{l_N}||_2} ||\sum_{t=1}^T\Bigg(alg\Big(\frac{df(\theta^{(t-1)},X^{(t)},Y^{(t)})}{d\theta_{l_N}}\Big)-alg\Big(\frac{df(\theta^{(t-1)},X^{(t)},Y^{(t)})}{d\theta_{l_N}} - res^\lambda_N(f, \theta^{(t-1)},X^{(t)},Y^{(t)})\Big)\Bigg)||_2\\
&\le&\frac{Tc'}{cr^N}\sum_{i=\lambda}^\infty(\frac{T}{c})^i\sum_{\psi \in \Psi^i}\prod_{n \in \psi}\frac{1}{r^n}\\
&=&\frac{Tc'}{cr^N}\sum_{i=\lambda}^\infty(\frac{T}{c})^i\sum_n\frac{K(i, n)}{r^n}\\
&=&\frac{Tc'}{cr^N}\sum_{i=\lambda}^\infty(\frac{T}{c})^i\sum_{n=\frac{i(i+1)}{2}}^\infty\frac{K(i, n)}{r^n}\\
&\le&\frac{Tc'}{cr^N}\sum_{i=\lambda}^\infty(\frac{T}{c})^i\sum_{n=\frac{i(i+1)}{2}}^\infty\frac{(n - \frac{i(i-1)}{2})^i}{r^n}\\
&=&\frac{Tc'}{cr^N}\sum_{i=\lambda}^\infty(\frac{T}{c})^i r^{-\frac{i(i-1)}{2}}\sum_{n=1}^\infty\frac{n^i}{r^n}\\
&<&\frac{Tc'}{cr^N}\sum_{i=\lambda}^\infty(\frac{T}{c})^i r^{-\frac{i(i-1)}{2}}i^i(\ln r)^{-i}\\
&\le&c'\sum_{i=\lambda}^\infty(\frac{T}{c})^{i+1} r^{-\frac{i(i-1)}{2}}i^i(\ln r)^{-i}\\
&\le&c'\sum_{i=\lambda}^\infty(\frac{\frac{ch^\frac{1}{\lambda}}{32rc'}(\ln r)^3r^{\frac{\lambda}{4}}}{c})^{i+1} r^{-\frac{i(i-1)}{2}}i^i(\ln r)^{-i}\\
&\le&\sum_{i=\lambda}^\infty h^\frac{i+1}{\lambda}(\frac{1}{2})^{i+1} (\frac{1}{16})^{i+1}(\ln r)^{2i+3} r^{-\frac{1}{4}i^2-\frac{1}{4}i-1}i^i\\
\end{eqnarray*}

Let $\nu(r,i) := (\frac{1}{16})^{i+1}(\ln r)^{2i+3} r^{-\frac{1}{4}i^2-\frac{1}{4}i-1}i^i$. Now we show that for all $r > 1$ and positive integers $i$, $\nu(r,i) \le 1$.

\paragraph{Case 1: $r^{\frac{i}{4}} \le i$.} We have 

\begin{eqnarray*}
&&(\frac{1}{16})^{i+1}(\ln r)^{2i+3} r^{-\frac{1}{4}i^2-\frac{1}{4}i-1}i^i \\
&<&(\frac{1}{16})^{i+1}(\frac{4}{i}\ln i)^{2i+3} 1^{-\frac{1}{4}i^2-\frac{1}{4}i-1}i^i \\
&=& (\frac{1}{16})(\frac{4}{i}\ln i)^3(\frac{1}{16}(\frac{4}{i}\ln i)^2i)^i\\
&\le &(\frac{(\ln i)^2}{i})^i\\
&\le& 1
\end{eqnarray*}

\paragraph{Case 2: $r \ge e$.} $\nu(r,i)$ is smooth in $r$ in the domain $r \ge e$. Also we have that $\frac{d\nu(r,i)}{dr} = \frac{\nu(r,i)}{r}(\frac{2i+3}{\ln r}-\frac{1}{4}i^2-\frac{1}{4}i-1)$ is eventually negative as $r$ tends to infinity. Therefore as long as $\nu(r,i) \le 1$ holds at the boundary of the domain and wherever $\frac{d\nu(r,i)}{dr}=0$, $\nu(r,i) \le 1$ holds in the entire domain. The boundary is $r=e$. We have 

\begin{eqnarray*}
&&(\frac{1}{16})^{i+1}(\ln e)^{2i+3} e^{-\frac{1}{4}i^2-\frac{1}{4}i-1}i^i \\
&=&\frac{1}{16}e^{-1}(\frac{1}{16}e^{-\frac{1}{4}i-\frac{1}{4}}i)^i\\
&\le &(\frac{1}{16}e^{-\frac{1}{4}i-\frac{1}{4}}i)^i\\
&<&1
\end{eqnarray*}

Now let's look at points where $\frac{d\nu(r,i)}{dr}=0$. There, we have $\frac{2i+3}{\ln r}-\frac{1}{4}i^2-\frac{1}{4}i-1 = 0$ and so $\ln r = \frac{8i+12}{i^2+i+4}$ and so $r = e^{\frac{8i+12}{i^2+i+4}}$. Plugging this value of $r$ into $\nu$, we can compute the value of $\nu$ for, say, $i \le 20$ and find that it is less than 1. For $i > 20$, clearly $e^{\frac{8i+12}{i^2+i+4}} < e$ and so $\frac{d\nu(r,i)}{dr}=0$ has no solution in the domain. 

\paragraph{Case 3: $r^{\frac{i}{4}} \ge i$ and $r \le e$.} We have 

\begin{eqnarray*}
&&(\frac{1}{16})^{i+1}(\ln r)^{2i+3} r^{-\frac{1}{4}i^2-\frac{1}{4}i-1}i^i \\
&<&(\ln e)^{2i+3} (r^{-\frac{1}{4}i}i)^i\\
&\le &(\frac{1}{i}i)^i\\
&=& 1
\end{eqnarray*}

So in all cases, we obtain that the desired $\nu(r,i) \le 1$. So we have

\begin{eqnarray*}
&&\frac{1}{||i_{l_N}||_2} ||\sum_{t=1}^T\Bigg(alg\Big(\frac{df(\theta^{(t-1)},X^{(t)},Y^{(t)})}{d\theta_{l_N}}\Big)-alg\Big(\frac{df(\theta^{(t-1)},X^{(t)},Y^{(t)})}{d\theta_{l_N}} - res^\lambda_N(f, \theta^{(t-1)},X^{(t)},Y^{(t)})\Big)\Bigg)||_2\\
&\le& \sum_{i=\lambda}^\infty h^\frac{i+1}{\lambda}(\frac{1}{2})^{i+1}\nu(r,i)\\
&\le&\sum_{i=\lambda}^\infty h^\frac{i+1}{\lambda}(\frac{1}{2})^{i+1}\\
&<&h
\end{eqnarray*}

\end{JMLRproof}

$alg$ represents the gradient-based algorithm and the quantity that is ultimately bounded by $h$ is the first-order approximation of the relative $\lambda$-contribution at layer $l_N$ until time $T$. To obtain that the network has effective depth $\lambda$, all we need is to set $h$ to a small value. In that case, the first-order approximation is sufficient.

Now, we analyze the four conditions in turn.

Condition (1) states that the algorithm computes the update. For convenience, we write the algorithm as a deterministic function of the gradient of the layer for which the update is computed. The proof can be trivially extended to algorithms that use the gradients of other layers, past gradients and as well randomness if we add the same dependencies to condition (3). Also for convenience, we assume a batch size of 1. We can apply the result to larger batch sizes, for example, by having $alg$ use past gradients and setting the majority of updates to zero.

Condition (2) reflects the argument from section \ref{analysisSection} that the area around the current parameter value in which the gradient is reflective of the function is bounded by a hypersphere of relative radius $\frac{1}{GSC(l_n, 0)}$, and the assumption that gradients explode, i.e. $GSC(l_n, 0) \ge cr^{l_n}$. Note that for convenience, we divide the size of the update $||\Delta \theta_{l_n}^{(t)}||_2$ by the weight matrix in the initialized state $||i_{l_n}||_2$ instead of $||\theta_{l_n}^{(t-1)}||_2$. This is realistic given the general observation that the largest useful update size decreases in practice when training a deep network. Therefore, we can bound all updates by the largest useful update size in the initialized state.

The strongest condition is (3). It can be understood as making two distinct assertions. 

Firstly, ignoring the $alg()$ function, it bounds the length of the sum of the $\lambda$-residual terms. In essence, it requires that on average, the size of these terms is ``what one would expect'' given the $L2$ norm of the initial and residual weight matrices up to some constant $c'$. In other words, we assume that in $\lambda$-residual terms, the large singular values of layer-wise Jacobians do not compound disproportionately compared to the full gradient. The bound is however also very conservative in the sense that it can tolerate all $\lambda$-residual terms having the same orientation.

Secondly, it asserts that $alg()$ is ``relatively Lipschitz'' over the gradient. This is fulfilled e.g. for SGD and SGD with custom layer-wise step sizes as used in our experiments. It is fulfilled by SGD with momentum as long as step sizes are non-increasing. In theory, it is not fulfilled by RMSprop or Adam as gradients on individual weight matrix entries can be ``scaled up'' arbitrarily via the denominator. In practice, the $\epsilon$ regularization term used in the denominator prevents this, although this is rarely necessary. 

Finally, condition (4) states that the training time is limited. Importantly, the bound on $T$ is exponential in $\lambda$ and independent of both $L$ and $\mathcal{N}$. Note that we did not attempt to make the bound tight. As it stands, unfortunately, the bound is too loose to have much practical value. It would indicate that networks can be trained to far greater depth than is possible in practice. See section \ref{depthexperimentssection} for an experimental study of effective depth.

\subsection{Theorem \ref{explodeTheorem} - why gradients compound} \label{theo2proof}

\begin{reptheorem}{explodeTheorem}

Consider a neural network $f$ with random parameter $\theta$. Define $\mathcal{J}_l := \mathcal{J}^l_{l+1}$. Let $c_l$, $1\le l \le L+1$ be fixed positive constants. Let $F_l$, $1\le l \le L$, be vectors of dimensionality $d_l$ respectively that vary among vectors of length $c_l$. Let $F_{L+1}$ be a vector of dimensionality $d_{L+1}$ that varies in the domain of data inputs $\mathcal{X}$. Let $U_l$, $1\le l \le L+1$, be vectors of dimensionality $d_l$ respectively that vary among vectors of length $1$. 

 Assume:

\begin{enumerate}
\item The $\theta_l$ are independent of each other. \label{t2c1}
\item For all $1 \le l \le L$ and $F_{l+1}$, $f_l(\theta_l,F_{l+1})$ is a uniformly random vector of length $c_l$.\label{t2c3}
\item For all $1 \le l \le L$, $U_{l+1}$ and $F_{l+1}$, $\mathcal{J}_l(\theta_l,F_{l+1})U_{l+1}$ can be written as $\ell_l(\theta_l,F_{l+1}, U_{l+1})u_l$. Here, $\ell_l$ is a random scalar independent of both $u_l$ and $f_l(\theta_l, F_{l+1})$. $u_l$, conditioned on $f_l(\theta_l,F_{l+1})$, is uniformly distributed in the space of unit length vectors orthogonal to $f_l(\theta_l,F_{l+1})$.\label{t2c5}
\item For all $1 \le l \le L$ and $F_{l+1}$, $\frac{df_l(\theta_l,cF_{l+1})}{dc}|_{c=1} = 0$ \label{t2c6}
\end{enumerate}


Then we have \begin{equation*}\sqrt{\frac{d_k}{d_k-1}}\mathbb{Q}_{\theta} GSC(k,l) = \prod_{l'=l+1}^{k} \sqrt{\frac{d_{l'}}{d_{l'}-1}} \mathbb{Q}_\theta GSC(l',l'-1)\end{equation*} for all $0 \le l < k \le L+1$.

\end{reptheorem}

\begin{JMLRproof}
Let $\theta^l_k := (\theta_l, \theta_{l+1}, .., \theta_{k-1})$. We begin by proving the following claim by induction on $k-l$ starting from $k-l=1$.\\

(A) For all $1 \le l < k \le L+1$ and $F_k$, $f_l(\theta^l_k, F_k)$ is a uniformly random vector of length $c_l$ independent of $\theta^1_l$. \\

The base case is $k-l=1$. By condition (\ref{t2c3}), we directly obtain that $f_l(\theta_l, F_k)$ is uniformly distributed of length $c_l$. Also, since $F_k$ is fixed, it is independent of $\theta^1_l$. Since $\theta_l$ is also independent of $\theta^1_l$ by condition (\ref{t2c1}), $f_l(\theta_l, F_k)$ is independent of $\theta^1_l$. This completes the base case.

Now for the induction step. Let $k-l \ge 2$. By the induction hypothesis, $f_{l+1}(\theta^{l+1}_k, F_k)$ is uniformly random of length $c_{l+1}$ and independent of $\theta^1_{l+1}$ and thus independent of $\theta^1_l$. Then by condition (\ref{t2c3}), $f_l(\theta^l_k, F_k) = f_l(\theta_l, f_{l+1}(\theta^{l+1}_k, F_k))$ is uniformly random of length $c_l$. Also, since both $\theta_l$ and $f_{l+1}(\theta^{l+1}_k, F_k)$ are independent of $\theta^1_l$, so is $f_l(\theta^l_k, F_k)$. This completes the induction step. 

Now we will prove the following claim by induction on $k-l$ starting from $k-l=1$.\\

(B) For all $1 \le l < k \le L+1$, $U_k$ and $F_k$, $\mathcal{J}^l_k(\theta^l_k, F_k)U_k$ can be written as $\ell^l_k(\theta^l_k,F_k,U_k)u_l$. Here, $\ell^l_k$ is a random scalar independent of both $u_l$ and $f_l(\theta^l_k,F_k)$. $u_l$, conditioned on $f_l(\theta^l_k,F_k)$, is uniformly distributed in the space of unit length vectors orthogonal to $f_l(\theta^l_k,F_k)$.\\

The case $k-l=1$ is equivalent to condition (\ref{t2c5}). For the induction step, consider

\begin{eqnarray*}
&&\mathcal{J}^l_k(\theta^l_k, F_k)U_k\\
&=&\mathcal{J}^l_{l+1}(\theta_l,f_{l+1}(\theta^{l+1}_k,F_k))\mathcal{J}^{l+1}_k(\theta^{l+1}_k,F_k)U_k\\
&=&\mathcal{J}^l_{l+1}(\theta_l,f_{l+1}(\theta^{l+1}_k,F_k))\ell^{l+1}_k(\theta^{l+1}_k,F_k,U_k)u_{l+1}\\
&=&\ell^{l+1}_k(\theta^{l+1}_k,F_k,U_k)\mathcal{J}^l_{l+1}(\theta_l,f_{l+1}(\theta^{l+1}_k,F_k))u_{l+1}\\
&=&\ell^{l+1}_k(\theta^{l+1}_k,F_k,U_k)\ell^l_{l+1}(\theta_l,f_{l+1},u_{l+1})u_l
\end{eqnarray*}

Here, we used claim (A) and the induction hypothesis twice.

Let $\ell^l_k(\theta^l_k,F_k,u) := \ell^{l+1}_k(\theta^{l+1}_k,F_k,u)\ell^l_{l+1}(\theta_l,f_{l+1},u_{l+1})$. From when we used the induction hypothesis the first time, we obtained that $\ell^{l+1}_k$ is independent of $f_{l+1}$ and $u_{l+1}$, and therefore of $f_l(\theta_l,f_{l+1})$ and $\mathcal{J}^l_{l+1}(\theta_l,f_{l+1})$, and therefore of $u_l$. From when we used the induction hypothesis the second time, we obtain that $\ell^l_{l+1}$ is independent of $u_l$ and $f_l$. Therefore, both $\ell^{l+1}_k$ and $\ell^l_{l+1}$ are independent of both $u_l$ and $f_l$, and then so is $\ell^l_k$. From using the induction hypothesis the second time we also obtained that conditioned on $f_l$, $u_l$ is uniformly distributed among unit vectors orthogonal to $f_l$. Therefore, the decomposition $\mathcal{J}^l_k(\theta^l_k, F_k)u = \ell^l_ku_l$ fulfills all requirements, so the induction step is complete.

Now we will prove the theorem by induction on $k-l$ starting from $k-l=1$.


The base case is trivial. Consider $k-l \ge 2$. Let $u_k$ be a $d_k$-dimensional, uniformly random unit length vector. We have 

\begin{eqnarray*}
&&\mathbb{Q}_\theta GSC(k,l)\\
&=&\mathbb{Q}_\theta \frac{||\mathcal{J}^l_k||_{qm}||f_k||_2}{||f_l||_2}\\
&=&c_k\mathbb{Q}_\theta \frac{||\mathcal{J}^l_{l+1}\mathcal{J}^{l+1}_k||_{qm}}{||f_l||_2}\\
&=&c_k\mathbb{Q}_{\theta,u_k} \frac{||\mathcal{J}^l_{l+1}\mathcal{J}^{l+1}_ku_k||_2}{||f_l||_2}\\
&=&c_k\mathbb{Q}_{\theta,u_k} \frac{||\mathcal{J}^l_{l+1}(\theta_l,f_{l+1}(\theta^{l+1}_L,x))\mathcal{J}^{l+1}_k(\theta^{l+1}_k,f_k(\theta^k_L,x))u_k||_2}{||f_l||_2}
\end{eqnarray*}

$f_k(\theta^k_L,x)$ has length $c_k$ and is independent of $\theta^{l+1}_k$ by claim (A). So by claim (B), we have that $\mathcal{J}^{l+1}_ku_k$ can be written as $\ell^{l+1}_k(\theta^{l+1}_k,f_k,u_k)u_{l+1}$ with the properties stated in claim (B). Using those properties, we obtain

\begin{eqnarray*}
&&\mathbb{Q}_\theta GSC(k,l)\\
&=& c_k\mathbb{Q}_{\theta,u_k} \frac{||\mathcal{J}^l_{l+1}(\theta_l,f_{l+1}(\theta^{l+1}_L,x))\mathcal{J}^{l+1}_k(\theta^{l+1}_k,f_k(\theta^k_L,x))u_k||_2}{||f_l(\theta_l,f_{l+1})||_2}\\
&=& c_k\mathbb{Q}_{\theta,u_k} \frac{||\mathcal{J}^l_{l+1}(\theta_l,f_{l+1}(\theta^{l+1}_L,x))\ell^{l+1}_k(\theta^{l+1}_k,f_k,u_k)u_{l+1}||_2}{||f_l(\theta_l,f_{l+1})||_2}\\
&=& c_k\mathbb{Q}_{\theta_l,u_{l+1},f_{l+1},\ell^{l+1}_k} \frac{\ell^{l+1}_k(\theta^{l+1}_k,f_k,u_k)||\mathcal{J}^l_{l+1}(\theta_l,f_{l+1})u_{l+1}||_2}{||f_l(\theta_l,f_{l+1})||_2}\\
&=& c_k(\mathbb{Q}_{\ell^{l+1}_k} \ell^{l+1}_k)(\mathbb{Q}_{\theta_l,f_{l+1},u_{l+1}}\frac{||\mathcal{J}^l_{l+1}(\theta_l,f_{l+1})u_{l+1}||_2}{||f_l(\theta_l,f_{l+1})||_2})
\end{eqnarray*}

Let's look at the first term $\mathbb{Q}_{\ell^{l+1}_k} \ell^{l+1}_k$. We have

\begin{eqnarray*}
&& \mathbb{Q}_{\ell^{l+1}_k} \ell^{l+1}_k\\
&=& \mathbb{Q}_{\theta,u_k} ||\mathcal{J}^{l+1}_ku_k||_2\\
&=& \frac{c_{l+1}}{c_k}\mathbb{Q}_\theta GSC(k,l+1)\\
&=& \frac{c_{l+1}}{c_k} \sqrt{\frac{d_k-1}{d_k}}\prod_{l'=l+2}^{k} \sqrt{\frac{d_{l'}}{d_{l'}-1}} \mathbb{Q}_\theta GSC(l',l'-1)
\end{eqnarray*}

The last line comes from the induction hypothesis.

Now, let's look at the second term $\mathbb{Q}_{\theta_l,f_{l+1},u_{l+1}}\frac{||\mathcal{J}^l_{l+1}(\theta_l,f_{l+1})u_{l+1}||_2}{||f_l||_2}$. $u_{l+1}$ is uniformly random among unit length vectors orthogonal to $f_{l+1}$. Let $u'_{l+1}$ be a uniformly random, $d_{l+1}$-dimensional unit length vector. Then we have $u_{l+1} = \sqrt{\frac{d_{l+1}}{d_{l+1}-1}}(u'_{l+1}-\frac{u'_{l+1}.f_{l+1}}{f_{l+1}.f_{l+1}}f_{l+1})$. By claim (A), $f_{l+1}$ has length $c_{l+1}$ and is independent of $\theta_l$, so by condition (\ref{t2c6}) we have $\mathcal{J}^l_{l+1}(\theta_l,f_{l+1})f_{l+1} = 0$. 


Putting those results together, we obtain 

\begin{eqnarray*}
&& \sqrt{\frac{d_k}{d_k-1}}\mathbb{Q}_\theta GSC(k,l)\\
&=&\sqrt{\frac{d_k}{d_k-1}}c_k(\mathbb{Q}_{\ell^{l+1}_k} \ell^{l+1}_k)(\mathbb{Q}_{\theta_l,f_{l+1},u_{l+1}}\frac{||\mathcal{J}^l_{l+1}(\theta_l,f_{l+1})u_{l+1}||_2}{||f_l(\theta_l,f_{l+1})||_2})\\
&=&\sqrt{\frac{d_k}{d_k-1}}c_k \bigg(\sqrt{\frac{d_k-1}{d_k}}\frac{c_{l+1}}{c_k}\prod_{l'=l+2}^{k} \sqrt{\frac{d_{l'}}{d_{l'}-1}} \mathbb{Q}_\theta GSC(l',l'-1)\bigg)\\
&&(\mathbb{Q}_{\theta_l,f_{l+1},u_{l+1}}\frac{||\mathcal{J}^l_{l+1}(\theta_l,f_{l+1})\sqrt{\frac{d_{l+1}}{d_{l+1}-1}}(u'_{l+1}-\frac{u'_{l+1}.f_{l+1}}{f_{l+1}.f_{l+1}}f_{l+1})||_2}{||f_l(\theta_l,f_{l+1})||_2})\\
&=&c_{l+1}\sqrt{\frac{d_{l+1}}{d_{l+1}-1}}\bigg(\prod_{l'=l+2}^{k} \sqrt{\frac{d_{l'}}{d_{l'}-1}} \mathbb{Q}_\theta GSC(l',l'-1)\bigg)\\
&&(\mathbb{Q}_{\theta_l,f_{l+1},u'_{l+1}}\frac{||\mathcal{J}^l_{l+1}(\theta_l,f_{l+1})u'_{l+1}-\frac{u'_{l+1}.f_{l+1}}{f_{l+1}.f_{l+1}}\mathcal{J}^l_{l+1}(\theta_l,f_{l+1})f_{l+1}||_2}{||f_l(\theta_l,f_{l+1})||_2})\\
&=&c_{l+1}\sqrt{\frac{d_{l+1}}{d_{l+1}-1}}\bigg(\prod_{l'=l+2}^{k} \sqrt{\frac{d_{l'}}{d_{l'}-1}} \mathbb{Q}_\theta GSC(l',l'-1)\bigg)(\mathbb{Q}_{\theta_l,f_{l+1},u'_{l+1}}\frac{||\mathcal{J}^l_{l+1}(\theta_l,f_{l+1})u'_{l+1}||_2}{||f_l(\theta_l,f_{l+1})||_2})\\
&=&c_{l+1}\sqrt{\frac{d_{l+1}}{d_{l+1}-1}}\bigg(\prod_{l'=l+2}^{k} \sqrt{\frac{d_{l'}}{d_{l'}-1}} \mathbb{Q}_\theta GSC(l',l'-1)\bigg)(\mathbb{Q}_{\theta_l,f_{l+1}}\frac{||\mathcal{J}^l_{l+1}(\theta_l,f_{l+1})||_{qm}}{||f_l(\theta_l,f_{l+1})||_2})\\
&=&c_{l+1}\sqrt{\frac{d_{l+1}}{d_{l+1}-1}}\bigg(\prod_{l'=l+2}^{k} \sqrt{\frac{d_{l'}}{d_{l'}-1}} \mathbb{Q}_\theta GSC(l',l'-1)\bigg)\frac{1}{c_{l+1}}\mathbb{Q}_\theta GSC(l+1,l)\\
&=&\prod_{l'=l+1}^{k} \sqrt{\frac{d_{l'}}{d_{l'}-1}} \mathbb{Q}_\theta GSC(l',l'-1)\\
\end{eqnarray*}

This completes the induction hypothesis.

\end{JMLRproof}

Condition (\ref{t2c1}) is standard for randomly initialized weight matrices. Consider an MLP made up of blocks containing a layer normalization layer, a nonlinearity layer and a linear layer, such as our layer-tanh and layer-ReLU architectures. Conditions (\ref{t2c3}), (\ref{t2c5}) and (\ref{t2c6}) are fulfilled exactly in this MLP if we reinterpret multiple layers as a single layer. Define `length-only layer normalization' (LOlayer) as the operation that divides a vector by the standard deviation of its entries. We can insert two LOlayer layers before the layer normalization layer without altering forward or backward evaluation of this MLP. Then we cast the block (LOlayer, layer normalization, nonlinearity, linear, LOlayer) as a single layer. A network defined this way fulfills conditions (\ref{t2c3}), (\ref{t2c5}) and (\ref{t2c6}) if the weight matrices are Gaussian or orthogonally initialized, which are the two most popular styles. 

The LOlayer operations ensure that the length of the input is immaterial (condition (\ref{t2c6})) and that the length of the output is constant. They also ensure that $\mathcal{J}_l(\theta_l,F_{l+1})U_{L+1}$ and $f_l(\theta_l, F_{l+1})$ are orthogonal by inducing a left-singular vector in $\mathcal{J}_l(\theta_l,F_{l+1})$ in the direction of $f_l(\theta_l, F_{l+1})$ with a zero singular value, because $f_l(\theta_l, F_{l+1})$ is the normal of its own co-domain, the hypersphere. The linear operation ensures that both $f_l(\theta_l, F_{l+1})$ and $\mathcal{J}_l(\theta_l,F_{l+1})U_{L+1}$ are mapped to uniformly random orientations, independent of the length of $\mathcal{J}_l(\theta_l,F_{l+1})U_{L+1}$, that depend on each other only via their angle, which is then set to $\frac{\pi}{2}$ by the final LOlayer operation.

The conclusion of this theorem still applies approximately to practical MLPs that do not use LOlayer. In those MLPs, the forward activation vectors at each layer corresponding to different data inputs have approximately the same length, due to the law of large numbers. Thus, inserting a pseudo-LOlayer operation after each linear layer that scales each activation vector to have length equal to the average length across data inputs will not drastically change the forward or backward dynamics of the MLP. We can approximate the original MLP in this way to have the GSC decompose, therefore the GSC decomposes approximately in the original MLP.

The factors $\sqrt{\frac{d_k}{d_k-1}}$ can be $\sqrt{\frac{d_{l'}}{d_{l'}-1}}$ can be understood by observing that each layer $f_l$ maps vectors from a $d_{l-1}-1$-dimensional subspace of vectors of length $c_{l-1}$ to a $d_l-1$-dimensional subspace of vectors of length $c_l$. In some sense, the ``correct'' denominator in the definition of the qm norm is not the number of columns of the matrix, but the intrinsic dimensionality of the input space, which in the case of this theorem is the number of columns minus 1. Making that change would eliminate the $\sqrt{\frac{d_k}{d_k-1}}$ and $\sqrt{\frac{d_{l'}}{d_{l'}-1}}$ factors.

\subsection{Theorem \ref{bigGradTheorem} - why gradients explode} \label{theo2p5proof}

\begin{reptheorem}{bigGradTheorem}

Let $X$ be a random variable with a real-valued probability density function $p$ on $\mathbb{R}^d$ and let $f$ be an endomorphism on $\mathbb{R}^d$. Let $\sigma_{|s|}$ be the standard deviation of the absolute singular values of $\mathcal{J}^f_X$ and let $\mu_{|s|}$ be the mean of the absolute singular values of $\mathcal{J}^f_X$. Assume $\sigma_{|s|} \ge \sqrt{\epsilon(\epsilon+2\mu_{|s|})}$ with probability $\delta$. Then \begin{equation}\mathbb{E}_X||\mathcal{J}^f_X||_{qm} \ge \epsilon\delta + e^{\frac{1}{d}(H(f(X)) - H(X))}\end{equation}

Further, assume that the value of $p$ is independent of $\mathcal{J}^f_X$. Then \begin{equation}\mathbb{E}_X||\mathcal{J}^f_X||_{qm} \ge \epsilon\delta + \frac{E(f(X))}{E(X)}\end{equation}

\end{reptheorem}

\begin{JMLRproof}
Denote by $s_i$ the singular values of $\mathcal{J}^f_X$ and let ${\bf 1}$ denote the indicator that $\sigma_{|s|} \ge \sqrt{\epsilon(\epsilon+2\mu_{|s|})}$. We have $H(X) = \mathbb{E}_X -\log p(X)$ and $H(f(X)) \le \mathbb{E}_X -\log \frac{p(X)}{|\det(\mathcal{J}^f_X(X))|} = \mathbb{E}_X -\log \frac{p(X)}{\prod_i |s_i|} = \mathbb{E}_X -\log p(X) + \mathbb{E}_X \log \prod_i |s_i|$ and so $\frac{1}{d}(H(f(X)) - H(X)) \le \mathbb{E}_X \log \sqrt[d]{\prod_i |s_i|}$. We have equality if $f$ is injective. So we have

\begin{eqnarray*}
&&\mathbb{E}_X||\mathcal{J}^f_X||_{qm}\\
&=&\mathbb{E}_X\sqrt{\frac{1}{d}\sum_i s_i^2}\\
&=&\mathbb{E}_X\sqrt{\mu_{|s|}^2 + \sigma_{|s|}^2}\\
&\ge&\mathbb{E}_X\sqrt{\mu_{|s|}^2 + {\bf 1}\epsilon(\epsilon+2\mu_{|s|})}\\
&=&\mathbb{E}_X\sqrt{(\mu_{|s|} + {\bf 1}\epsilon)^2}\\
&=&\mathbb{E}_X {\bf 1}\epsilon + \mathbb{E}_X\mu_{|s|}\\
&\ge&\delta \epsilon + \mathbb{E}_X\sqrt[d]{\prod_i |s_i|}\\
&=&\delta \epsilon +  e^{\log \mathbb{E}_X\sqrt[d]{\prod_i |s_i|}}\\
&\ge&\delta \epsilon +  e^{\mathbb{E}_X\log\sqrt[d]{\prod_i |s_i|}}\\
&\ge&\delta \epsilon + e^{\frac{1}{d}(H(f(X)) - H(X))}\\
\end{eqnarray*}

We use Jensen's inequality and the fact that the arithmetic mean exceeds the geometric mean.

Now let's look at the second part of the theorem. We have $E(X) = \mathbb{E}_X p(X)^{-\frac{1}{d}}$ and $E(f(X)) \le \mathbb{E}_X \frac{p(X)}{|\det(\mathcal{J}^f_X(X))|}^{-\frac{1}{d}} = \mathbb{E}_X \frac{p(X)}{\prod_i |s_i|}^{-\frac{1}{d}} = \mathbb{E}_X p(X)^{-\frac{1}{d}} \mathbb{E}_X \sqrt[d]{\prod_i |s_i|}$, where again equality holds if $f$ is injective. To attain the last equality, we used the independence of the value of $p$ and the Jacobian. Hence $\frac{E(f(X))}{E(X)} \le \mathbb{E}_X \sqrt[d]{\prod_i |s_i|}$. But recall that we have $\mathbb{E}_X||\mathcal{J}^f_X||_{qm} \ge \delta \epsilon + \mathbb{E}_X\sqrt[d]{\prod_i |s_i|}$, and so $\mathbb{E}_X||\mathcal{J}^f_X||_{qm} \ge \delta \epsilon + \frac{E(f(X))}{E(X)}$ as required.

\end{JMLRproof}

We present this theorem as two separate results because both have advantages. The result in terms of $H$ may be more interesting in general because $H$ is a more widely used and understood quantity. Also, this result does not require the independence condition. Conversely, the result in terms of $E$ is tighter because we can avoid the use of Jensen's inequality. This is especially significant if there is a very small region of space where $f$ has zero singular values. This would not significantly affect the bound in terms of $E$ but would make the bound in terms of $H$ meaningless.

As stated, these results apply only to MLPs where each layer function preserves both the nominal dimensionality, i.e. the number of neurons, as well as the intrinsic dimensionality of the domain. The theorem can be easily extended to cover layer functions that change nominal dimensionality and only preserve intrinsic dimensionality. Extending this theorem to nonlinearities that reduce intrinsic dimensionality across part of the domain, such as ReLU or hard tanh, or to width reductions that reduce intrinsic dimensionality is left for future work.

The simplest way to achieve independence of $p$ and the Jacobian is to make use of the conditions of theorem \ref{explodeTheorem}. Let each layer map inputs from a hypersphere to outputs on another hypersphere with the same intrinsic dimensionality. Let the orientation of the output be uniformly random when viewed as a function of the parameter. Then $p$, when viewed as a random function of the parameter, has the same distribution everywhere on the hypersphere. Hence, we obtain the required independence if we replace $\mathbb{E}_X$ with $\mathbb{E}_{X,\theta}$. The architectures that fulfill the conditions of theorem \ref{explodeTheorem} exactly / approximately are discussed in section \ref{theo2proof}.

\subsection{Theorem \ref{dilutionTheorem} - skip connections reduce the gradient} \label{theo3proof}

\begin{reptheorem}{dilutionTheorem}

Let $g$ and $v$ be random vectors. Consider a function $f_b$ that is $k_b$-diluted with respect to $v$, a matrix $S_b$ and a function $\rho_b$. Let $\mathcal{R}_b(v)$ be the Jacobian of $\rho_b$ at input $v$. Let $r := \frac{\mathbb{Q}||g\mathcal{R}_b(v)||_{qm}\mathbb{Q}||v||_2}{\mathbb{Q}||\rho_b(v)||_2\mathbb{Q}||g||_{qm}}$. 
Assume:
\begin{enumerate}
\item $\mathbb{E}(S_bv).(\rho_b(v)) = 0$ \label{t3c1}
\item $\mathbb{E}(g\mathcal{R}_b(v)).(gS_b) = 0$ \label{t3c2}
\item $\frac{\mathbb{Q}||gS_b||_{qm}\mathbb{Q}||v||_2}{\mathbb{Q}||S_bv||_2\mathbb{Q}||g||_{qm}} = 1$ \label{t3c3}
\end{enumerate}

Then \begin{equation*}\frac{\mathbb{Q}||g\mathcal{R}_b(v) + gS_b||_{qm}\mathbb{Q}||v||_2}{\mathbb{Q}||\rho_b(v) + S_bv||_2\mathbb{Q}||g||_{qm}} = 1 + \frac{r-1}{k^2 + 1} + O((r-1)^2)\end{equation*}.
\end{reptheorem}

\begin{JMLRproof}
Throughout this proof, we omit $b$ subscripts. We have

\begin{eqnarray*}
&&\frac{\mathbb{Q}||g\mathcal{R}(v) + gS||_{qm}\mathbb{Q}||v||_2}{\mathbb{Q}||\rho(v) + Sv||_2\mathbb{Q}||g||_{qm}} \\
&=&\frac{\frac{1}{\sqrt{d_{b+1}}}\mathbb{Q}||g\mathcal{R}(v) + gS||_2\mathbb{Q}||v||_2}{\mathbb{Q}||\rho(v) + Sv||_2\mathbb{Q}||g||_{qm}} \\
&=&\frac{\frac{1}{\sqrt{d_{b+1}}}\mathbb{Q}\sqrt{||g\mathcal{R}(v) + gS||_2^2}\mathbb{Q}||v||_2}{\mathbb{Q}\sqrt{||\rho(v) + Sv||_2^2}\mathbb{Q}||g||_{qm}}\\
&=&\frac{\frac{1}{\sqrt{d_{b+1}}}\sqrt{\mathbb{E}||g\mathcal{R}(v) + gS||_2^2}\mathbb{Q}||v||_2}{\sqrt{\mathbb{E}||\rho(v) + Sv||_2^2}\mathbb{Q}||g||_{qm}}\\
&=&\frac{\frac{1}{\sqrt{d_{b+1}}}\sqrt{\mathbb{E}(g\mathcal{R}(v) + gS).(g\mathcal{R}(v) + gS)}\mathbb{Q}||v||_2}{\sqrt{\mathbb{E}(\rho(v) + Sv).(\rho(v) + Sv)}\mathbb{Q}||g||_{qm}}\\
&=&\frac{\frac{1}{\sqrt{d_{b+1}}}\sqrt{\mathbb{E}[(g\mathcal{R}(v)).(g\mathcal{R}(v)) + 2(gS).(g\mathcal{R}(v)) + (gS).(gS)]}\mathbb{Q}||v||_2}{\sqrt{\mathbb{E}[\rho(v).\rho(v) + 2(Sv).\rho(v) + (Sv).(Sv)]}\mathbb{Q}||g||_{qm}}\\
&=&\frac{\frac{1}{\sqrt{d_{b+1}}}\sqrt{\mathbb{E}[(g\mathcal{R}(v)).(g\mathcal{R}(v)) + (gS).(gS)]}\mathbb{Q}||v||_2}{\sqrt{\mathbb{E}[\rho(v).\rho(v) + (Sv).(Sv)]}\mathbb{Q}||g||_{qm}}\\
&=&\frac{\frac{1}{\sqrt{d_{b+1}}}\sqrt{(\mathbb{Q}||g\mathcal{R}(v)||_2)^2 + (\mathbb{Q}||gS||_2)^2}\mathbb{Q}||v||_2}{\sqrt{(\mathbb{Q}||\rho(v)||_2)^2 + (\mathbb{Q}||Sv||_2)^2}\mathbb{Q}||g||_{qm}}\\
&=&\frac{\sqrt{(\mathbb{Q}||g\mathcal{R}(v)||_{qm})^2 + (\mathbb{Q}||gS||_{qm})^2}\mathbb{Q}||v||_2}{\sqrt{(\mathbb{Q}||\rho(v)||_2)^2 + (\mathbb{Q}||Sv||_2)^2}\mathbb{Q}||g||_{qm}}\\
&=&\frac{\sqrt{(\frac{r\mathbb{Q}||\rho(v)||_2\mathbb{Q}||g||_{qm}}{\mathbb{Q}||v||_2})^2 + (\mathbb{Q}||gS||_{qm})^2}\mathbb{Q}||v||_2}{\sqrt{(\mathbb{Q}||\rho(v)||_2)^2 + (\mathbb{Q}||Sv||_2)^2}\mathbb{Q}||g||_{qm}}\\
&=&\frac{\sqrt{(\frac{r\mathbb{Q}||Sv||_2\mathbb{Q}||g||_{qm}}{k\mathbb{Q}||v||_2})^2 + (\mathbb{Q}||gS||_{qm})^2}\mathbb{Q}||v||_2}{\sqrt{(\frac{\mathbb{Q}||Sv||_2}{k})^2 + (\mathbb{Q}||Sv||_2)^2}\mathbb{Q}||g||_{qm}}\\
&=&\frac{\sqrt{(\frac{r\mathbb{Q}||v||_2\mathbb{Q}||g||_{qm}}{k\mathbb{Q}||v||_2})^2 + (\mathbb{Q}||g||_{qm})^2}\mathbb{Q}||v||_2}{\sqrt{(\frac{\mathbb{Q}||v||_2}{k})^2 + (\mathbb{Q}||v||_2)^2}\mathbb{Q}||g||_{qm}}\\
&=&\frac{\sqrt{\frac{r^2}{k^2}+1}}{\sqrt{\frac{1}{k^2}+1}}\\
&=&\frac{\sqrt{k^2+r^2}}{\sqrt{k^2+1}}\\
&=&\frac{\sqrt{k^2 + (1 + (r-1))^2}}{\sqrt{k^2+1}}\\
&=&\frac{\sqrt{k^2 + 1 + 2(r-1) + (r-1)^2}}{\sqrt{k^2+1}}\\
&=&\sqrt{1 + \frac{2}{k^2+1}(r-1) + \frac{1}{k^2+1}(r-1)^2}\\
&=&1 + \frac{r-1}{k^2+1} + O((r-1)^2)
\end{eqnarray*}

\end{JMLRproof}

$v$ represents the incoming activation vector of some residual block and $g$ the incoming gradient. $r$ represents a type of expectation over the ratio $\frac{GSC(b+1,0)}{GSC(b,0)} = \frac{\frac{||\mathcal{J}^0_{b+1}||_{qm}||f_{b+1}||_2}{||f_0||_2}}{\frac{||\mathcal{J}^0_{b}||_{qm}||f_{b}||_2}{||f_0||_2}} = \frac{||g_{b+1}||_{qm}||f_{b+1}||_2}{||g_{b}||_{qm}||f_{b}||_2} = \frac{||g_b\mathcal{R}_b(f_{b+1})||_{qm}||f_{b+1}||_2}{||g_{b}||_{qm}||\rho(f_{b+1})||_2}=\frac{||g\mathcal{R}_b(v)||_2||v||_2}{||\rho_b(v)||_2||g||_{2}}$. Therefore $r$ can be viewed as the growth of the GSC. Similarly, $\frac{\mathbb{Q}||g\mathcal{R}_b(v) + gS_b||_2\mathbb{Q}||v||_2}{\mathbb{Q}||\rho_b(v) + S_bv||_2\mathbb{Q}||g||_2}$ represents the growth of the GSC after the skip connection has been added.

With conditions (\ref{t3c1}) and (\ref{t3c2}), we assume that the function computed by the skip block is uncorrelated to the function computed by the residual block and that the same is true for the gradient flowing through them. For the forward direction, this is true if either the skip block is Gaussian / orthogonally initialized or the last layer of the residual block is linear and Gaussian / orthogonally initialized and if the randomness of the initialization is absorbed into the expectation. Unfortunately, for the backward direction, we cannot ensure a zero correlation because there is a complex interdependence between the forward and backward evaluations of the network. However, the assumption that the forward and backward evaluations are conducted with different, independently sampled weights has been widely used and proven realistic in mean field theory based studies (see section \ref{meanFieldSection}). Assuming the forward and backward evaluations use independent weights, to fulfill condition (\ref{t3c2}), similar to condition (\ref{t3c1}), we require that either the skip connection is Gaussian / orthogonally initialized or the first layer of the residual block is linear and Gaussian / orthogonally initialized.

Condition (\ref{t3c3}) holds if $S$ is an orthogonal matrix and so specifically if $S$ is the identity matrix. If $S$ is Gaussian initialized, it holds if the randomness of the initialization is absorbed into the expectation. 

In general, if these conditions only hold approximately, it is not catastrophic to the theorem. We expect similar levels of gradient reductions in all practical MLP ResNets. 

An implicit assumption made is that the distribution of the incoming gradient $g$ is unaffected by the addition of the skip connection, which is of course not quite true in practice. The addition of the skip connection also has an indirect effect on the distribution and scale of the gradient as it flows further towards the input layer.

\subsection{Theorem \ref{dilutionComposition} - how dilution compounds}\label{theo3p5proof}

We say a random function $f_b$ is `$k$-diluted in expectation' with respect to random vector $v$, a random matrix $S_b$ and a random function $\rho_b$ if $f_b(v) = S_bv + \rho_b(v)$ and $\frac{\mathbb{Q}_{v,S_b}||S_bv||_{2}}{\mathbb{Q}_{v,\rho_b}||\rho_b(v)||_{2}} = k$.

We say a random function $\rho$ is `scale-symmetric decomposable' (SSD) if $\rho(v)$ can be written as $u_\rho\ell_\rho(\frac{v}{||v||_2})||v||_2$, where $\ell_\rho$ is a random scalar function independent of the length of $v$, and $u_\rho$ is a uniformly random unit length vector that is independent of both $\ell_\rho$ and $v$. 

We say a random matrix $S$ is `scale-symmetric decomposable' (SSD) if $Sv$, when viewed as a function of the vector $v$ is SSD.

\begin{reptheorem}{dilutionComposition}

Let $u$ be a uniformly distributed unit length vector. Consider random functions $f_b$, $1 \le b \le B$, which are $k_b$-diluted in expectation with respect to $u$, a matrix $S_b$ and a random function $\rho_b := f_b - S_b$.

Assume:

\begin{enumerate}
\item All the $S_b$ and $\rho_b$ are independent. \label{t3p5c1}
\item Each $S_b$ is either SSD or a fixed multiple of the identity. \label{t3p5c2}
\item Each $\rho_b$ is SSD. \label{t3p5c3}
\end{enumerate}

Then $f_1 \circ f_2 \circ .. \circ f_B$ is $\Big(\big(\prod_b(1+\frac{1}{k_b^2})\big)-1\Big)^{-\frac{1}{2}}$-diluted in expectation with respect to $u$, $S_1S_2 .. S_B$ and $f_1 \circ f_2 \circ .. \circ f_B - S_1S_2 .. S_B$.

\end{reptheorem}

\begin{JMLRproof}
We will prove the theorem by induction over $B$, where the induction hypothesis includes the following claims.

\begin{enumerate}
\item The claim of the theorem.
\item $\mathbb{Q}||f_1 \circ f_2 \circ .. \circ f_B(u)||_2 = \sqrt{(\mathbb{Q}||S_1S_2 .. S_Bu||_2)^2 + (\mathbb{Q}||f_1 \circ f_2 \circ .. \circ f_B(u) - S_1S_2 .. S_Bu||_2)^2}$
\item $S_1S_2 .. S_Bu$, $f_1 \circ f_2 \circ .. \circ f_B(u)$ and $f_1 \circ f_2 \circ .. \circ f_B(u) - S_1S_2 .. S_Bu$ are all radially symmetric
\end{enumerate}

Let's start by looking at the base case $B=1$. Claim (1) follows directly from the conditions of the proposition. We have:

\begin{eqnarray*}
&& \mathbb{Q}||f_1(u)||_2\\
&=& \mathbb{Q}||S_1u + \rho_1(u)||_2\\
&=& \mathbb{Q}\sqrt{(S_1u + \rho_1(u)).(S_1u + \rho_1(u))}\\
&=&\sqrt{ \mathbb{E}(S_1u + \rho_1(u)).(S_1u + \rho_1(u))}\\
&=&\sqrt{ \mathbb{E}(S_1u).(S_1u) + 2\mathbb{E}(S_1u).\rho_1(u) + \mathbb{E}\rho_1(u).\rho_1(u)}\\
&=&\sqrt{ \mathbb{E} ... + 2\mathbb{E}(S_1u).(u_{\rho_1}\ell_{\rho_1}(\frac{u}{||u||_2})||u||_2) + \mathbb{E} ...}\\
&=&\sqrt{ \mathbb{E} ... + 2\mathbb{E}_{u,S_1,\ell_{\rho_1}}(S_1u).(\ell_{\rho_1}(\frac{u}{||u||_2})||u||_2\mathbb{E}_{u_{\rho_1}}u_{\rho_1}) + \mathbb{E} ...}\\
&=&\sqrt{ \mathbb{E}(S_1u).(S_1u) + 0 + \mathbb{E}\rho_1(u).\rho_1(u)}\\
&=&\sqrt{ (\mathbb{Q}||S_1u||_2)^2 + (\mathbb{Q}||\rho_1(u)||_2)^2}\\
&=&\sqrt{ (\mathbb{Q}||S_1u||_2)^2 + (\mathbb{Q}||f_1(u) - S_1u||_2)^2}\\
\end{eqnarray*}

This is claim (2). For any $u$, $\rho_1(u)$ is radially symmetric because $\rho_1$ is SSD. If $S_1$ is SSD, $S_1u$ is also radially symmetric for arbitrary $u$. If $S_1$ is a multiple of the identity, $S_1u$ is radially symmetric because $u$ is radially symmetric. In either case, $S_1u$ is radially symmetric. Because the orientation of $\rho_1(u)$ is governed only by $u_{\rho_1}$ which is independent of both $u$ and $S_1$, the orientations of $S_1u$ and $\rho_1(u)$ are independent. But the sum of two radially symmetric random variables with independent orientations is itself radially symmetric, so $f_1(u)$ is also radially symmetric. This yields claim (3).

Now let's turn to the induction step. Set $B$ to some value and also define $k_\circ := \Big(\big(\prod_{b=2}^B(1+\frac{1}{k_b^2})\big)-1\Big)^{-\frac{1}{2}}$, $S_\circ := S_2S_3..S_B$, $f_\circ:=f_2 \circ .. \circ f_B$ and $\rho_\circ := f_\circ - S_\circ$. Then the induction hypothesis yields that $f_\circ$ is $k_\circ$-diluted in expectation with respect to $u$, $S_\circ$ and $\rho_\circ$, that $\mathbb{Q}||f_\circ(u)||_2 = \sqrt{\mathbb{Q}||S_\circ u||_2 + \mathbb{Q}||\rho_\circ(u)||_2}$ and that $f_\circ(u)$, $\rho_\circ(u)$ and $S_\circ u$ are radially symmetric. This implies that $\frac{f_\circ(u)}{||f_\circ(u)||_2}$, $\frac{\rho_\circ(u)}{||\rho_\circ(u)||_2}$ and $\frac{S_\circ u}{||S_\circ u||_2}$ are uniformly distributed unit length vectors. Define $c_\circ := \mathbb{Q}||S_\circ u||_2$. Then $k_\circ$-dilution in expectation implies $\mathbb{Q}||\rho_\circ(u)||_2 = \frac{c_\circ}{k_\circ}$ and hence $\mathbb{Q}||f_\circ(u)||_2 = \sqrt{1 + \frac{1}{k_\circ^2}}c_\circ$. Similarly, let $c_1 := \mathbb{Q}||S_1u||_2$ and then $\mathbb{Q}||\rho_1(u)||_2 = \frac{c_1}{k_1}$ and analogously to the $B=1$ case we have $\mathbb{Q}||f_1(u)||_2 = \sqrt{ (\mathbb{Q}||S_1u||_2)^2 + (\mathbb{Q}||f_1(u) - S_1u||_2)^2} = \sqrt{c_1^2 + (\frac{c_1}{k_1})^2} =  \sqrt{1 + \frac{1}{k_1^2}}c_1$. We have

\begin{eqnarray*}
&& \mathbb{Q}||\rho_1(f_\circ(u))||_2\\
&=& \sqrt{\mathbb{E}\rho_1(f_\circ(u)).\rho_1(f_\circ(u))}\\
&=& \sqrt{\mathbb{E}(u_{\rho_1}\ell_{\rho_1}(\frac{f_\circ(u)}{||f_\circ(u)||_2})||f_\circ(u)||_2).(u_{\rho_1}\ell_{\rho_1}(\frac{f_\circ(u)}{||f_\circ(u)||_2})||f_\circ(u)||_2)}\\
&=& \sqrt{\mathbb{E}_{||f_\circ(u)||_2, \frac{f_\circ(u)}{||f_\circ(u)||_2}, \rho_1}||f_\circ(u)||_2^2(u_{\rho_1}\ell_{\rho_1}(\frac{f_\circ(u)}{||f_\circ(u)||_2})).(u_{\rho_1}\ell_{\rho_1}(\frac{f_\circ(u)}{||f_\circ(u)||_2}))}\\
&=& \sqrt{\mathbb{E}_{||f_\circ(u)||_2}||f_\circ(u)||_2^2\mathbb{E}_{\frac{f_\circ(u)}{||f_\circ(u)||_2}, \rho_1}(u_{\rho_1}\ell_{\rho_1}(\frac{f_\circ(u)}{||f_\circ(u)||_2})).(u_{\rho_1}\ell_{\rho_1}(\frac{f_\circ(u)}{||f_\circ(u)||_2}))}\\
&=& \sqrt{(1 + \frac{1}{k_\circ^2})c_\circ^2\mathbb{E}_{\frac{f_\circ(u)}{||f_\circ(u)||_2}, \rho_1}\rho_1(\frac{f_\circ(u)}{||f_\circ(u)||_2}).\rho_1(\frac{f_\circ(u)}{||f_\circ(u)||_2})}\\
&=& \sqrt{(1 + \frac{1}{k_\circ^2})c_\circ^2(\mathbb{Q}||\rho_1(\frac{f_\circ(u)}{||f_\circ(u)||_2})||_2)^2}\\
&=& \sqrt{1 + \frac{1}{k_\circ^2}}c_\circ\frac{c_1}{k_1}\\
\end{eqnarray*}

If $S_1$ is SSD, analogously, we have $\mathbb{Q}||S_1S_\circ u||_2 = c_\circ c_1$, $\mathbb{Q}||S_1\rho_\circ(u)||_2 = \frac{c_\circ}{k_\circ}c_1$ and $\mathbb{Q}||S_1f_\circ(u)||_2 = \sqrt{1+\frac{1}{k_\circ^2}}c_\circ c_1$. If $S_1$ is a multiple of the identity, by the definition of $c_1$, it is $c_1$ times the identity. Therefore $\mathbb{Q}||S_1S_\circ u||_2 = \mathbb{Q}||c_1S_\circ u||_2 = c_\circ c_1$, $\mathbb{Q}||S_1\rho_\circ(u)||_2 = \mathbb{Q}||c_1\rho_\circ(u)||_2  = \frac{c_\circ}{k_\circ}c_1$ and $\mathbb{Q}||S_1f_\circ(u)||_2 = \mathbb{Q}||c_1f_\circ(u)||_2 = \sqrt{1+\frac{1}{k_\circ^2}}c_\circ c_1$ also hold. Then we have

\begin{eqnarray*}
&& \mathbb{Q}||f_1(f_\circ(u)) - S_1S_\circ u||_2\\
&& \mathbb{Q}||S_1(f_\circ(u)) + \rho_1(f_\circ(u)) - S_1S_\circ u||_2\\
&& \mathbb{Q}||S_1(S_\circ u + \rho_\circ(u)) + \rho_1(f_\circ(u)) - S_1S_\circ u||_2\\
&& \mathbb{Q}||S_1\rho_\circ(u) + \rho_1(f_\circ(u))||_2\\
&=&\sqrt{ \mathbb{E}(S_1\rho_\circ(u)).(S_1\rho_\circ(u)) + \mathbb{E}(S_1\rho_\circ(u)).\rho_1(f_\circ(u)) + \mathbb{E}\rho_1(f_\circ(u)).\rho_1(f_\circ(u))}\\
&=&\sqrt{ \mathbb{E} ... + 2\mathbb{E}(S_1\rho_\circ(u)).(u_{\rho_1}\ell_{\rho_1}(\frac{f_\circ(u)}{||f_\circ(u)||_2})||f_\circ(u)||_2) + \mathbb{E} ... }\\
&=&\sqrt{ \mathbb{E} ... + 2\mathbb{E}_{f_\circ(u),\rho_\circ(u),S_1,\ell_{\rho_1}}(S_1\rho_\circ(u)).(\ell_{\rho_1}(\frac{f_\circ(u)}{||f_\circ(u)||_2})||f_\circ(u)||_2\mathbb{E}_{u_{\rho_1}}u_{\rho_1}) + \mathbb{E} ... }\\
&=&\sqrt{ \mathbb{E}(S_1\rho_\circ(u)).(S_1\rho_\circ(u)) + 0 + \mathbb{E}\rho_1(f_\circ(u)).\rho_1(f_\circ(u))}\\
&=&\sqrt{ (\mathbb{Q}||S_1\rho_\circ(u)||_2)^2 + (\mathbb{Q}||\rho_1(f_\circ(u))||_2)^2}\\
&=&\sqrt{ \Big(\frac{c_\circ}{k_\circ}c_1\Big)^2 + \Big(\sqrt{1 + \frac{1}{k_\circ^2}}c_\circ\frac{c_1}{k_1}\Big)^2}\\
&=&c_\circ c_1\sqrt{ \frac{1}{k_\circ^2} + (1 + \frac{1}{k_\circ^2})\frac{1}{k_1^2}}\\
&=&c_\circ c_1\sqrt{ -1 + (1 + \frac{1}{k_\circ^2})(1 + \frac{1}{k_1^2})}\\
\end{eqnarray*}

And analogously we have

\begin{eqnarray*}
&& \mathbb{Q}||f_1(f_\circ(u))||_2\\
&& \mathbb{Q}||S_1(f_\circ(u)) + \rho_1(f_\circ(u))||_2\\
&=&\sqrt{ (\mathbb{Q}||S_1f_\circ(u)||_2)^2 + (\mathbb{Q}||\rho_1(f_\circ(u))||_2)^2}\\
&=&\sqrt{ \Big(\sqrt{1 + \frac{1}{k_\circ^2}}c_\circ c_1\Big)^2 + \Big(\sqrt{1 + \frac{1}{k_\circ^2}}c_\circ\frac{c_1}{k_1}\Big)^2}\\
&=&c_\circ c_1\sqrt{(1 + \frac{1}{k_\circ^2})(1 + \frac{1}{k_1^2})}\\
\end{eqnarray*}

So $\frac{\mathbb{Q}||S_1S_\circ u||_2}{\mathbb{Q}||f_1(f_\circ(u)) - S_1S_\circ u||_2} = \frac{c_\circ c_1}{c_\circ c_1\sqrt{ -1 + (1 + \frac{1}{k_\circ^2})(1 + \frac{1}{k_1^2})}} = (-1 + (1 + \frac{1}{k_\circ^2})(1 + \frac{1}{k_1^2}))^{-\frac{1}{2}}$. Substituting back $k_\circ$, $S_\circ$ and $f_\circ$, we obtain claim (1). Claim (2), when substituting in $k_\circ$, $S_\circ$ and $f_\circ$ becomes $ \mathbb{Q}||f_1(f_\circ(u))||_2 = \sqrt{(\mathbb{Q}||S_1S_\circ u||_2)^2 + (\mathbb{Q}||f_1(f_\circ(u)) - S_1S_\circ u||_2)^2}$. Substituting the identities we obtained results in 

$c_\circ c_1\sqrt{(1 + \frac{1}{k_\circ^2})(1 + \frac{1}{k_1^2})} = \sqrt{(c_\circ c_1)^2 + \Big(c_\circ c_1\sqrt{ -1 + (1 + \frac{1}{k_\circ^2})(1 + \frac{1}{k_1^2})}\Big)^2}$. This is a true statement, so we have claim (2). 

Consider $S_1S_2 .. S_Bu = S_1S_\circ u$. We know $S_\circ u$ is radially symmetric by the induction hypothesis, so if $S_1$ is a multiple of the identity, so is $S_1S_\circ u$. If $S_1$ is SSD, then $S_1S_\circ u$ is radially symmetric for any value of $S_\circ u$. In either case, $S_1S_\circ u$ is radially symmetric.

Consider $f_1f_2 .. f_Bu = f_1f_\circ u$. We know $f_\circ u$ is radially symmetric by the induction hypothesis. We also have $f_1f_\circ u = S_1f_\circ u + \rho_1(f_\circ u)$. We just showed $S_1f_\circ u$ is radially symmetric. Because $\rho_1$ is SSD, $\rho_1(f_\circ u)$ is radially symmetric with an orientation independent of that of $S_1f_\circ u$ because it is governed only by $u_{\rho_1}$. The sum of two radially symmetric random variables with independent orientation is itself radially symmetric, so $f_1f_\circ u$ is radially symmetric.

Finally, consider $f_1f_2 .. f_Bu - S_1S_2 .. S_Bu = S_1\rho_\circ(u) + \rho_1(f_\circ(u))$. We know $\rho_\circ(u)$ is radially symmetric by the induction hypothesis so as before, $S_1\rho_\circ(u)$ is radially symmetric. And again, $\rho_1(f_\circ(u))$ is radially symmetric with independent orientation, so the sum $f_1f_2 .. f_Bu - S_1S_2 .. S_Bu$ is radially symmetric. 

So we also have claim (3). This completes the proof.

\end{JMLRproof}

Condition (\ref{t3p5c1}) is standard for randomly initialized networks.

Condition (\ref{t3p5c2}) is fulfilled if the skip block $S_b$ is either a multiple of the identity or orthogonally or Gaussian initialized. Hence, it is fulfilled for all popular skip block types.

Condition (\ref{t3p5c3}) is satisfied, for example, by any residual block consisting only of linear and ReLU layers, where the final layer is Gaussian or orthogonally initialized. In general, condition (\ref{t3p5c3}) is only fulfilled approximately in practice. If the last operation of each $\rho_b$ is multiplication with an SSD matrix, a popular choice, the orientation of $\rho_b$ is indeed governed by an independent, uniformly random unit length vector $u_\rho$ as required. However, $\rho_b$ generally does not preserve the length of the incoming vector $||v||_2$ as required by (\ref{t3p5c3}). As for theorems \ref{explodeTheorem} and corollary \ref{bigGradCorollary}, we appeal to the fact that due to the law of large numbers, the lengths of latent representations corresponding to different data inputs do not vary much in practical networks, and thus we can approximate both the input and output lengths with constants.

.

\section{Taylor approximation of a neural network} \label{taylorDetailsSection}

We define the first-order Taylor approximation $T_l$ of the bottom layers up to layer $l$ recursively. Write $i_l(X)$ as the short form of $i_l(i_{l+1}(..i_L(X)..))$. Then

\begin{eqnarray*}
T_L(\theta, X) &=& f_L(\theta_L,X)\\
T_l(\theta, X) &=& i_l(T_{l+1}(\theta,X)) + r_l(\theta_l,i_{l+1}(X)) + \sum_{k=l+1}^L \frac{dr_l(\theta_l,i_{l+1}(X))}{di_k(X)}r_k(\theta_k,i_{k+1}(X)) \text{ for } l < L
\end{eqnarray*}

The maximum number of parametrized residual functions composed in $T_l$ is 2. Otherwise, only addition and composition with fixed functions is used. Hence, the compositional depth of $T_l$ is $\min(L-l,2)$. Hence, the network $f_{\text{Taylor}(l)} := f_0(y,f_1(..f_{l-1}(T_l(X))..))$ has compositional depth $\max(l + 1,L)$.

For ResNet architectures, as in section \ref{depthComputationSection}, we divide each layer in the residual block into its initial and residual function. Then the definition of the Taylor expansion remains as above, except a term $s_l(T_m(\theta, X))$ is added at each layer $l$ where a skip connection, represented by skip block $s_l$, terminates. $T_m$ is the Taylor expansion at the layer where the skip connection begins.

\section{Looks-linear initialization} \label{looklineardetails}

The looks-linear initialization (`LLI') of ReLU MLPs achieves an approximate orthogonal initial state. Consider a ReLU MLP with some number of linear layers and a ReLU layer between each pair of linear layers. LLI initializes the weight matrix of the lowest linear layer differently from the weight matrix of the highest linear layer and differently from the weight matrices of the intermediate linear layers. Consider an $m \times n$-dimensional weight matrix $W$, where $n$ is the dimensionality of the incoming vector and $m$ is the dimensionality of the linear layer itself. Also, we require that the dimensionality of all ReLU layers, and thus the dimensionality of all linear layers except the highest linear layer, is even. Then the weight matrices are initialized as follows.

\begin{itemize}
\item Lowest linear layer: Draw a uniformly random orthogonal $\max(\frac{m}{2},n) * \max(\frac{m}{2},n)$-dimensional matrix $W'$. Then, for all $1 \le i \le \frac{m}{2}$ and $1 \le j \le n$, set $W(2i,j) = \max(\sqrt{\frac{m}{2n}},1)W'(i,j)$ and $W(2i+1,j) = -\max(\sqrt{\frac{m}{2n}},1)W'(i,j)$.
\item Highest linear layer: Draw a uniformly random orthogonal $\max(m,\frac{n}{2}) * \max(m,\frac{n}{2})$-dimensional matrix $W'$. Then, for all $1 \le i \le m$ and $1 \le j \le \frac{n}{2}$, set $W(i,2j) = \max(\sqrt{\frac{2m}{n}},1)W'(i,j)$ and $W(i,2j+1) = -\max(\sqrt{\frac{2m}{n}},1)W'(i,j)$.
\item Intermediate linear layers: Draw a uniformly random orthogonal $\max(\frac{m}{2},\frac{n}{2}) * \max(\frac{m}{2},\frac{n}{2})$-dimensional matrix $W'$. Then, for all $1 \le i \le \frac{m}{2}$ and $1 \le j \le \frac{n}{2}$, set $W(2i,2j) = \max(\sqrt{\frac{m}{n}},1)W'(i,j)$, $W(2i+1,2j) = -\max(\sqrt{\frac{m}{n}},1)W'(i,j)$, \\$W(2i,2j+1) = -\max(\sqrt{\frac{m}{n}},1)W'(i,j)$ and $W(2i+1,2j+1) = \max(\sqrt{\frac{m}{n}},1)W'(i,j)$.
\end{itemize}

Under LLI, pairs of neighboring ReLU neurons are grouped together to effectively compute the identity function. The incoming signal is split between ReLU neurons of even and odd indeces. Each of the two groups preserves half the signal, which are then ``stitched together'' in the next linear layer only to be re-divided in a different way to pass through the next ReLU layer.

An LLI initialized ReLU network can be said to be approximately orthogonal. Let $X_{l_r}$ be the representation computed by the $r$'th ReLU layer. Then let $\chi_{l_r}'(i) = X_{l_r}(2i) - X_{l_r}(2i+1)$ for $1 \le i \le \frac{d_{l_r}}{2}$. Since $X_{l_r}(2i)$ or $X_{l_r}(2i+1)$ is 0, this transformation is bijective. Then the transformation from $\chi_{l_r}$ to $\chi_{l_{r-1}}$ is an orthogonal transformation.

\section{Experimental details} \label{detailssection}

We conducted three types of experiments in this paper. 

\begin{itemize}
\item `Gaussian noise experiments': Static experiments where the forward and backward dynamics were investigated in the initialized state as the network was presented with Gaussian noise inputs and labels. Results are shown in figures \ref{GSCfigure} and \ref{GSCresfigure} as well as tables \ref{gaussianTable} and \ref{GSCdynamic}. See section \ref{gaussianProtocol} for details.
\item `color map experiments': A single input from the CIFAR10 dataset is evaluated by a dense cluster of weight configurations to investigate the functional complexity of an architecture. Results are shown in figures \ref{im_expl} and \ref{im_collapse}. See section \ref{colorMapProtocol} for details.
\item `CIFAR10 experiments': Networks were trained on CIFAR10 to track training error and other important metrics such as effective depth on a popular dataset. Results are shown in figures \ref{dyna}, \ref{dynaCollapse}, \ref{OIS}, \ref{dynaRes} and \ref{dynaLL} as well as table \ref{CIFARtable}. See section \ref{cifarProtocol} for details.
\end{itemize}

See the upcoming section for details about the neural architectures used.

\subsection{Architectures used} \label{architectureSection}

\paragraph{Vanilla networks without skip connections}

All networks are MLPs composed of only fully-connected linear layers and unparametrized layers. The following types of layers are used.

\begin{itemize}
\item linear layer: $f_l(\theta_l, F_{l+1}) = W_lF_{l+1}$ where the entries of the weight matrix $W_l$ are the entries of the parameter sub-vector $\theta_l$. Trainable bias parameters are not used.
\item ReLU layer: $f_l(F_{l+1}) = \sigma_{\text{ReLU}}.(F_{l+1})$, where the scalar function $\sigma_{\text{ReLU}}$ is applied elementwise as indicated by $.()$ We have $\sigma_{\text{ReLU}}(a) = a$ if $a \ge 0$ and $\sigma_{\text{ReLU}}(a) = 0$ if $a < 0$.
\item leaky ReLU layer: $f_l(F_{l+1}) = \sigma_{\text{leaky ReLU}}.(F_{l+1})$, where $\sigma_{\text{leaky ReLU}}(a) = a$ if $a \ge 0$ and $\sigma_{\text{leaky ReLU}}(a) = ca$ if $a < 0$. $c$ is the leakage parameter, which we varied between 0 and 1. $c$ is fixed during training.
\item tanh layer: $f_l(F_{l+1}) = \sigma_{\text{tanh}}.(F_{l+1})$, where $\sigma_{\text{tanh}}(a) = \tanh(a)$.
\item SELU layer: $f_l(F_{l+1}) = \sigma_{\text{SeLU}}.(F_{l+1})$. We have $\sigma_{\text{SeLU}}(a) = c_{\text{pos}}a$ if $a \ge 0$ and $\sigma_{\text{SeLU}}(a) = c_{\text{neg}}(e^a - 1)$ if $a < 0$. We set $c_{\text{pos}} = 1.0507$ and  $c_{\text{neg}} = 1.0507 * 1.6733$ as suggested by \cite{selu}.
\item batch normalization layer: $f_l(F_{l+1}) = \frac{F_{l+1} - \mu}{\sigma}$, where $\mu$ is the component-wise mean of $F_{l+1}$ over the current batch and $\sigma$ is the componentwise standard deviation of $F_{l+1}$ over the current batch.
\item layer normalization layer: $f_l(F_{l+1}) = \frac{F_{l+1} - \mu}{\sigma}$, where $\mu$ is mean of the entries of $F_{l+1}$ and $\sigma$ is the standard deviation of the entries of $F_{l+1}$.
\item dot product error layer: $f_0(F_1,y) = F_1.y$
\item softmax layer: $f_l(F_{l+1})(i) = \frac{e^{F_{l+1}(i)}}{\sum_j e^{F_{l+1}(j)}}$
\item cross-entropy error layer: $f_0(F_1,y) = \ln F_1(y)$ where $y$ is an integer class label and $F_1$ has one entry per class.
\end{itemize}

Note that normalization layers do not use trainable bias and variance parameters.

A network of compositional depth $N$ contains $N$ linear layers and $N-1$ nonlinearity layers (ReLU, leaky ReLU, tanh or SELU) inserted between those linear layers. If the network uses normalization layers, one normalization layer is inserted after each linear layer. For Gaussian noise experiments, the error layer is the dot product error layer. For CIFAR10 experiments, a softmax layer is inserted above the last linear or normalization layer and the error layer is the cross-entropy error layer. For color map experiments, we did not use error layers.

For Gaussian noise experiments, data inputs as well as predictions and labels have dimensionality 100. We used a compositional depth of 50. We generally used a uniform width of 100 throughout the network. However, we also ran experiments where the width of all layers from the first linear layer to the layer before the last linear layer had width 200. We also ran experiments where linear layers alternated in width between 200 and 100. For CIFAR10 experiments, data inputs have dimensionality 3072 and predictions have dimensionality 10. We use a compositional depth of 51. The first linear layer transforms the width to 100 and the last linear layer transforms the width to 10. For color map experiments, networks receive a single data input of dimensionality 3072 and predictions have dimensionality 3. We use a compositional depth of 50. The first linear layer transforms the width to 100 and the last linear layer transforms the width to 3.

We now describe the schemes used in this paper to initialize weight matrices. Note that after each weight matrix is generated via one of those schemes as outlined below, if the corresponding linear layer is directly preceded by a ReLU layer, the matrix is multiplied by a factor of $\sqrt{2}$. This follows the convention introduced by \cite{heInit} and ensures stability of forward activations in ReLU networks. We did not use the $\sqrt{2}$ factor when using looks-linear initialization, as this type of initialization already takes this factor into account.

\begin{itemize}
\item Gaussian: Each entry of the weight matrix is drawn independently from a Gaussian distribution with mean 0 and variance equal to one over the dimensionality of the incoming vector.
\item orthogonal: An $m \times n$-dimensional weight matrix is set to be an $m \times n$-submatrix of a $\max(m,n) \times \max(m,n)$ uniformly random orthogonal matrix, multiplied by $\max(1,\sqrt{\frac{m}{n}})$.
\item looks-linear: See section \ref{looklineardetails}. 
\item skew-symmetric: The matrix is set to $\frac{C - C^T}{||C - C^T||_{qm}}$, where $C$ is Gaussian initialized.
\item $-C^TC$: The matrix is set to $\frac{-C^TC}{||-C^TC||_{qm}}$, where $C$ is Gaussian initialized.
\item correlated Gaussian: The matrix is set to $\frac{\sqrt{\frac{3}{4}}C + \frac{1}{2}W_{l-1}}{||\sqrt{\frac{3}{4}}C + \frac{1}{2}W_{l-1}||_{qm}}$, where $C$ is Gaussian initialized and $W_{l-1}$ is the weight matrix of the preceding linear layer.
\end{itemize}

The reason for normalizing the qm norm under various schemes is to ensure consistency with the Gaussian and orthogonal initialization schemes. The qm norm of Gaussian initialized matrices is 1 in quadratic expectation and the qm norm of orthogonally initialized weight matrices is exactly 1. Also note that only the Gaussian, orthogonal and looks-linear initialization schemes are ever used for matrices that are not square. The correlated Gaussian initialization scheme is only used when the weight matrix as well as the weight matrix of the preceding linear layer is square.

\paragraph{ResNet}

In all cases, the first layer is a linear layer. For experiments discussed in section \ref{ResNetsection}, this is followed by 25 blocks. Each skip connection bypasses a residual block of 6 layers: a normalization layer, a nonlinearity layer, a linear layer, another normalization layer, another nonlinearity layer, and another linear layer. For experiments discussed in section \ref{dynamicalSection}, the first linear layer is followed by 50 blocks. Each skip connection bypasses a residual block of 3 layers: a normalization layer, a nonlinearity layer and a linear layer.

Above the last block, a final normalization layer is inserted, followed by softmax (CIFAR10 only) and then the error layer. For Gaussian noise experiments, we use a constant width of 100. For CIFAR10, the first linear layer transforms the width from 3072 to 100, and the last skip connection as well as the last linear linear in the last residual block transform the width from 100 to 10.

Regarding skip blocks, networks fall in one of two categories. In the first category, skip blocks are identity skip blocks. The only exception to this is the highest skip block in CIFAR10 experiments. This highest skip block is responsible for reducing the width of the incoming vector. It multiplied that vector with an orthogonally initialized matrix. In the second category, all skip blocks multiply the incoming vector with a Gaussian initialized matrix.

For experiments discussed in section \ref{ResNetsection}, all weight matrices in linear layers are Gaussian initialized. For experiments discussed in section \ref{dynamicalSection}, the first linear layer (below the first block) as well as the last linear layer (in the last block) are Gaussian initialized. All other linear layers use the Gaussian, skew-symmetric, $-C^TC$ or correlated Gaussian scheme. Note that under the correlated Gaussian scheme, the second lowest linear layer (in the lowest block) is still Gaussian initialized because there is no other weight matrix to correlate that weight matrix with.

\subsection{Protocol for Gaussian noise experiments} \label{gaussianProtocol}

For Gaussian noise experiments, both inputs and labels are 100-dimensional vectors where each entry is drawn independently from a Gaussian with mean $0$ and variance $\frac{1}{100}$. We normalized the input vectors to have exactly length 10. For each architecture we studied (see table \ref{gaussianTable} for the full list), we drew 100 datasets of 10.000 datapoints each and associated each dataset with its own random weight initialization. We computed both the forward activations and the gradients for each datapoint. For architectures with batch normalization, all 10.000 datapoints within each dataset were considered part of a single batch. Note that no training was conducted. We then computed the following metrics for each architecture and dataset:

\begin{itemize}
\item $GSC(l,0)$: At each layer $l$, we computed $\frac{\mathbb{Q}_D||\mathcal{J}^0_l||_{qm}\mathbb{Q}_D||f_l||_{2}}{\mathbb{Q}_D||f_0||_2}$. Note that $\mathcal{J}^0_l$ is simply the ``regular gradient'' of the network. The expectation is taken over the dataset. For ResNets, we computed $GSC(b,0)$ only between blocks, at the end of the last block and at the beginning of the first block.
\item Pre-activation standard deviation: For each nonlinearity layer $l$, we computed 

$\sqrt{\frac{1}{d_{l+1}} \sum_{i=1}^{d_{l+1}} \mathbb{E}_D [f_{l+1}(i)^2] - (\mathbb{E}_D f_{l+1}(i))^2}$
\item Pre-activation quadratic expectation: For each nonlinearity layer $l$, we computed 

$\sqrt{\frac{1}{d_{l+1}}\sum_{i=1}^{d_{l+1}} \mathbb{E}_D [f_{l+1}(i)^2]}$
\item Pre-activation domain bias: For each nonlinearity layer $l$, we computed 

$1 - \frac{\sum_{i=1}^{d_{l+1}} (\mathbb{E}_D f_{l+1}(i))^2}{\sum_{i=1}^{d_{l+1}} \mathbb{E}_D [f_{l+1}(i)^2]}$
\item Pre-activation sign diversity: For each nonlinearity layer $l$, we computed

$\frac{1}{d_{l+1}}\sum_{i=1}^{d_{l+1}} \min(\mathbb{E}_D\mathbbm{1}(f_{l+1}(i) > 0), \mathbb{E}_D\mathbbm{1}(f_{l+1}(i) < 0))$, where $\mathbbm{1}$ is the indicator function of a boolean expression.
\item Linear approximation error: For each nonlinearity layer $l$ and neuron in that layer $i$ where $1 \le i \le d_l$, we compute constants $a_i$ and $b_i$ such that $\mathbb{E}_D[(f_l(i)-(a_if_{l+1}(i)+b_i))^2]$ is minimized. This is the least squares linear approximation of that neuron. The linear approximation error is then $1 - \frac{\sum_{i=1}^{d_{l+1}} \mathbb{E}_D [(a_if_{l+1}(i)+b_i)^2]}{\sum_{i=1}^{d_{l+1}} \mathbb{E}_D [f_{l}(i)^2]}$
\item Dilution level (ResNet only): For each block $s_b + \rho_b$, we computed $k_b = \frac{\mathbb{Q}_D||s_b||_2}{\mathbb{Q}_D||\rho_b||_2}$. We obtain one value for each block.
\item $GSC(b,0)$ dilution-corrected (ResNet only): This was computed recursively as 

$GSC_{corr}(1,0) = GSC(1,0)$ and $GSC_{corr}(b+1,0) = GSC(b,0)(1 + (k_b^2+1)(\frac{GSC(b+1,0)}{GSC(b,0)}-1))$, where $GSC(1,0)$ and $GSC_{corr}(1,0)$ denote the (corrected) GSC between the end of the highest block and the error layer.
\end{itemize}

Finally, we averaged the results over the 100 datasets / initializations.

For ResNets, the pre-activations of a nonlinearity layer were the activations of the preceding layer within the same block. 

\subsection{Protocol for color map experiments} \label{colorMapProtocol}

For a given architecture, a single random Gaussian initialization $\theta$ for all weights in the network is chosen. For a single linear layer $l$, three different random Gaussian initializations are chosen. Call them $\theta^1_l$, $\theta^2_l$ and $\theta^3_l$. Consider points $(x,y,z)$ on the unit sphere in $\mathbb{R}^3$. We associate each of these points with the weight configuration that is equal to $\theta$ at all layers except $l$, and equal to $x\theta^1_l + y\theta^2_l + z\theta^3_l$ at layer $l$. Because each $\theta^i_l$ is Gaussian initialized, for each fixed tuple $(x,y,z)$, $x\theta^1_l + y\theta^2_l + z\theta^3_l$ is also Gaussian initialized. We call the sphere of points associated with these weight configurations the ``weight sphere''.

For each of these weight configuration, we take the input shown in figure \ref{im_expl}A from the CIFAR10 dataset and propagate it forward through the network with that weight configuration. We obtain a 3-dimensional output at the prediction layer, which we divide by its length. Now the output lies on the unit sphere in $\mathbb{R}^3$. Each point on that ``output sphere'' is associated with a color as shown in figure \ref{im_expl}B. Finally, we color each point on the weight sphere according to its respective color on the output sphere. The colored weight spheres are shown in figure \ref{im_expl}C-K as well as figure \ref{im_collapse}. 

All spheres are shown as azimuthal projections. The center of the depicted discs corresponds to the point $(\frac{\sqrt{3}}{3},\frac{\sqrt{3}}{3},\frac{\sqrt{3}}{3})$. The RGB values of colors on the output sphere are chosen so that the R component is largest whenever the first output neuron is largest, the G component is largest whenever the second output neuron is largest and the B component is largest whenever the third output neuron is largest. If we imagine that the prediction is fed into a softmax layer for 3-class classification, then ``purer'' colors correspond to more confident predictions.

\subsection{Protocol for CIFAR10 experiments} \label{cifarProtocol}

For CIFAR10 experiments, we preprocessed each feature to have zero mean and unit variance. We used the training set of 50.000 datapoints and disregarded the test set. We used batches of size 1.000 except for the vanilla batch-ReLU architecture with Gaussian initialization, for which we used a batch size of 50.000. (See section \ref{batchNormSection} for the explanation.)

We trained each architecture we studied (see table \ref{CIFARtable} for the full list) with SGD in one or both of two ways: with a single step size for all layers; and with a custom step size for each layer.

\paragraph{Single step size} We perform a grid search over the following starting step sizes: $\{10^5, 3*10^4, 10^4, 3*10^3, .., 10^{-4}, 3*10^{-5}, 10^{-5}\}$. For each of those 21 starting step sizes, we train the network until termination. This is done as follows. We train with the starting step size until the end-of-epoch training classification error across the whole dataset has not decreased for 5 consecutive epochs. Once that point is reached, the step size is divided by 3 and training continues. Once the error has again not decreased for 5 epochs, the step size is divided by 3 again. This process is repeated until training terminates. Termination occurs either after 500 epochs or after the step size is divided 11 times, whichever comes first. The starting step size that obtains the lowest final training classification error is selected as the representative step size for which results are presented in the paper.

\paragraph{Custom step sizes} In this scenario, we use a different starting step size for each layer. After the network is pre-trained and those step sizes are computed, smoothed and scaled as described in section \ref{customSection}, we continue training the pre-trained network with those step sizes. As before, when the error has not improved for 5 epochs, we divide all step sizes jointly by 3. And as before, training is terminated after 11 divisions or when 500 epochs are reached, whichever comes first.\\

In all experiments, we compute the following metrics:

\begin{itemize}
\item Largest relative update size for each layer induced by the estimated optimal step size during the epoch the estimate was computed. See section \ref{customSection} for details.
\item Effective depth throughout training: see section \ref{depthComputationSection} for details. $\lambda$-contributions are accumulated from batch to batch.
\item Training classification error across the whole dataset at the end of each epoch.
\item Training classification error when compositional depth is reduced via Taylor expansion after training: see section \ref{taylorDetailsSection} for details.
\item $GSC(L+1,0)$, pre-activation standard deviation and pre-activation sign diversity: for details, see the end of section \ref{gaussianProtocol}. Note that the expectations over the dataset were computed by maintaining exponential running averages across batches.
\item Operator norms of residual weight matrices after training.
\end{itemize}

\subsubsection{Selecting custom step sizes} \label{stepSizeSection} \label{customSection}

We estimated the optimal step size for each linear layer under SGD for our CIFAR10 experiments. This turned out to be more difficult than expected. Below, we describe the algorithm we used. It has five stages. 

\paragraph{Pre-training} We started by pre-training the network. We selected a set of linear layers in the network that we suspected would require similar step sizes. In exploding architectures (vanilla batch-ReLU with Gaussian initialization, vanilla layer-tanh, vanilla batch-tanh, SELU), we chose the second highest linear layer through the sixth highest linear layer for pretraining, i.e. 5 linear layers in total. We expected these layers to require a similar step size because they are close to the output and the weight matrices have the same dimensionality. For vanilla ReLU, vanilla layer-ReLU, vanilla tanh and looks-linear initialization, we chose the second lowest linear layer through the second highest linear layer (i.e. 49 linear layers in total) because the weight matrices have the same dimensionality. Finally, for ResNet, we chose the second lowest through the third highest linear layer (i.e. 48 linear layers in total), because the blocks those layers are in have the same dimensionality.

We then trained those layers with a step size that did not cause a single relative update size of more than 0.01 (exploding architectures) or 0.001 (other architectures) for any of the pre-trained layers or any batch. We chose small step sizes for pre-training to ensure that pre-training would not impact effective depth. We pre-trained until the training classification error reached 85\%, but at least for one epoch and at most for 10 epochs. The exact pre-training step size was chosen via grid search over a grid with multiplicative spacing of 3. The step size chosen was based on which step size reached the 85\% threshold the fastest. Ties were broken by which step size achieved the lowest error.

\paragraph{Selection} In the selection stage, we train each linear layer one after the other for one epoch while freezing the other layers. After each layer is trained, the change to the parameter caused by that epoch of training is undone before the next layer is trained. For each layer, we chose a step size via grid search over a grid with multiplicative spacing 1.5. The step size that achieved the lowest training classification error after the epoch was selected. Only step sizes that did not cause relative update sizes of 0.1 or higher were considered, to prevent weight instability.

Now we can explain the need for pre-training. Without pre-training, the selection stage yields very noisy and seemingly random outcomes for many architectures. This is because it was often best to use a large step size to jump from one random point in parameter space to the next, hoping to hit a configuration at the end of the epoch where the error was, say, 88\%. Since we used a tight spacing of step sizes, for most layers, there was at least one excessively large step size that achieved this spurious ``success''. Since we only trained a single layer out of 51 for a single epoch, the error of the ``correct'' step size after pre-training often did not reach, say, 88\%. When we trained the network for 500 epochs with those noisy estimates, we obtained very high end-of-training errors. Pre-training ensures that jumping to a different point in parameter space causes the error to exceed 85\% again. Therefore, this behavior is punished and step sizes that ultimately lead to a much better end-of-training error are selected.

\paragraph{Clipping}

Even though pre-training was used, for some architectures, it was still beneficial to add the following restriction: as we consider larger and larger step sizes during grid search, as soon as we find a step size for which the error after the training epoch is at least, in absolute terms, $0.1\%$ higher than for the current best step size, the search is terminated. Clipping is capable of further eliminating outliers and was used if and only if it improved the end-of-training error. It was used for vanilla tanh, ResNet layer-tanh and looks-linear layer-ReLU.

For each linear layer, the largest relative update size induced by the step size obtained for that layer after the clipping stage (or after the selection stage if clipping was not used) during the epoch of training conducted in the selection stage is shown in figures \ref{dyna}A, \ref{dynaCollapse}A, \ref{dynaRes}A and \ref{dynaLL}A.

\paragraph{Smoothing}

In this stage, we built a mini-regression dataset of $(X,Y)$ points as follows. For each $X$ from 1 to 51, we include the point $(X,Y)$ where $Y$ is the largest relative update size the step size selected for linear layer $X$ after clipping induced during the epoch of training in the selection stage. We then fit a line via least-squares regression on that mini-dataset in log scale. For each $X$, we thus obtain a smoothed value $Y'$. The ratio $\frac{Y'}{Y}$ was multiplied to the step size obtained for each layer at the end of the clipping stage.

The smoothing stage essentially replaces the relative update sizes in figures \ref{dyna}A, \ref{dynaCollapse}A, \ref{dynaRes}A and \ref{dynaLL}A with their best linear fit. We added this stage because we found that the end-of-training error could still be significantly improved by reducing noise among the layer-wise step sizes. 

\paragraph{Scaling}

Finally, we jointly scale all layer-wise step sizes with a single constant. That value is chosen as in the selection stage by trying a small constant, training for one epoch, rewinding that epoch, multiplying that constant by 1.5, rinse, repeat. Again, that process was terminated once any layer experiences an update of relative size at least 0.1. The scaling stage is necessary because the size of the update of the entire parameter vector when all layers are trained jointly is $\approx \sqrt{51}$ times larger than when only single layers are trained as in the selection stage. Hence, a scaling constant less than 1 is usually needed to compensate. Again, some architectures benefited from using clipping, where we terminated the scaling constant search as soon as one exhibited an error more than 0.1\% above the current best scaling constant. Vanilla tanh, vanilla layer-tanh, ResNet layer-tanh and looks-linear layer-ReLU used this clipping.

Formally, for each architecture, we trained three networks to completion. One using no clipping, one using only clipping during the scaling stage, and one using the clipping stage as well as clipping during the scaling stage. Whichever of these three networks had the lowest end-of-training error was selected for presentation in the paper. To compare, for single step size training, we compared 21 end-of-training error values.


\vskip 0.2in
\bibliography{explodingGradient}

\end{document}